%% file: colm2024_conference.tex
\documentclass{article} %
\usepackage{colm2024_conference}
\usepackage[T1]{fontenc}
\usepackage{lmodern}
\usepackage{booktabs}
\usepackage{graphicx}
\usepackage{enumitem}
\usepackage{wrapfig}
\usepackage{algorithm}
\usepackage{algpseudocode}

\usepackage{microtype}
\usepackage{amsmath}
\usepackage{colortbl}
\usepackage[utf8]{inputenc}
\definecolor{lightgray}{rgb}{0.9,0.9,0.9}
\usepackage{caption}
\usepackage{subcaption}
\usepackage{xcolor}
\usepackage{setspace}
\usepackage{url}
\usepackage{multirow}
\usepackage{colortbl}
\usepackage{tabularx}
\usepackage{blindtext}
\usepackage{pgfplots}
\pgfplotsset{compat=1.18} 
\usepackage{tikz}
\usetikzlibrary{er,positioning,bayesnet}
\usepackage{makecell}
\usepackage{siunitx}
\usepackage{nicefrac}
\usepackage{tocloft}
\usepackage{listings}
\usepackage[raster,skins]{tcolorbox} %
\usepackage{xltabular}
\usepackage{adjustbox}
\usepackage{xurl}
\usepackage{pifont}
\newcommand{\featok}{\textcolor{green!70!black}{\ding{52}}}
\newcommand{\featno}{\textcolor{red!70!black}{\ding{55}}}

\definecolor{lightgray}{rgb}{0.9,0.9,0.9}
\definecolor{bestcell}{HTML}{FFF4CC}
\definecolor{safetygroup}{HTML}{B9A8C2}
\definecolor{generalgroup}{HTML}{E3D8D9}
\definecolor{reasongroup}{HTML}{B0BBCB}
\definecolor{featuresynthbase}{HTML}{EB687B}   % Security-Feature-Driven Synthetic Data
\definecolor{randomsynthbase}{HTML}{F1837B}    % Security-Related-Randomly Synthetic Data
\colorlet{featuresynthrow}{featuresynthbase!30}
\colorlet{randomsynthrow}{randomsynthbase!15}
\title{E-Commerce Bench: Evaluating LLM Agents on Long-Horizon Autonomous Business Operation}

\author{
\bf Wei Fan$^{1,3}$\thanks{Equal contribution. Work done during summer internship.}, Xinjie Shen$^{1}$\footnotemark[1],
Xudong Guo$^{1}$, Jianhong Tu$^{1}$, Yang Su$^{1}$, Yinger Zhang$^{1}$,\\
\bf Lianghao Deng$^{1}$, Fengyu Wang$^{2}$, Baohua Dong$^{2}$, Yangqiu Song$^{3}$, Dayiheng Liu$^{1}$\\
\mbox{$^{1}$Qwen Team, Alibaba Group} \quad \mbox{$^{2}$Taobao \& Tmall Group, Alibaba Group}\\
\mbox{$^{3}$Department of Computer Science and Engineering, HKUST}
}

\begin{document}

\renewcommand{\thefootnote}{\fnsymbol{footnote}}
\maketitle
\renewcommand{\thefootnote}{\arabic{footnote}}\setcounter{footnote}{0}

\begin{abstract}

Long-horizon agentic tasks go beyond chaining short tasks over more interaction turns. Their evolving dynamic environments and long-range dependencies require Large Language Models~(LLMs) to continually explore, learn from experience, and adapt their policies over thousands of steps.
We introduce \textbf{E-Commerce Bench}, the first open-source benchmark that integrates multi-round counterpart negotiation and dynamic events into a year-long business operation.
Over a 365-day year, an LLM agent concurrently runs multiple online stores, researching the market, negotiating with suppliers to source inventory, optimizing sales strategies, fulfilling orders, handling returns, and managing cash flow to maximize its end-of-year total assets.
To construct a realistic merchant-side operating environment, the product and supplier data are derived from a real e-commerce platform, while a year-long calendar of promotions, natural disasters, and supply-chain shocks continually reshapes demand.
For reproducibility, both sides of the market are deterministic: customer purchases and returns follow a fixed demand model, while a negotiation kernel determines supplier pricing, concessions, and decisions, with an LLM used only to verbalize them.
We evaluate 18 frontier models across seven dimensions, including year-end assets, and find that no single model dominates. GPT-5.6 Sol earns the most, growing the \textyen 100{,}000 opening stake into \textyen 1{,}431{,}425, yet it ranks 16th of 18 on fraud avoidance and trails Fable5 in operational efficiency.
Among open-weight models, Qwen3.8-Max-Preview leads with \textyen 416{,}252, 38\% above GLM 5.2 (high), and achieves the strongest learning over the horizon, progressively bargaining down prices across repeated orders.
Our code is available at \url{https://github.com/QwenLM/E-CommerceBench}.

\end{abstract}

\clearpage

\begin{figure}[t]
  \centering
  \includegraphics[width=\linewidth]{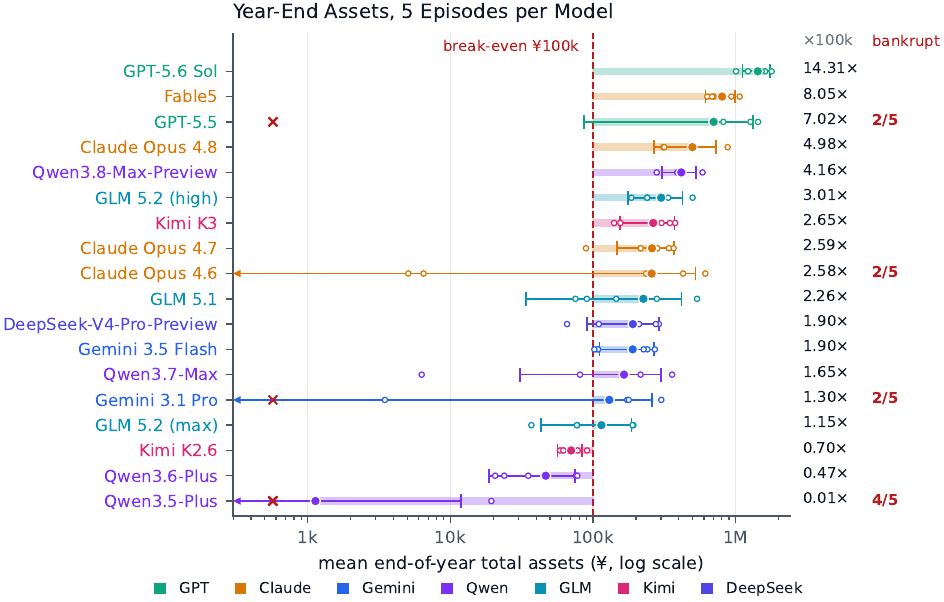}
  \caption{Mean end-of-year total assets, eighteen models, five 365-day episodes each. Bars start at the \textyen 100{,}000 stake. Whiskers span one 5-run standard deviation, hollow dots are single episodes, and red fractions count bankrupt episodes, which stay pooled into the mean. The right column gives mean assets as a multiple of the stake.}
  \label{fig:assets}
\end{figure}

\begin{figure}[t]
  \centering
  \includegraphics[width=\linewidth]{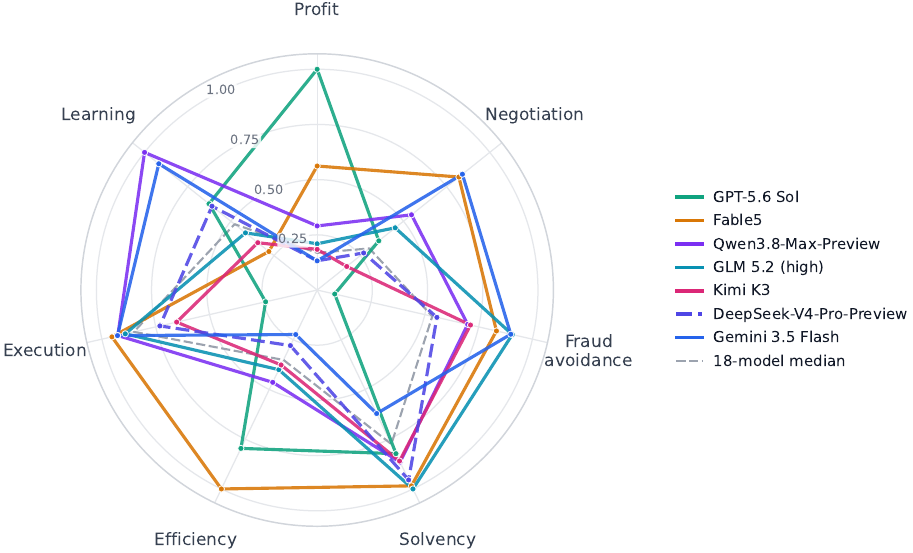}
  \caption{Seven evaluation axes for the strongest model of each vendor family by mean total assets, the primary score, and the six evaluation dimensions clockwise from the top, each on the ranking metric \S\ref{sec:eval} gives it. Axes are min-max normalized over the eighteen model means, so 1 is the best of the eighteen and 0 the worst, with a dashed polygon at the median. Fraud avoidance, solvency, execution, and learning are sign-flipped, so a vertex further from the center is better on every axis.}
  \label{fig:ranks}
\end{figure}

\clearpage

% main text lists every heading level, the appendix only its top level, which
% frag_app_00_head.tex sets before the appendix content is written to the .toc
\setcounter{tocdepth}{3}
\tableofcontents

\clearpage

\input{sections/1_introduction}

\clearpage
\input{sections/2_related_work}

\clearpage
\input{sections/3_ec_bench}

\clearpage
\input{sections/4_experiment}

\input{sections/5_conclusion}

\clearpage

\bibliography{biblio,xinjie}
\bibliographystyle{colm2024_conference}

\clearpage
\input{sections/appendix}

\end{document}

%% file: sections/1_introduction.tex
\section{Introduction}\label{sec:intro}

Large language model (LLM) agents are increasingly expected to handle tasks that span thousands of steps, planning and acting over long trajectories with external tools~\citep{yao2023react}.
Such tasks require an agent to make sequential decisions, adapt to past experiences, and remain coherent long after the initial instruction. Sustained competence of this kind marks a frontier for current models and eludes most agentic benchmarks.
In reinforcement learning, tasks are divided into \emph{episodic} and \emph{continuing} ones \citep{sutton2018reinforcement}, and this distinction also organizes the long-horizon agentic tasks.
Episodic benchmarks set a goal and score its completion, whether resolving a GitHub issue~\citep{jimenez2024swebench}, operating a real computer~\citep{xie2024osworld}, or producing knowledge work that an expert grades~\citep{patwardhan2025gdpval}.
Continuing tasks carry no terminal state, running without end in dynamic environments~\citep{qian2025trade, chen2026stockbench} and scoring an accumulated outcome that surfaces the failures long horizons produce, among them loss of coherence and an inability to turn experience into better decisions.
Business operations provide a natural continuing scenario with high-stakes, irreversible decisions, where compounding financial health tests sustained agent competence. Despite its promise, current benchmarks capture this environment only partially.
While the Vending-Bench series~\citep{backlund2025vending, andonlabs2026vending2} simulates multi-round supplier negotiations, its reliance on mock data and prompt-based simulation yields highly stochastic supplier feedback that is vulnerable to manipulation.
YC-Bench~\citep{he2026ycbench} runs a deterministic startup year yet involves no supplier sourcing. Data-grounded retail benchmarks such as MerchantBench~\citep{shi2026merchantbench} and RetailBench~\citep{zhang2026retailbench} add real products and open code while dropping negotiation and adversarial dynamics altogether.
Evidently, existing benchmarks either deviate from real-world market dynamics or rely on purely LLM-driven counterparts for multi-round negotiations, which introduces substantial sampling noise and ultimately compromises both the credibility and reproducibility of the evaluation.

To address these, we introduce \textbf{E-Commerce Bench}, the first open-source benchmark that unifies deterministic supplier negotiation and customer sales modeling within a real-world continuing business operation (Figure~\ref{fig:overview}).
Across a simulated 365-day year, an agent can run up to four online stores concurrently, researching the market, negotiating for cost-effective goods, formulating sales strategies, and managing cash flow to maximize end-of-year total assets.
Grounded in \textbf{Taobao \& Tmall data}, the environment encompasses a total of 6{,}886 products and 576 suppliers calibrated from empirical platform logs, while a year-long calendar of promotional events, natural disasters, and supply-chain shocks continually shifts dynamic market demand.
Critical market factors, such as baseline product costs and exact consumer demand, remain hidden, requiring the agent to continuously search, negotiate, and model market dynamics through a dedicated e-commerce toolset provided by the benchmark.
Meanwhile, the benchmark harness automatically manages the context window and offers a persistent memory module~\citep{packer2024memgpt, park2023generative} to sustain a year-long operational horizon.
To ensure rigorous reproducibility, determinism governs both sides of the market. Customer purchases and returns follow a fixed demand model that combines price elasticity, weekday, seasonality, scheduled promotions, market events, and store reputation, with its only randomness seeded from the episode so runs reproduce exactly. On the supply side, a deterministic negotiation kernel fixes every pricing, concession, and accept/reject decision through explicit bargaining policies, while an LLM only renders them as dialogue. Roughly a quarter of the suppliers turn adversarially fraudulent, on kernels that do not vary between runs. No sampling noise reaches the customers or the counterpart's pricing and accept/reject decisions, so a difference in outcome entirely belongs to the agent policy rather than to chance. Only the LLM renderer's wording varies between runs~(\S\ref{sec:nego_twolayer}).

We evaluate 18 leading frontier and open-weight models along multi-dimensional metrics spanning negotiation quality, fraud avoidance, cash flow and solvency, operational efficiency, operations execution, and learning over the horizon, rather than by total assets alone. A radar over the primary score and those six dimensions comes out visibly uneven, and six of the seven models it draws, one per vendor family, fall below the eighteen-model median on at least one axis (Figure~\ref{fig:ranks}). GPT-5.6 Sol grows its capital roughly 14-fold, yet ranks 16th of 18 on fraud avoidance and earns less from a tool call than Fable5 does, while Claude Opus 4.7 leads both bargaining and fraud avoidance from mid-pack on profit. Among the ten open-weight entries, first place goes to our own Qwen3.8-Max-Preview at 4.2 times the initial balance, and four of the eighteen models slip into bankruptcy on part of their runs. Qwen3.8-Max-Preview is the strongest model in the field, improving over the year. Across 8{,}647 repeat purchases of the same item from the same honest supplier, 16 models show no clear sign of pushing the purchase price down as the year goes on, exposing critical deficiencies in their long-horizon experiential learning.

We make three principal contributions.
\begin{enumerate}
\item We present \textbf{E-Commerce Bench}, a realistic business environment grounded in Taobao \& Tmall platform data, and the first open-source benchmark to combine deterministic counterpart negotiations, dynamic events, and concurrent multi-store management over a year-long horizon.
\item We keep the economy \textbf{deterministic on both sides of the market}, with a fixed demand model driving customer purchases and returns and a deterministic negotiation kernel setting every pricing, concession, and accept/reject decision while an LLM only renders dialogue. The kernel extends the negotiation-partner design of TERMS-Bench~\citep{zhang2026terms} to repeated, cross-supplier bargaining.
\item We evaluate 18 models along \textbf{multi-dimensional metrics}, so one total-assets figure can no longer hide where a model is weak. Reading six dimensions beside that total turns a leaderboard into a capability profile, and the profiles come out sharply divergent, the best earner being neither the best negotiator nor the efficient operator.
\end{enumerate}

The remainder of this paper is organized as follows. Section~\ref{sec:related} reviews long-horizon agentic benchmarks and positions E-Commerce Bench among them. Section~\ref{sec:bench} details the benchmark, covering its four-layer architecture, the deterministic negotiation kernel, the real-data economy, and the evaluation metrics. Section~\ref{sec:experiment} reports results for the 18 models on every measured dimension, together with an error analysis of recurring failure modes. Section~\ref{sec:conclusion} concludes.

%% file: sections/2_related_work.tex
\section{Related Work}\label{sec:related}

\subsection{Episodic Long-Horizon Agentic Tasks}

Long-horizon agentic benchmarks are divided by whether a run carries a terminal state~\citep{sutton2018reinforcement}. Episodic benchmarks define one, setting a goal and rewarding the multi-step execution that reaches it. \emph{Work-value} benchmarks grade an economically meaningful deliverable, GDPval~\citep{patwardhan2025gdpval} scoring knowledge work against expert judgment and TheAgentCompany~\citep{xu2025agentcompany} setting a simulated employee's tasks inside a software firm. \emph{Environment-grounded} benchmarks stretch the horizon through real software, SWE-bench~\citep{jimenez2024swebench} resolving a GitHub issue against its own tests and OSWorld~\citep{xie2024osworld} driving a real operating system over many GUI steps. \emph{Exploration-driven} benchmarks probe reasoning further still, UltraHorizon~\citep{luo2025ultrahorizon} hiding the rules of a partially observable world and EdgeBench~\citep{zhu2026edgebench} learning from one environment for 12 to 72 hours at a time.
However long these tasks run, each ends the moment its goal is met, reducing evaluation solely to the quality of the final deliverable. Consequently, this setup fails to assess an agent's ability in dynamic environments to track dependencies across hundreds of sub-tasks, recover from unexpected failures, or leverage accumulated history to optimize future actions.

\subsection{Continuing Long-Horizon Agentic Tasks}

Continuing tasks carry no terminal state, so the agent acts indefinitely and earns an accumulated score. The shape recurs across trading a live portfolio for cumulative return~\citep{qian2025trade, chen2026stockbench} and self-evolving agents that grow a skill library~\citep{wang2023voyager} or rewrite their own code~\citep{zhang2026dgm}. The setting E-Commerce Bench belongs to is business-operation simulation, where the agent runs a going concern and is judged by its accumulated balance.
Vending-Bench~\citep{backlund2025vending} established the line, running one vending machine over roughly two thousand messages and exposing failures of \emph{long-term coherence}, with LLM-generated supplier replies that leave runs stochastic. Vending-Bench~2~\citep{andonlabs2026vending2} extends it to a full year and adds adversarial suppliers, still LLM-driven. YC-Bench~\citep{he2026ycbench} runs a deterministic startup year against rule-based clients but no supplier negotiation. Two benchmarks are grounded in real data, MerchantBench~\citep{shi2026merchantbench} over 365 days of seller-side operation with suppliers arriving as stochastic events, and RetailBench~\citep{zhang2026retailbench} over 180 supermarket days against a privileged oracle. Neither offers negotiation or adversarial suppliers, and RetailBench lists strategic supplier behavior and multi-store coordination among the things it leaves out.
Table~\ref{tab:related_work} contrasts the five. The Vending-Bench line couples negotiation with adversarial suppliers in its sequel, but only through sampled LLM parties and without an open, real-data economy, while the data-grounded retail pair buys reproducibility by giving those dynamics up. E-Commerce Bench holds all six properties at once, and no other entry pairs a negotiating counterpart with reproducible supplier economics. TERMS-Bench~\citep{zhang2026terms} sits outside the business-operation line, diagnosing bilateral price bargaining against a rule-based counterpart over sampled regimes rather than data-grounded cost floors, so it supplies a scripted opponent but no economy for the resulting deals to act on. Our market is deterministic on both sides, a fixed demand model driving customers while a \emph{deterministic negotiation kernel} adapts that counterpart design to set every pricing and accept/reject decision, leaving an LLM only the dialogue rendering.

\begin{table}[t]
\centering
\caption{\small Comparison with continuing business-operation benchmarks. \featok\ present, \featno\ absent. \emph{Steps/run} is the per-episode agent-action count, taken or inferred from each paper. \emph{Op.\ units} counts business units~(e.g., stores) run concurrently. \emph{Reproducible} marks counterparts that follow fixed rules or seeded draws rather than a sampled LLM. $^{\ddagger}$In Vending-Bench 1/2, product and supplier names are real while prices, demand, and elasticity are LLM-generated. $^{\S}$Supplier economics are deterministic, while the renderer voicing them varies wording between runs (\S\ref{sec:nego_twolayer}).}
\label{tab:related_work}
\footnotesize
\setlength{\tabcolsep}{4.5pt}
\renewcommand{\arraystretch}{1.2}
\begin{adjustbox}{max width=\textwidth}
\begin{tabular}{l l c c c c c c c c}
\toprule
\textbf{Benchmark} & \textbf{Domain} & \makecell{\textbf{Steps/}\\\textbf{run}} & \makecell{\textbf{Op.}\\\textbf{units}} & \makecell{\textbf{Real}\\\textbf{data}} & \makecell{\textbf{Nego-}\\\textbf{tiation}} & \makecell{\textbf{Multi-}\\\textbf{agent}} & \makecell{\textbf{Adver-}\\\textbf{sarial}} & \makecell{\textbf{Repro-}\\\textbf{ducible}} & \makecell{\textbf{Open}\\\textbf{source}} \\
\midrule
Vending-Bench~\citep{backlund2025vending}     & Vending     & $\leq$2000       & 1 & \mbox{\featno$^{\ddagger}$} & \featok & \featok & \featno & \featno & \featno \\
Vending-Bench 2~\citep{andonlabs2026vending2} & Vending     & 3000--6000       & 1 & \mbox{\featno$^{\ddagger}$} & \featok & \featok & \featok & \featno & \featno \\
YC-Bench~\citep{he2026ycbench}                & Startup     & $\sim$500        & 1 & \featno   & \featno & \featok & \featok & \featok & \featok \\
MerchantBench~\citep{shi2026merchantbench}    & E-commerce  & $\sim$700        & 1 & \featok   & \featno & \featno & \featno & \featok & \featok \\
RetailBench~\citep{zhang2026retailbench}      & Supermarket & $\sim$1000       & 1 & \featok   & \featno & \featno & \featno & \featok & \featok \\
\midrule
\rowcolor{lightgray}
\textbf{E-Commerce Bench (Ours)}              & E-commerce  & 1000--4000       & \textbf{$\leq$4} & \featok & \featok & \featok & \featok & \mbox{\featok$^{\S}$} & \featok \\
\bottomrule
\end{tabular}
\end{adjustbox}

\end{table}

%% file: sections/3_ec_bench.tex
\section{E-Commerce Bench}\label{sec:bench}
\subsection{Overview}\label{sec:overview}

E-Commerce Bench hands the agent a merchant account, \textyen 100{,}000, and the 2026
calendar year on a simulated Chinese online marketplace. The job is to end the year with as much money as
possible.
The agent researches
which store types and categories carry demand, bargains suppliers down, buys inventory and
publishes it to one of at most four stores. From there it prices the goods, reads yesterday's sales, ships orders, absorbs the returns, and withdraws settled revenue before the next operating charge lands.
To evaluate the strategies driving the year-end profit, we independently score six core capabilities: negotiation quality, fraud avoidance, cash flow and solvency, operational efficiency, operational execution, and long-horizon learning (\S\ref{sec:eval}).
Next, we introduce the four layers of E-Commerce Bench in detail.

\begin{figure}[t]
  \centering
  \includegraphics[width=0.95\linewidth]{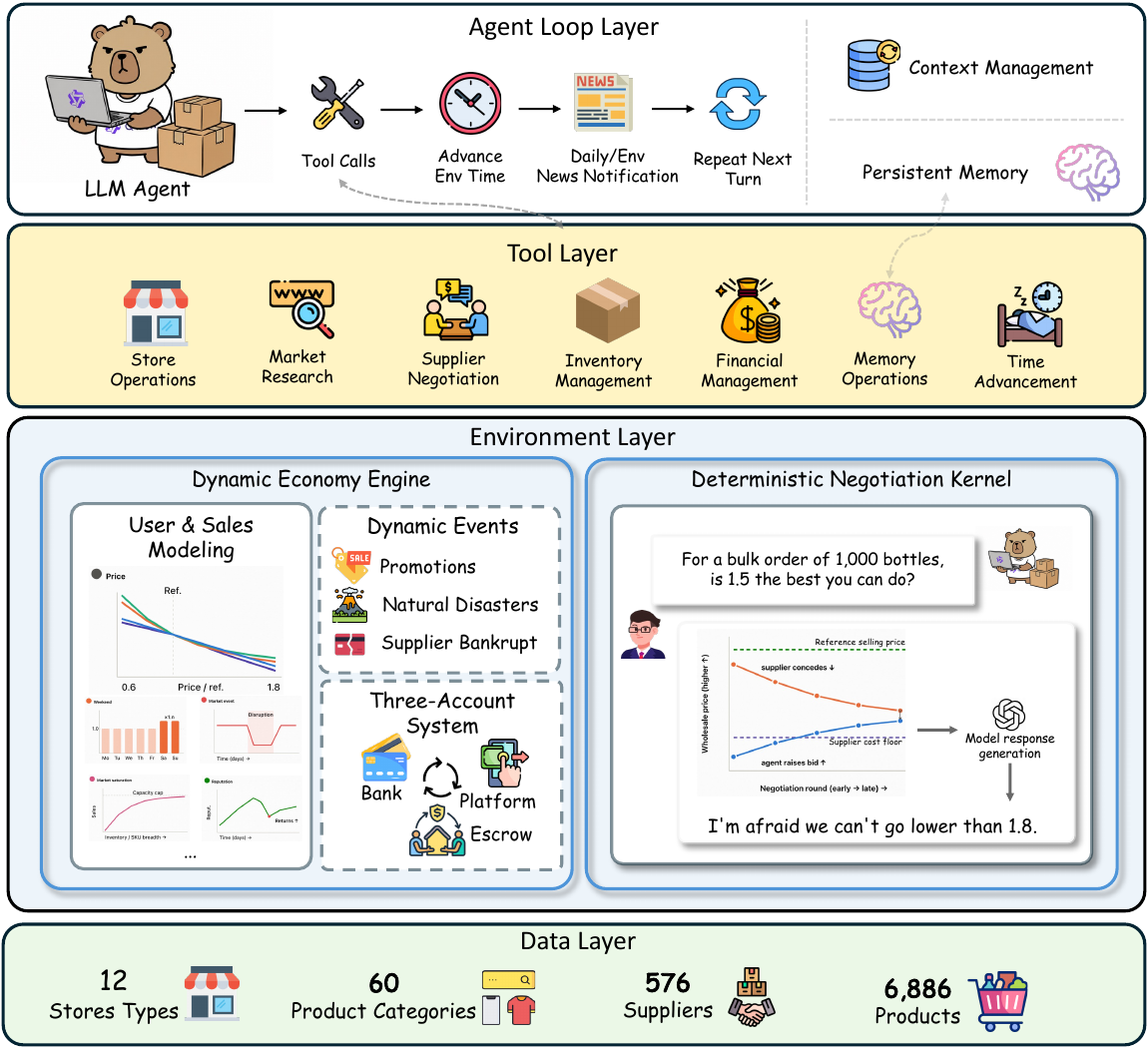}
  \caption{Overview of E-Commerce Bench. The agent works through four layers. An \emph{agent loop layer} carries turn-based control, context management, and persistent memory, and a \emph{tool layer} holds the 18 e-commerce tools. Under both sits a deterministic \emph{environment layer}, the sales-and-economy engine beside the two-layer negotiation engine, computing over a \emph{data layer} derived from a real e-commerce platform.}
  \label{fig:overview}
\end{figure}
\subsection{Agent Loop Layer}
\label{sec:agent_loop}

\begin{figure}[t]
  \centering
  \includegraphics[width=\linewidth]{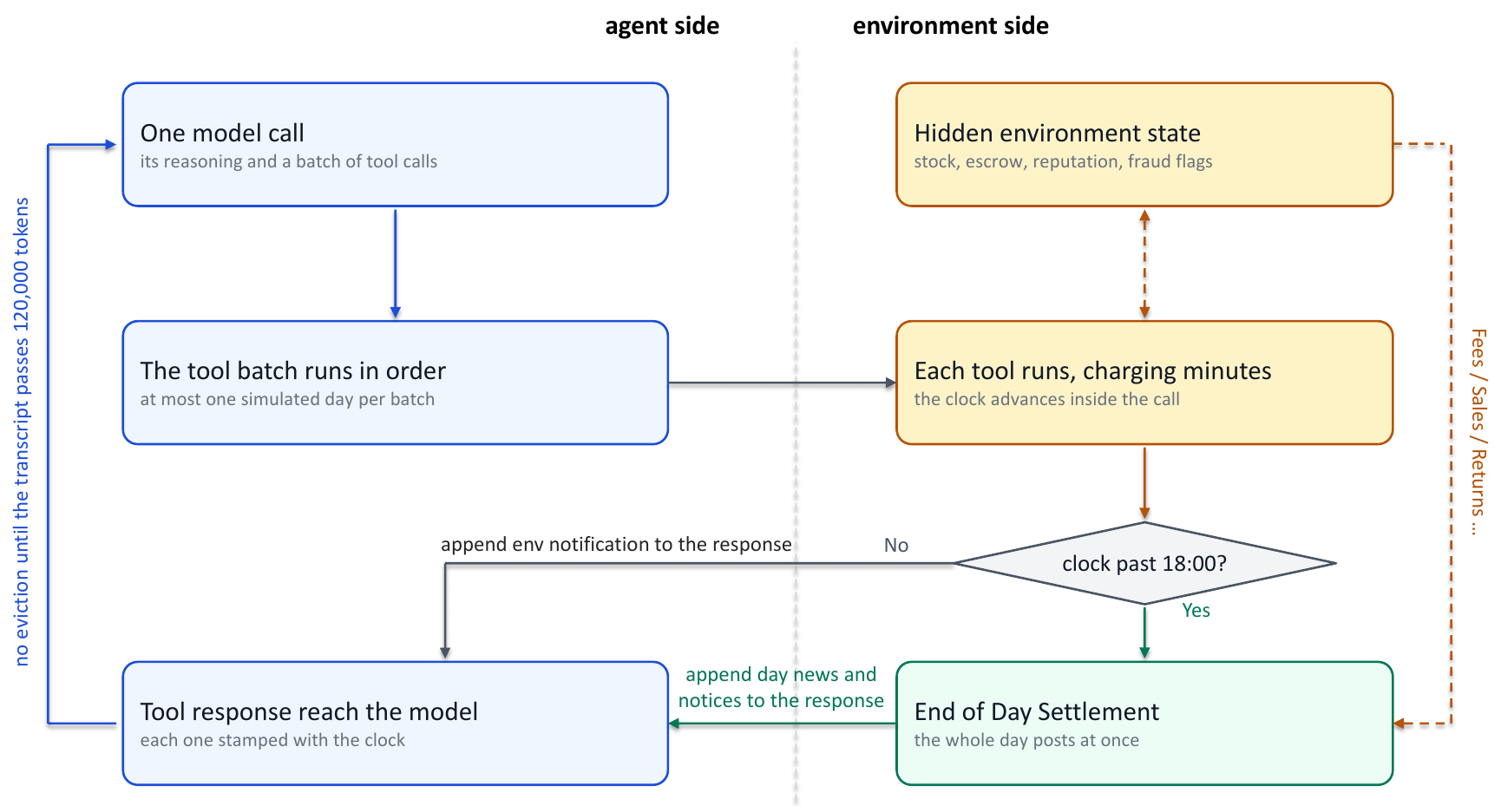}
  \caption{The agent loop layer. One model call produces a batch of tool calls; the batch
  runs in order, and a timed tool advances the simulated clock inside its own call. The
  daily settlement fires whenever the clock reaches the next 08:00, whether a timed tool
  carried it there or the agent asked to skip to tomorrow.
  Tools read and write the hidden environment state, and the
  settlement writes back to it as well, posting that day's fees, sales, and returns. The
  day's news and any notices are appended to the tool responses. The transcript is left
  alone until it passes 120{,}000 tokens, and only then are its oldest groups dropped.}
  \label{fig:agent_loop}
\end{figure}

Three mechanisms sit between the model and the environment. The turn loop~(Figure \ref{fig:agent_loop}) bounds what one
model call can do, and charges simulated minutes for every tool it uses. Every turn adds its
tool traffic to the message list, and a year of calls overruns the window many times, so the
context editor evicts the oldest turns first.
The window is not the real limit either,
since long contexts degrade well before they overflow \citep{liu2024lost, bai2024longbench}.
Because evicted text leaves nothing behind, the agent also gets a persistent memory system beyond the
reach of any eviction pass.

\subsubsection{Turn-Based Loop and Simulated Time}
\label{sec:turn_loop}

Within a single turn, the model can request several tools, which then run one by one in the
order given. Turns therefore range from a single balance check to a batch of a dozen
tools. An episode runs to a cap of 4{,}000 turns (\S\ref{sec:setup}), or ends early once
the model goes three turns without calling a tool.
Every tool but the day jump costs simulated minutes: 10 for a balance check and 60 for opening a store
(Table~\ref{tab:tools} in Appendix~\ref{app:harness}), spent out of a 600-minute day that runs from 08:00 to
18:00. Looking something up draws on the same clock as acting on it. A message costs 30
minutes whether it goes to one supplier or to ten, so twenty of them fill a day.

Once the clock passes 18:00, it jumps to 08:00 the next morning, which
\texttt{wait\_for\_next\_day} also does on request, costing no minutes but forfeiting the
rest of the day. Every jump runs the same fixed sequence, charging operating and storage
costs, turning the previous day's demand into orders, cancelling shipments that missed
their two-day deadline, settling matured escrow, delivering arriving purchase orders, and
updating reputation (\S\ref{sec:economy}).
What the agent gets back after advancing to the next day is four thin signals: the day's news feed, a reminder
once the bank balance no longer covers the day's operating cost, the current time, and a
warning about token usage (Appendix~\ref{app:harness}). Anything more precise costs a tool
call and the minutes that come with it.

\subsubsection{Context Management}
\label{sec:context_mgmt}

\begin{figure}[t]
  \centering
  \includegraphics[width=\linewidth]{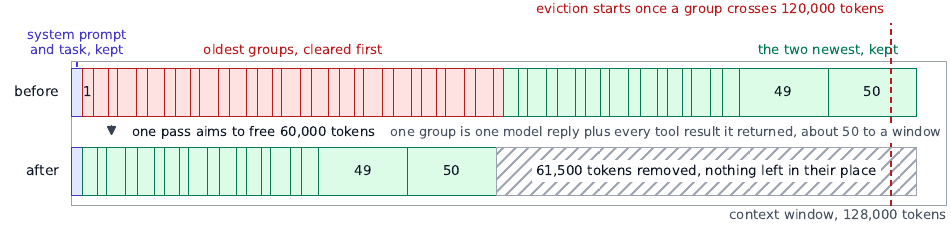}
  \caption{How eviction works. A group is what the editor removes as a unit: one model
  reply together with every tool result that reply produced, so a turn that called six
  tools is still one group, and about fifty groups fill the window. A pass fires only once
  a group pushes the count past 120{,}000 tokens, then drops whole groups, oldest first,
  until 60{,}000 tokens are released or the clearable groups run out. The system message,
  the first user turn and the two newest groups are kept.}
  \label{fig:context_mgmt}
\end{figure}

A single tokenizer counts each model's context to ensure all of them
work against the same 128{,}000-token budget.
A context eviction is triggered whenever the count reaches 120,000 tokens, taking 60,000 as its release target.
Whole tool-call groups go, an assistant message and every tool response after it, oldest first (Figure~\ref{fig:context_mgmt}).
The system message, the first user turn, and the two newest groups are spared.
The agent is told the policy verbatim, thresholds included, so preserving what matters in its memory is a choice it can make. 
Across the 90 evaluation episodes of
\S\ref{sec:experiment}, 1{,}495 evictions ran.
How much history a model loses scales with its total tool traffic, the turns it takes
times what each turn drags in, and the two factors trade off. Claude Opus 4.7 finishes the
year in 432 turns and has about four windows' worth of transcript discarded along the way,
while Gemini 3.5 Flash takes 2{,}628 turns and loses nearly nineteen windows' worth. Turns
alone do not settle it. Kimi K2.6 takes 1{,}017 turns and loses the same three and a half
windows as Claude Opus 4.7 at 432 turns, because it drags in $431$ tokens a turn against
Opus 4.7's $1{,}058$, and GPT-5.5 loses 28 windows in half Gemini 3.5 Flash's turns at
$2{,}759$ a turn.

\subsubsection{Persistent Agentic Memory}
\label{sec:memory}

% source: tools/operate_memory.py:5,12,21-62,81 (MAX_MEMO_COUNT=20, advance_minutes(10))
The memory store is a list of notes the agent keeps in the environment rather than in its own messages, so eviction never touches it. The store holds at most 20 entries, each a title and its content. The agent fully curates the store itself—creating, updating, and pruning entries as needed \citep{shinn2023reflexion, packer2024memgpt, zhong2024memorybank, sumers2024coala}. Reading the store back is cheap in minutes yet expensive in tokens, since returned entries re-enter the context and become evictable like everything else. A capable agent is thus incentivized to preserve high-value strategic knowledge, such as past settlement prices or identifying deceptive suppliers.
% Section~\ref{sec:nego_repeated} derives that bound and \S\ref{sec:nego_metrics} measures the drift away from it, so memory is scored rather than merely convenient.

\subsection{Tool Layer}
\label{sec:tools}

Eighteen tools cover a merchant's back office, in the seven groups Figure~\ref{fig:overview}
draws, and Table~\ref{tab:tools} gives each one its minute cost and its effect. 
Every call but the day jump is billed in simulated minutes (\S\ref{sec:turn_loop}), which makes the tool list a budget rather than a menu.
Checking the bank balance, the warehouse, and one store's status costs 10 minutes each. A morning round of all three spends 30 minutes a day, more than 18 working days across the year.
Only \texttt{chatbox} can create a purchase order, so every unit of inventory arrives
through free-form text carrying fenced \texttt{negotiate} blocks
(\S\ref{sec:nego_twolayer}). Routine mail, supplier offers, and fraudulent pitches
(\S\ref{sec:nego_fraud}) all cost the same 30 minutes through that one tool.

\subsection{Sales and Economy Engine}\label{sec:economy}

The sales and economy engine decides how much the agent sells and when it gets paid. Nothing about the market is sampled. A fixed formula computes how many units customers buy each day, and money moves between accounts on fixed delays. The simulator still draws a few random numbers, for
instance when it turns a fractional unit of demand into a whole one, and those come from one seed shared by every run, so two episodes that take the same actions end the year with exactly the same money.

\subsubsection{Multi-Store Setup}\label{sec:stores}

An agent trades through stores it opens itself. Twelve store types are on offer, each
covering a different set of product categories, and at most four may trade at once, one per type (Appendix~\ref{app:data_stores}). Opening one costs \textyen 500 and then
\textyen 60 to \textyen 130 a day to run, the harder types costing more. Closing a store stops the daily charge, and stock left behind keeps accruing storage unless the agent liquidates it for $10\%$ of purchase cost. Reopening later starts reputation over. A market lookup grades a category's margin low, moderate, or high, while the price a supplier will actually accept emerges only from bargaining (\S\ref{sec:nego_engine}).

\subsubsection{Three-Account Deferred Settlement}\label{sec:econ_settlement}

\begin{figure}[t]
  \centering
  \includegraphics[width=\linewidth]{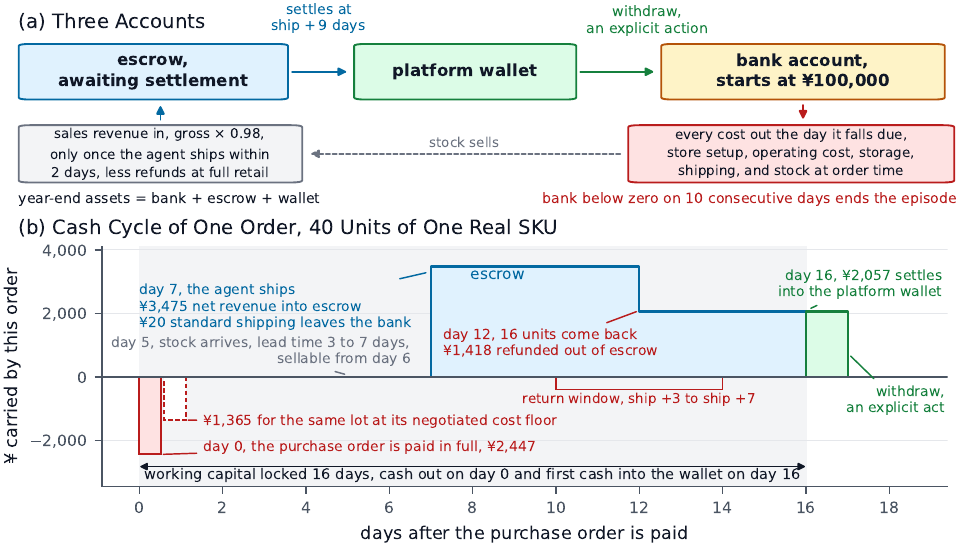}
  \caption{Three-account deferred settlement. (a) Costs leave the bank the day they are incurred, while revenue passes through escrow and the platform wallet first. (b) One 40-unit order bought at the opening quote and resold at reference, followed as cash for 16 days.}
  \label{fig:settlement}
\end{figure}

Money an agent has earned is not money it can spend. Cash sits in three places
(Figure~\ref{fig:settlement}). The \emph{bank account} pays every cost the moment it is incurred, a purchase order in full the moment a negotiation closes. A sale pays nothing at first and only creates a pending shipment. Once the agent ships it, the revenue enters \emph{escrow} with the platform's commission already deducted. Nine days later that balance matures into the \emph{platform wallet}, and only an explicit withdrawal moves it back to the bank account where costs are paid. A refund is taken out of the escrow batch its sale belongs to, so a return never touches spendable cash.
One order shows the gap. A 40-unit lot of a mid-priced accessory costs \textyen 2{,}447 the day it is placed and returns its first cash on day 16. Ten consecutive days closing with a negative bank balance end the episode as bankrupt
(Appendix~\ref{app:econ_constants}), so an agent selling at a profit while leaving its
revenue in the wallet can still fail (\S\ref{sec:experiment}).

\subsubsection{Deterministic Multi-Factor Demand Model}\label{sec:demand}

\begin{figure}[t]
  \centering
  \includegraphics[width=\linewidth]{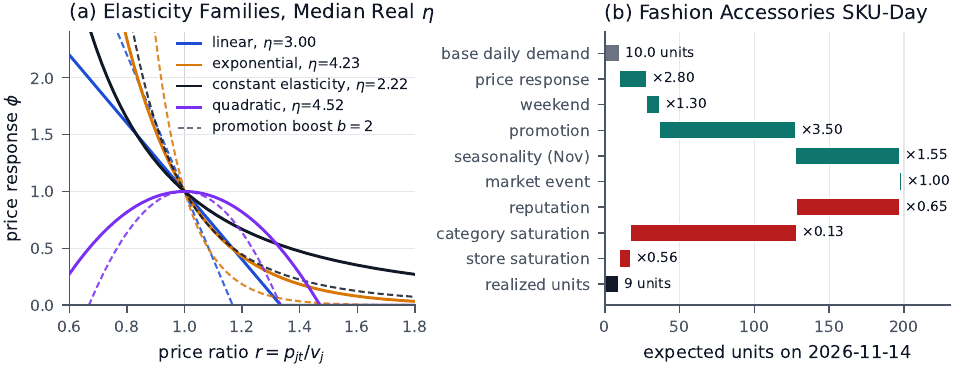}
  \caption{The demand model. (a) The four price-elasticity families at their median parameter, solid without a promotion and dashed with the largest promotion boost. (b) The chain of multipliers that turns one SKU's price into units sold.}
  \label{fig:demand_model}
\end{figure}

How many units are sold is a product of named factors rather than a random draw. Demand for one SKU on one day starts from what its category sells in a store of that type. Six multipliers then act on that base: the response to the posted price, a weekend uplift, the promotion the agent joined, the month's seasonality, whatever the calendar holds that day, and the store's reputation. Two capacity terms trim the result, one for the category and one for the store, each a ceiling on how much demand that category or that store can absorb in a day. What survives is capped at the units actually on the shelf.
\begin{equation}
\small
\label{eq:demand}
\begin{aligned}
\tilde{d}_{ijt} &= \beta_{c\tau}\;\phi_{f_c}\!\left(r_{ijt};\,\eta_c\,b_{it}\right)\,w_t\;
  m_{it}(\delta_{it})\;\sigma_{\tau,t}\;\varepsilon_{\tau,t}\;\rho_{it},
  \qquad r_{ijt} = \frac{(1-\delta_{it})\,p_{ijt}}{v_j},\\[2pt]
\gamma_{ijt} &= \begin{cases}
  \dfrac{\kappa_{c}}{\kappa_{c} + \sum_{j' \in c} \tilde{d}_{ij't}} & \text{if } n_{\tau} \ge 2,\\[4pt]
  1 & \text{if } n_{\tau} = 1,
\end{cases}\\[4pt]
d_{ijt} &= \tilde{d}_{ijt}\;\gamma_{ijt}\;
  \frac{\kappa_{\tau}}{\kappa_{\tau} + \hat{D}_{it}},\\[2pt]
q_{ijt} &= \min\!\left(x_{ijt},\;\lfloor d_{ijt} \rfloor +
  \mathbf{1}\!\left[u_{jt} < d_{ijt} - \lfloor d_{ijt} \rfloor\right]\right).
\end{aligned}
\end{equation}
Indices run over stores $i$, SKUs $j$, dates $t$, store types $\tau$ and categories $c$, and Appendix~\ref{app:econ_demand} defines every symbol alongside the seasonality tables of Appendix~\ref{app:data}. Price enters through one of four elasticity families carried per category \citep{gallego1994pricing}. One of them, the quadratic family, is symmetric about the public reference price rather than monotone, so in the categories that carry it, a discount below reference loses demand instead of gaining it. The two capacity terms can then dominate everything the six multipliers did. For one accessory SKU on a promotion Saturday, they cut a compounded expectation of $128.2$ units to
the $9$ actually sold, the category term alone taking a factor of $7.5$
(Figure~\ref{fig:demand_model}).
How much they bite depends on how heavily the agent stocks that category and store, meaning that adding more homogenized SKUs cannibalizes demand rather than boosting aggregate sales.

\subsubsection{Fulfillment, Shipping Speed, and Returns}

Every sale has to be shipped, and how it ships is a priced choice. Faster delivery costs more freight and brings back fewer returns, and a bulkier parcel costs more again. An order left unshipped past its two-day deadline is canceled outright, the units going back to the warehouse unsold and the cancellation counting against reputation. Returns are decided when goods ship and arrive days later, so a fulfillment decision costs money long after it is taken \citep{petersen2009returns}. Four channels set a unit's return probability, its category's natural rate, whether the supplier shipped defective stock, how far above reference the agent priced it, and the shipping speed chosen (Appendix~\ref{app:econ_returns}). A defective unit carries a return rate of at least $0.40$
and at least double the natural rate, capped at $0.95$. Pricing at $1.3$ times reference multiplies the rate by $1.50$, and slow shipping by a further $1.30$. Every arriving return refunds the customer out of escrow and puts the unit back in the warehouse on a fresh storage clock. Two of the four channels answer to the
agent, its pricing and its freight tier, so a model's return rate reads partly as a
management outcome rather than only as a property of the catalog (\S\ref{sec:eval}).

\subsubsection{Reputation System}

A store's reputation multiplies its demand directly (\S\ref{sec:demand}), so service
failures cost sales. Credit accrues when an order ships rather than when it
sells, so an unshipped order earns nothing and draws the cancellation penalty as well.
Returns and cancellations subtract from the same standing, weighted $0.6$ and $1.0$
\citep{resnick2000reputation}, so a missed deadline costs more than a change of mind. A store pinned at the floor receives $15\%$ of the demand it would receive at the top, and the capacity terms return part of that loss once they bind (Appendix~\ref{app:econ_reputation}).

\subsubsection{Dynamic Event System}

The calendar moves demand whether or not the agent is ready. Market events arrive on fixed dates, scaling demand per store type and stretching delivery lead times
(Appendix~\ref{app:data_calendar}, Table~\ref{tab:events} and Figure~\ref{fig:calendar}). Direction is rarely uniform, so a storm that lifts food cuts fashion and footwear. News of an event reaches the agent's feed as the event begins, while a promotion is announced a week ahead without its dates or its size. Joining reveals the size, and the year's first sale is already running when its announcement lands. The boost from joining peaks at a $30\%$ discount, so a deeper cut adds demand only through the price response while margin falls on every unit. Suppliers exit on a count rather than a date, each retiring after a hidden number of filled orders, so a counterpart whose floor the agent has learned can be gone by the next re-order.

\subsubsection{Daily Settlement and Processing Lifecycle}

Costs are charged before any revenue is recognized, so every morning tests
solvency at the day's low point. Sales are then computed for the \emph{previous} calendar date against the inventory and prices the agent left standing overnight, so the agent steers a system that lags it by a day. Refunds leave escrow before matured escrow settles, so a return can empty the batch that was about to pay out. Appendix~\ref{app:econ_tick} lists all thirteen steps with the state each one touches.

\subsection{Deterministic Negotiation Engine}
\label{sec:nego_engine}

The negotiation engine sets what the agent pays for stock. Every unit it sells has to be bought from a named supplier, one chatbox message at a time, at a price the two of them argue over. Roughly a quarter of the roster takes the money instead of earning it
(\S\ref{sec:nego_fraud}), so a counterpart has to be chosen as well as bargained with.

\subsubsection{Bargaining Policy Preliminaries}
\label{sec:nego_prelim}

Sourcing one SKU begins with a message to one supplier and ends in a price or in nothing. Neither side sees the other's limit. A supplier never reveals the floor below which it
will not sell, so each purchase is a bilateral bargaining problem under incomplete information
\citep{harsanyi1967games, chatterjee1983bargaining, rubinstein1982perfect}. One such cycle is
a session $z=(s,j,c)$, the $c$-th between supplier $s$ and SKU $j$. No rule closes a session
at a fixed round. A deadline scale sets the time pressure instead, and past its midpoint a supplier can walk away from an offer under its floor
(Appendix~\ref{app:nego_kernel}).

One pair from the evaluation fixes the scale. Ridge Express quotes a single SKU at
\textyen 149.81, and the agent resells against a market reference of \textyen 199.75.
Beneath both lies a floor of \textyen 101.27 the agent never sees, leaving \textyen 98.48
for the two sides to divide.
Figure~\ref{fig:anchoring_example} follows one such session offer by offer in panel (a), and
the nine deals the pair produced across the year in panel (b).

Two prices bracket what a session can be worth. The agent resells against a public market
reference price $v_j$ (\S\ref{sec:economy}), while supplier $s$ holds a private floor
$c_j^s$. Bargaining runs in the gap between the two, the zone of possible agreement (ZOPA),
so a deal at unit price $p_z$ for $q_z$ units hands the agent $(v_j - p_z)\,q_z$ of it. How
wide that gap is, how badly the supplier needs the sale and how hard it will push are drawn
once and stay hidden. The agent bargains without knowing how much room it has.

Sessions are divided by counterpart type, one term of the objective per side.
Writing $\mathcal{Z}_{\mathcal{G}}$
for those concluded with the honest suppliers $\mathcal{G}$ and
$\mathcal{Z}_{\mathcal{B}}$ for those concluded with the fraudulent ones $\mathcal{B}$
(\S\ref{sec:nego_fraud}), the year-end asset goal of \S\ref{sec:overview} becomes a
procurement objective.
\begin{equation}
\label{eq:nego_objective}
\max_{\pi} \;
\underbrace{\sum_{z \in \mathcal{Z}_{\mathcal{G}}}
\mathbf{1}[f_z \neq \bot]\,(v_{j(z)} - p_z)\,q_z}_{\text{surplus extracted from honest suppliers}}
\;-\;
\underbrace{\sum_{z \in \mathcal{Z}_{\mathcal{B}}}
\mathbf{1}[f_z \neq \bot]\; \ell_z}_{\text{losses paid to fraudulent suppliers}}.
\end{equation}
Here $f_z$ is the session outcome, $\bot$ a disagreement worth nothing, and $\ell_z$ what a
fraudulent deal costs in the end, whether overpayment, fees, short shipment or defective
stock.

Two competencies follow, and they pull against each other. Winning surplus from an honest
supplier takes firmness; refusing a fraudulent one takes walking away, and the firmness that
wins a good price is also what produces impasse \citep{raiffa1982art}. Both are paid in the
cash that sets the primary score.

\subsubsection{Two-Layer Supplier Design}
\label{sec:nego_twolayer}

An outcome difference says something about the agent only if every model met the same
counterpart economics in every run. A supplier played entirely by an LLM breaks that twice over.
Sampling moves its prices from run to run, and it stays \emph{persuadable}, so an eloquent
enough agent can talk it below its own floor and turn the benchmark into a jailbreaking
contest. Every supplier therefore splits into two layers.

\textbf{Layer 1, the Deterministic Negotiation Kernel.} Every price, concession and accept
or walk-away decision comes from a deterministic kernel rather than from a model. The kernel
builds on the counterpart policy of \citet{zhang2026terms}, reimplemented independently in a
seller-only role and re-parameterized from the data layer (Appendix~\ref{app:nego}). Acceptance
and walk-away are stochastic \citep{baarslag2014acceptance}, and a latent cue carries
sentiment and posture (Appendix~\ref{app:nego_kernel}). Counter-offers answer the agent's own
concession speed rather than the clock \citep{faratin1998negotiation}, so an agent that
concedes fast meets a \emph{firmer} counterpart. Fast concession carries the same disadvantage
in human bargaining \citep{tey2021impact}.

One kernel serves one (supplier, SKU) pair. Its reservation price and opening quote come
from the data layer, so both are properties of the SKU rather than draws. Its seed makes an
identical agent trajectory meet an identical supplier response
(Appendix~\ref{app:nego_grounding}). Two sources of run-to-run variance survive. The Layer 2
renderer is an LLM sampled at its provider's default decoding, and the delivery days a
supplier's catalog advertises are drawn outside the episode seed, so they can move between
runs. Prices and accept or
walk decisions cannot move, but the narrative a model reads to detect fraud does, so the
fraud axis of \S\ref{sec:res_fraud} carries renderer sampling noise on top of policy
sampling noise.

Honest suppliers draw from six behavior templates. The templates differ in how sharply the
kernel answers the agent's concession speed and in how much its cues give away. Which supplier gets
which is fixed in the released roster, so the difficulty mix is set rather than drawn per
run (Appendix~\ref{app:nego_presets}). Neither the template nor the honest or
fraudulent label is visible to the agent.

\textbf{Layer 2, the NPC renderer.} A second LLM turns each committed kernel decision into
supplier chat. The renderer has to relay the exact price and decision and takes only tone
from the kernel's sentiment cue, which keeps the dialogue rich and economically inert.
Compliance does not rest on that instruction. An accept is refused unless its price matches the
kernel's standing quote to within $\pm 0.005$, so a renderer that misstated a price could
never make it a transaction (Appendix~\ref{app:nego_protocol}). In the other direction, a
deterministic parser reads the agent's own offer, accept and reject actions out of the same
free-form message. One economic action travels in the prose instead, enrolment in a
supplier's membership program, which an intent classifier recognizes from the agent's own
words and bills at \textyen 1{,}000 (Appendix~\ref{app:nego_fraud_catalog}). Prices carry no
signal about fraud by design, so on the pre-deal class the narrative holds the decisive
evidence, and no fixed template could mount it.
Figure~\ref{fig:session_example} prints one session end to end, every supplier line beside
the kernel decision it had to relay.

\subsubsection{Multi-Round, Cross-Supplier, Repeated Bargaining}
\label{sec:nego_repeated}

One session is usually the unit of evaluation in existing negotiation benchmarks, and the
kernel as described so far governs exactly one. Across a full year the game becomes
\emph{repeated}, which asks whether an agent turns what one encounter taught it about a
counterpart into a better price at the next. Inferring a private reservation value from past
encounters \citep{baarslag2016learning} therefore becomes part of the task rather than an
assumption. Closed deals are not re-surfaced, so an agent keeps its own price history or does
without it.

Once a session over $(s, j)$ concludes, the next offer on that pair opens a fresh cycle under a new
kernel instance and a new seed, carrying the \emph{same} persistent type and reservation
price. A seller-side kernel never accepts below that reservation, so the best price won so
far is an upper bound on $c_j^s$ that only tightens. The competence it rewards is to
\emph{anchor} each reopening at or below that historical best. Calling the rule optimal
would go too far, since against a pre-deal fraudulent counterpart the historical best tightens
onto a floor inflated on purpose (\S\ref{sec:nego_fraud}).

\begin{figure}[t]
  \centering
  \includegraphics[width=\linewidth]{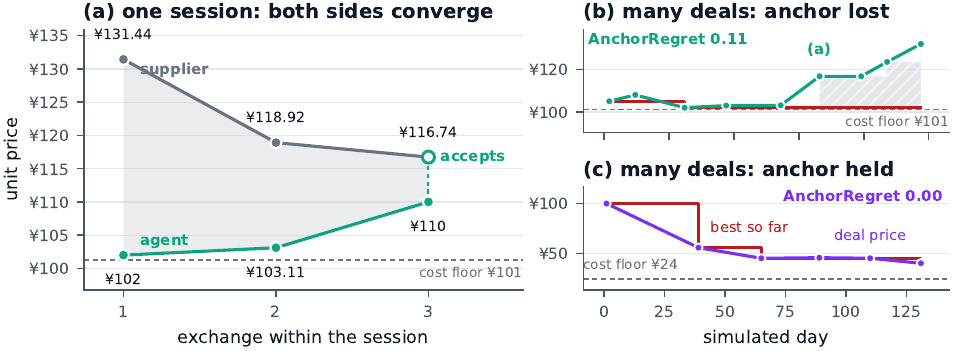}
  \caption{Negotiation at two timescales, from real episodes, each panel on its own price
  axis and the cost floor hidden from the agent. \textbf{(a)} One session offer by offer, the agent bidding up and the kernel conceding
  down until the agent accepts its standing quote.
  \textbf{(b)} The same (supplier, SKU) pair all year, where that session ends a run of
  near-floor deals and the price never recovers, the hatched wedge being the overpayment
  $\mathrm{AnchorRegret}$ averages. \textbf{(c)} Another agent instead pushes lower at
  nearly every restock.}
  \label{fig:anchoring_example}
\end{figure}

Learning runs on both sides of the table. Each supplier's deal history goes into its prompt,
so suppliers ``remember'' what they sold before. Several honest suppliers serve every
category alongside a few fraudulent ones, so the agent can also arbitrate \emph{across}
counterparts, probing one supplier's floor and holding it as an outside option against
another. Single-session skill and year-long accumulation then separate cleanly, the first
measured in per-session surplus, the second in how fast the price for a repeatedly sourced
SKU walks down toward the floor (\S\ref{sec:nego_metrics}).

Every session costs twice over besides. Bargaining spends minutes out of the working day
(\S\ref{sec:turn_loop}), and an agreement spends cash out of a bank balance that refills
only when escrow settles (\S\ref{sec:econ_settlement}). A session excellent in surplus terms
can still be the wrong one to have opened.

\subsubsection{Adversarial Supplier Fraud}
\label{sec:nego_fraud}

Of the 576 suppliers on the roster, 152 are fraudulent. Each runs the hardball kernel preset
under one of five scripted scam overlays that the renderer performs. The five split into two
classes by \emph{when} the damage lands
(Appendix~\ref{app:nego_fraud_catalog}). No single observation gives a fraudulent
counterpart away. Opening quotes are uninformative by construction
(\S\ref{sec:nego_twolayer}), and the behavior family stays latent and is shared with honest
suppliers. What does separate the two populations is the \emph{rate} of concession, since
fraudulent kernels hold out longer than the hardest honest preset. That signal lives only in
the shape of a multi-round exchange, so an agent that accepts early gives up its one
bargaining-side cue and has to read the narrative instead.

\textbf{Pre-deal scams} make their money inside the negotiation. Their kernels carry a
reservation price inflated toward $1.5\times$ the honest cost floor, so \emph{any} deal
overpays what an honest competitor would have charged. The inflated floor still sits inside
the range an honest supplier might quote, so only the narrative gives it away, offering
membership fees, discounts that never arrive, or manufactured urgency.

\textbf{Post-deal scams} keep the honest cost floor and quote from that same range, then
defect at fulfillment. A short delivery brings 60\% to 70\% of the quantity paid for, and
defective stock returns at $0.40$ or double the SKU's natural rate, whichever is higher,
capped at $0.95$ (\S\ref{sec:economy}). The defective rate is
blended across the SKU's pooled stock, so interleaving honest and fraudulent deliveries
dilutes the evidence and no observed return rate points back to a culprit. Scoring therefore
counts the money actually lost to these counterparts rather than judging intent
(\S\ref{sec:nego_metrics}), which keeps the axis on the cash of
Equation~\eqref{eq:nego_objective}.

\subsubsection{Negotiation Metrics}
\label{sec:nego_metrics}

Both terms of Equation~\eqref{eq:nego_objective} are scored as the fraction of the
bargaining range an agent captures, standard practice in automated negotiation
\citep{raiffa1982art, baarslag2013practical}. The metrics build on the surplus efficiency of
\citet{zhang2026terms}, re-based from sampled ZOPA geometry onto data-grounded floors, and Appendix~\ref{app:nego_metrics} carries every formal definition.
Agreement is close to universal on the honest sessions a model concludes, so
$\mathrm{CSE}^+$ carries the headline, the share of a pair's bargaining range that a closed deal captures. Beside it $\%\mathrm{Oracle}$ compares the same surplus against closing every honest deal at the supplier's floor, and the agreement rate stays in
Appendix~\ref{app:nego_metrics}. Neither headline metric weights a
deal by quantity, and both are rates over a session set the agent chose itself, so they measure per-deal bargaining skill rather than procurement cash
saved. Two models compare on them only if their portfolios carry similar bargaining ranges, which \S\ref{sec:res_nego} checks by reporting deal volume beside the rate.
Conclusion is the operative word. A session abandoned mid-bargain enters no denominator and costs nothing in any of the three metrics, so \S\ref{sec:res_nego} reports abandonment counts beside them rather than folding them into a rate. Fraud avoidance is reported as $\mathrm{BadSpend}\%$ rather than as an agreement rate over fraudulent sessions, because the agent also chooses whom to contact at all. A session-rate denominator would reward never looking, and cash is in any case the quantity a scam actually costs.

\paragraph{Sequential price discipline}
Over 365 days, the same supplier is asked for the same SKU many times, which raises a
question no session-level metric answers. Does an agent's next order benefit from the price it already won? Holding an anchor like that has documented value in human bargaining \citep{petrowsky2025power}. The measure looks at every honest (supplier, SKU) pair traded more than once. Each agreed price is expressed as its position $\pi_z$ inside that pair's bargaining range, then scored after the first deal against the best the agent had already won there. $\mathrm{AnchorRegret}$ averages the resulting overpayment.
$\mathrm{AnchorRatio}$ divides that average by a null which holds each pair's prices fixed and reshuffles only the order they arrived in. Significance travels separately in the permutation $z$-score $z_{\mathrm{anchor}}$.

The null assumes a pair's cycles are equally hard, and that assumption is checkable rather
than asserted. Two of the three quantities that set that difficulty are constants of the SKU, and the
archive records the cost floor and the nominal wholesale reference unchanged across all
$8{,}647$ re-orders. The quote the kernel actually opens with is redrawn each cycle and the
archive does not keep it, so a rising $\pi_z$ is read against a fixed floor rather than
against a fixed opening.
The measure reports an outcome and not its cause, and \S\ref{sec:res_learning} takes up
which mechanism produces it. Figure~\ref{fig:anchoring_example} shows both outcomes on real
trajectories, one agent giving back a price it had already won while another walks its price
down the range all year.

Three limits bound the scope. Only honest pairs count, since a pre-deal fraudulent floor is
inflated by construction (\S\ref{sec:nego_fraud}). Each re-order counts equally regardless
of order size, so the measure captures per-unit discipline and not its cash consequence.
No observations come from an episode that ends early.

\subsection{Data Layer}
\label{sec:data}

Every number E-Commerce Bench runs on comes from a real e-commerce platform, either
desensitized straight off it or synthesized by hand against an analysis of its data logs, the supplier price floors of \S\ref{sec:nego_engine} and the demand parameters of \S\ref{sec:economy} among them.

\subsubsection{Catalog and Suppliers}
\label{sec:data_real}

The environment ships 6{,}886 products in 60 categories, served by 576 suppliers, 424 of them honest and 152 fraudulent. The 12 store types span 9 categories down to 1, and the calendar fixes 10 market events and 8
promotions. Desensitization maps the products, the categories, the per-category
return-rate tiers, the monthly-sales bands and the holiday calendar straight off a real e-commerce platform. Synthesis from real logs supplies the rest, namely the four price-elasticity families, seasonality curves, reputation parameters and supplier attributes.
Promotion names are fictional, since a model that recognizes China's largest shopping festival could price from memory rather than from evidence \citep{zhu2025agenticbench}. Fraud reaches every product line rather than concentrating in a few. Each supplier serves exactly one category, no category holds fewer than 7 honest and 2 fraudulent, and coverage stays flat across catalog depth (Figure~\ref{fig:catalog}d). Every fraudulent one runs a scam of \S\ref{sec:nego_fraud}. The supplier list is fixed at release rather than redrawn per episode, and Appendix~\ref{app:data} lists all 60 categories with their parameters.

\subsubsection{Systematic Information Asymmetry}
\label{sec:data_hidden}

An agent sees the catalog and the outcomes of its own actions, never the demand and
cost-floor parameters that decide what an action is worth. A market lookup answers at one of three levels: the coarsest naming every store type and the finest one category's sales band (Table~\ref{tab:tools}), each substituting a label for the quantity behind it. A profit-potential grade replaces demand capacity, a margin class the wholesale ratio, and a sentence about returns the per-SKU rate. Supplier lookup withholds most, naming no type and ranking nothing (\S\ref{sec:nego_twolayer}). Pre-deal scams show in the narrative from the opening quote (\S\ref{sec:nego_fraud}). Post-deal ones are written to read like ordinary suppliers, so only the goods that arrive expose them, and the cash has left by then. Table~\ref{tab:asymmetry} in Appendix~\ref{app:data} sets each observable against the quantity it withholds and prices the discovery. Portfolio decisions therefore rest on inference, and hidden parameters differ widely in how fast they can be learned. The best price already won from a supplier bounds its floor from above (\S\ref{sec:nego_repeated}), and a SKU's natural return rate becomes readable once its stock has moved. An elasticity parameter stays hidden all year, since four unprinted factors multiply the same observation. A store type's earning ceiling shows only after one has been operated for a year. Exploration and operation both cost time (\S\ref{sec:agent_loop}), so the year rewards allocating discovery rather than exhausting it.

\subsection{Evaluation and Metrics}
\label{sec:eval}

The primary score is \textbf{end-of-year total assets}, the sum of the bank account, the platform wallet, and unsettled escrow, reported as the \emph{asset multiplier} over the \textyen 100{,}000 opening stake.
Bankruptcy ends a run before the horizon (\S\ref{sec:econ_settlement}). Truncated runs enter the average rather than being dropped, so a five-run mean can mix them with full years. Six capabilities are measured around the year-end assets:

(1) Bargaining is scored on how far down the agent pushes a price. $\mathrm{CSE}^+$ ranks \textbf{negotiation quality}, how much of the distance from the public reference price down to the supplier's cost floor a closed honest deal covers, where $1$ means buying at the floor and $0$ means paying the reference price, averaged over sessions with no weight for order size. Recorded around it are $\mathrm{SE}^+$, $\%\mathrm{Oracle}$, $\mathrm{AGR}^+$, rounds to deal, deals closed, and both surplus and spending broken out by the supplier's behavior template (\S\ref{app:metrics_d1}).
(2) \textbf{Fraud avoidance} follows the cash rather than the conversation. Ranking runs on $\mathrm{BadSpend}\%$, the share of order spending that cleared into a fraudulent supplier's account, low being the better direction.
We also record the fraud agreement rate (the percentage of fraudulent sessions ending in a deal), alongside diagnostics covering losses across the five scams, membership fees paid, and fraudulent suppliers contacted or ordered from (\S\ref{app:metrics_d2}).
(3) Solvency is judged by how much of its own best position a year gave back. Peak drawdown over peak total assets ranks \textbf{cash flow and solvency}, since the archive records that fall in absolute \textyen{}, and a business ten times the size books ten times the fall for the same discipline. The unnormalized fall, the bankrupt runs, the overdrawn days, the deepest bank balance, the idle platform wallet, and the largest warehouse held round out the dimension (\S\ref{app:metrics_d3}).
(4) \textbf{Operational efficiency} prices a year of profit against the calls that produced it. Mean profit over mean tool calls orders the models. Sitting next to it are the call count, its split over eight activity bands, the turns, the eviction passes the transcript needed, and the memory-store traffic
(\S\ref{app:metrics_d4}).
(5) \textbf{Operations execution} is ranked by the controllable return rate—the excess returns driven by the model's pricing above the baseline (accrued at dispatch; negative for discounted prices), excluding fraud-induced defects. Supporting diagnostics include realized returns, refund and freight losses, on-time delivery rates, orders lost to the two-day deadline, tier mix, and store counts (\S\ref{app:metrics_d5}).
(6) Learning is read from what an agent does every time after the first that it buys the same item from the same supplier, which over a year is usually several times.
$\mathrm{AnchorRatio}$ ranks \textbf{learning over the horizon} by measuring how much each repeat order overpays against the cheapest that agent had yet paid that supplier for that item. Beside it lie the same overpayment before its permutation null,
$\mathrm{AnchorRegret}$, the $\mathrm{NewLow}$ rate, the combined $z_{\mathrm{anchor}}$, the repeat-order count and three half-year comparisons (\S\ref{app:metrics_d6}).

%% file: sections/4_experiment.tex
\section{Experimental Results}\label{sec:experiment}

\subsection{Experimental Setup}
\label{sec:setup}

Eighteen models ran the 365-day task, eight proprietary and ten open-weight by
our own assignment, ranked within tier by mean total assets
(Table~\ref{tab:models}).
Each model received five independent episodes, 90 in all, under one fixed world. Requests use each provider's default settings, with no
temperature or \texttt{top\_p} sent and thinking force-enabled wherever a family
exposes it (Table~\ref{tab:models}). Five episodes is a small sample, so every mean in Table~\ref{tab:leaderboard} is printed with its standard deviation
\citep{miller2024errorbars, kapoor2025agents}.
Otherwise, all models shared the same 18 tools (Table~\ref{tab:tools}), context window and group-wise eviction (\S\ref{sec:context_mgmt}), and the memory store outside it (\S\ref{sec:memory}). The longest episode took 3{,}599 turns of the 4{,}000
allowed, and all 90 ended on the simulated calendar or in bankruptcy rather than on
that cap.
One renderer voices every supplier for every model, so wording varies between
episodes while the economics it relays cannot (\S\ref{sec:nego_twolayer}).
Scoring reduces the year to one number, the asset multiplier, end-of-year total
assets over the \textyen 100{,}000 stake (Appendix~\ref{app:metrics}). A bankrupt
episode stops on the tenth consecutive negative bank day and still enters the
average.

\subsection{Overall Performance}
\label{sec:overall}

Starting from the same stake in the same simulated year, the models finish orders of
magnitude apart. A factor of $1{,}264$ separates GPT-5.6 Sol at the top from Qwen3.5-Plus
at the bottom (Table~\ref{tab:leaderboard}, Figure~\ref{fig:assets}). The
environment is deterministic given the action sequence (\S\ref{sec:setup}), so that
range belongs to the agents rather than to resampled demand.
Proprietary systems take the top four places (Table~\ref{tab:leaderboard}).
Qwen3.8-Max-Preview leads the open-weight tier and places fifth overall. Inside the
Qwen family each release earns more than the one before it, and  Qwen3.8-Max-Preview earns the most. The 3.6 to 3.7 step is worth noting, since Qwen3.7-Max swings from ¥6,290 to ¥358,790 across its five years, and its worst year finishes below every Qwen3.6-Plus year. Strong models go bankrupt too. Ten of the ninety episodes
ended insolvent, six of them under closed-source models. GPT-5.5, Claude Opus 4.6,
Gemini 3.1 Pro and Qwen3.5-Plus account for all ten, and the same four surrender more than half of peak total assets (\S\ref{sec:res_cashflow}).
Ordering by assets recovers only part of the skill behind it
(Figure~\ref{fig:ranks}). Section~\ref{sec:multidim} takes the axes one at a time, and
Appendix~\ref{app:results} carries the per-model table of every dimension.

\input{tables/tab_4_2_leaderboard}

\subsection{Multi-Dimensional Analysis}
\label{sec:multidim}

\begin{figure}[t]
  \centering
  \includegraphics[width=\linewidth]{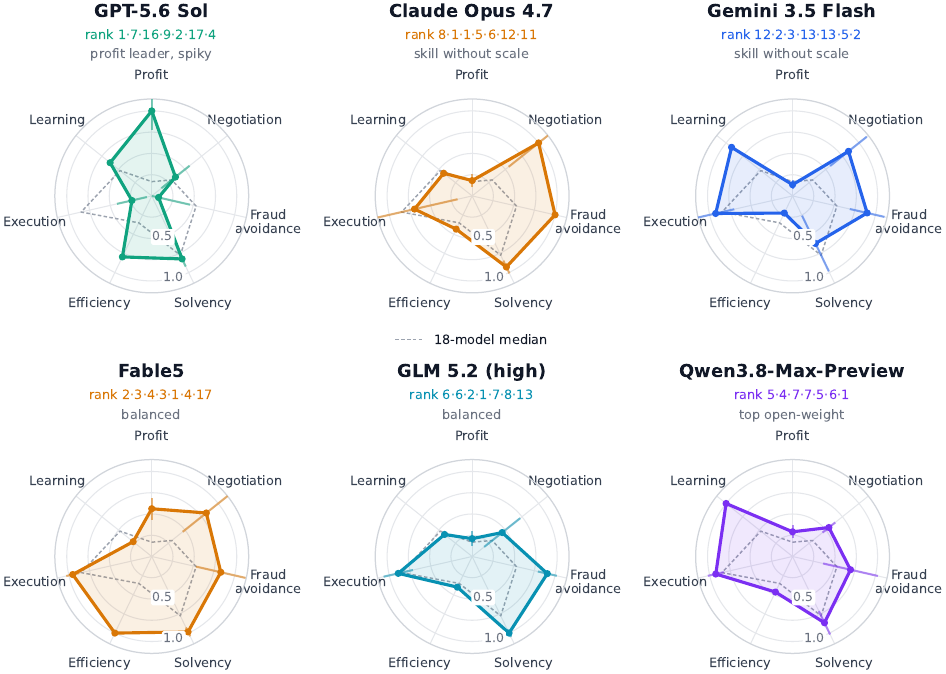}
  \caption{Capability profiles for six of the 18 models. The primary score and the six dimensions of this section run clockwise from the top, profit as mean end-of-year total assets, negotiation as $\mathrm{CSE}^+$, fraud avoidance as $\mathrm{BadSpend}\%$, solvency as peak drawdown over peak total assets, efficiency as profit per tool call, execution as the controllable return rate and learning as $\mathrm{AnchorRatio}$. Fraud avoidance, solvency, execution and learning are sign-flipped so that higher is better on every axis. Each axis is min-max normalized over the 18 model means, so 1 marks the best mean and 0 the worst, with whiskers over the five episodes and a dashed polygon at the median. Efficiency carries no episode spread, being a ratio of 5-run means, and the archived $\mathrm{AnchorRatio}$ spread is a population std over a varying number of episodes, so the learning axis draws no whisker either. No profile fills the polygon.}
  \label{fig:profiles}
\end{figure}

Six dimensions pull the year apart, each ranked by one recorded measure and defined in \S\ref{sec:eval}, with the formulas, field names, and caveats in one appendix subsection per dimension (\S\ref{app:metrics_d1} to \S\ref{app:metrics_d6}).
Figure~\ref{fig:profiles} sets the primary score beside the six axes for six models, and Appendix~\ref{app:res_profiles} carries all 18.

\subsubsection{Negotiation Quality}
\label{sec:res_nego}

% source: summary.json CSE+ for the two endpoints, 0.8107 Claude Opus 4.7 and 0.5957 Kimi
% K2.6, both re-printed by verify/num_4_3_1_negotiation.py section 1. The 0.50 reference is
% the wr_j = round(0.5 + 0.5 cfr_j, 2) identity over all 60 rows of
% data/category_params.csv, worked out in \S\ref{app:nego_grounding}, where accepting the
% opening quote scores (1 - wr_j)/(1 - cfr_j), exactly 0.5 before the rounding and 0.4667 to
% 0.5185 after it. Per-template spans are printed by figscripts/fig_app_family_heatmap.py.
% Every remaining per-model value stays in Table~\ref{tab:res_d1}.
No model bargains prices down to the supplier's cost floor. Claude Opus 4.7 keeps the most of
each bargaining range at $0.811$ and Kimi K2.6 the least at $0.596$, with the per-model
values in Table~\ref{tab:res_d1}. Accepting the opening quote already scores $0.50$, so the
field lands $0.10$ to $0.31$ above what taking the first offer would have cost it
(Appendix~\ref{app:nego_grounding}). Which counterpart a model draws matters less than which
model it is, since the field spreads $0.051$ across the six behavior templates against
$0.204$ across models, though the per-model ordering inside a column still moves by as much
as thirteen places (Figure~\ref{fig:family_heatmap}).
Agreement rates sit at the top of their scale and separate nobody. Of the $12{,}060$
concluded honest sessions, 73 end without a deal, and 72 of those are the agent's own
rejection rather than a supplier walking away. Every model settles at least $96\%$ of the
honest sessions it concludes, so the ranking rests on how much a deal captures rather than on
how often one happens (Appendix~\ref{app:metrics_d1}). The rate also leaves out the $10.9\%$
of sessions a model opened mid-year and never came back to.
Agents buy least from the counterparts that concede least. The hardest of the six honest
behavior templates (Appendix~\ref{app:nego_presets}) covers 48 of the 424 honest names on
the supplier list, and those 48 take $4.8\%$ of the \textyen 55.1M of honest procurement,
against at least $14\%$ for each of the five easier ones (Figure~\ref{fig:honest_spend}). Buying the same item from the same honest
supplier again costs more than a reshuffle of the prices already won would have, for
16 of the 18 models, and \S\ref{sec:res_learning} takes that up with the rest of the year's
trends.
\subsubsection{Fraud Avoidance}
\label{sec:res_fraud}

Every model sends part of its procurement budget to suppliers that defraud it. Spending on
them spans more than 160-fold, cleanest at Claude Opus 4.7 and heaviest at Qwen3.5-Plus
(Figure~\ref{fig:fraud}), and none spends as though it never screened.
$\mathrm{BadSpend}\%$ counts money rather than refusals (\S\ref{app:metrics_d2}), unlike a
refusal rate on scripted solicitations \citep{yang2025fraudr1}, so declining an offer the
agent never sought earns no credit.
A share and a sum of money rank the models differently, because procurement budgets differ
as widely as the shares do. Qwen3.5-Plus gives up the largest share, yet GPT-5.6
Sol hands over fifty times the cash on a smaller one (Table~\ref{tab:res_d2}).
Both readings rest on a handful of orders, and across five episodes the standard deviation
of the share exceeds its own mean for 15 of the 18 models.
Losing money to fraud does not cost a model its year. The cleanest buyer finishes mid-pack on assets and the profit leader is 16th of 18 on $\mathrm{BadSpend}\%$
(\S\ref{sec:overall}), because a pre-deal scam inflates a price that still sits inside the honest range and a post-deal scam delivers most of the goods (\S\ref{sec:nego_fraud}). Most of the money disappears after a deal is signed. Defective lots take $53.4\%$ of everything reaching fraudulent suppliers and short shipments the next largest slice (Figure~\ref{fig:fraud}), while membership-fee scams almost never land, three payments in the whole evaluation.
Models separate when they order, not when they choose whom to approach. Fraudulent
suppliers take close to their share of the supplier list in every model's outreach
(\S\ref{sec:nego_twolayer}), and what differs is how often an approach becomes an order,
from $4.0\%$ for Claude Opus 4.7 to $31.7\%$ for GPT-5.6 Sol (Figure~\ref{fig:funnel},
Appendix~\ref{app:res_funnel}). A negotiation that reaches its end almost always closes a deal (\S\ref{sec:res_nego}), so the gap lies in the approaches a model abandons.
\subsubsection{Cash Flow and Solvency}
\label{sec:res_cashflow}

Money leaves the bank account weeks before it returns. Revenue waits nine days in escrow,
then waits again for a withdrawal (\S\ref{sec:econ_settlement}). No episode sees cash in
the wallet before day 15, and 49 of the 90 bottom out in January. Two GPT-5.5 episodes
went bankrupt on day 17, before a single yuan of revenue had settled. Returns claw more
money back out of escrow (\S\ref{sec:res_execution}).
A fall recorded in yuan grows with the balance at risk, and the largest of the field belongs
to GPT-5.6 Sol, which also earned the most (Appendix~\ref{app:metrics_d3}). Dividing each fall
by the peak total assets of its own episode separates the field on risk, and the five
models that surrendered more than half of peak include all four that ever went bankrupt
(Figure~\ref{fig:cashflow}a, Table~\ref{tab:res_d3}).
Running the bank account below zero was rare and usually fatal. Seventeen of the 90
episodes went negative at a morning check across 177 such days, and ten of them went
bankrupt. A run ends only after ten consecutive overdrawn mornings, so one Claude Opus 4.6
episode absorbed fifteen scattered overdrafts and survived. On 66 of those days the wallet
already held the shortfall (Figure~\ref{fig:cashflow}b). No tool reports the streak
(\S\ref{sec:agent_loop}), so an agent cannot count the mornings it has failed.
A large cash buffer buys survival and caps growth. Three models kept the bank above
\textyen 15{,}000 in every episode and finished sixth, tenth and eleventh, while the two
biggest earners held the thinnest average buffers of the fourteen solvent models.
Qwen3.8-Max-Preview leads the open-weight field on a mid-field buffer, its bank above zero
all year.
\subsubsection{Operational Efficiency}
\label{sec:res_efficiency}

\begin{figure}[t]
  \centering
  \includegraphics[width=\linewidth]{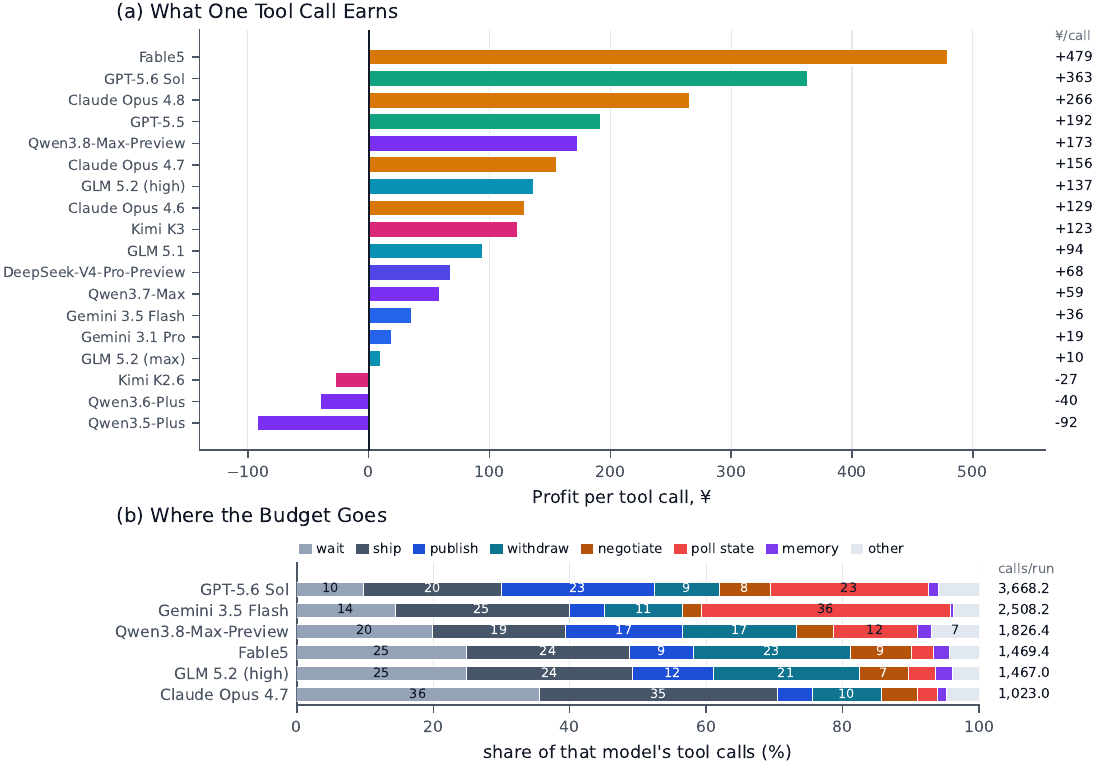}
  \caption{(a) Profit per tool call for the 18 models, best first, mean total assets less the \textyen 100{,}000 stake divided by mean tool calls per episode. Bankrupt episodes are pooled into both means, and the three models that finish under the stake sit left of zero. (b) Where six models spent their calls, over eight groups of the 18 tools, scaled within each model.}
  \label{fig:efficiency}
\end{figure}

Fable5 earns more from a tool call than the leader does. Sitting second on total
assets, it turns \textyen 479 of profit out of each call against GPT-5.6 Sol's
\textyen 363, on $59.9\%$ fewer calls (Table~\ref{tab:res_d4}). Three
models finish under their opening stake, so their calls average a loss over the year,
\textyen 92 a call at the worst. No fee inside the economy charges for a call, so
the bill falls on whoever deploys the agent and the score stays blind to
it~\citep{kapoor2025agents}. The Berkeley function-calling leaderboard reports that bill
directly, printing dollar cost and latency beside accuracy~\citep{patil2025bfcl}.
Where the calls go describes a policy better than their number does
(Figure~\ref{fig:efficiency}b). Shipping, advancing the day, withdrawing settled revenue
and publishing inventory absorb $71.8\%$ of all 18 models' calls, a daily loop rather than
a plan. Reading state back takes another $14.7\%$ and supplier conversation $6.0\%$, so the
one channel that can change what an item costs draws less than a fifteenth of the
effort.
Pricing barely registers in the call budget. One call can revise many prices at once, yet
$0.66\%$ of all calls do so. Nine of the 18 models average under three such calls a year,
leaving their listed prices to stand while season, promotion and reputation move demand
underneath them.
The memory store draws almost no traffic. Notes the eviction pass cannot
reach (\S\ref{sec:memory}) take $2.1\%$ of the field's calls. Claude Opus 4.8 gives it the largest
share of its own budget, $4.7\%$ at $69.6$ calls a year against 20 slots, while GPT-5.5 makes
the most calls outright at $135.2$ and Qwen3.5-Plus wrote to it in none of its five
episodes. Twenty notes
revised a few dozen times is what a year of trading leaves behind.

\subsubsection{Operations Execution}
\label{sec:res_execution}

The controllable return rate counts the percentage points a model's own pricing adds above the natural rate of the goods it chose to sell, and it is the one return channel this axis
charges to the agent. The points run from $-0.05$ for Qwen3.7-Max to $5.88$ for Qwen3.5-Plus (Table~\ref{tab:res_d5}). The natural rate underneath is larger, $5.31$ to $21.25$ points, and it follows what a model chose to sell rather than how it priced. Twenty-six of the 90 episodes carry a negative rate, priced under the reference. Added returns are worth buying when the price that causes them pays for them. GPT-5.6 Sol books $4.46$ points and the year's largest refund bill, \textyen 1.62M, and finishes first on assets. Qwen3.5-Plus adds more return points than any other model and finishes last.
Freight speed is chosen once and left alone. The archive holds $92{,}182$ fast dispatches beside $171{,}642$ standard and exactly one slow, and 72 of the 88 episodes that sold anything put at least $98\%$ of shipments in one tier. Fast doubles the freight bill and cuts the return draw by a quarter, yet nine models swing over 30 points in fast share between their own episodes, so the tier reads as sampling, not policy. Store openings order nothing. Cumulative openings run from $3.0$ for two Qwen models to $9.0$ for GPT-5.6 Sol, each charging the \textyen 500 setup fee again and restarting a reopened type's reputation at the floor. GLM 5.2 (high) opened exactly four every episode and placed sixth, while Gemini 3.5 Flash opened 13 in one episode and placed twelfth, and neither is a profitable habit.

\subsubsection{Learning over the Horizon}
\label{sec:res_learning}

Two of the 18 models pay less on a repeat order than a reshuffling of the prices they already won from that supplier would have cost. Qwen3.8-Max-Preview scores
$0.834$ on $\mathrm{AnchorRatio}$ and Gemini 3.5 Flash $0.918$, where $1.0$ means an agent's price ordering is indistinguishable from a random ordering of its own prices. The field median is $1.369$, and Qwen3.5-Plus reaches $1.860$ (Table~\ref{tab:res_d6}). Fifteen models miss their own permutation null by more than two standard deviations in the expensive direction, which is what an unrevised anchor on an early price produces \citep{lou2026anchoringllm}. GPT-5.5 contributes three episodes instead of five; its two bankrupt runs never buy the same item twice from the same supplier. A second reading of the same year runs the same way, with sixteen of the 18 models capturing less surplus in the second half of an episode than in the first. The surplus half-lift splits an episode's honest sessions at the median day and differences the two means (\S\ref{app:metrics_d6}).
Three mechanisms would hold that surplus trend flat, and the evaluation separates none of them. Eviction destroys the tool result that carried a price already won
(\S\ref{sec:context_mgmt}). The memory store that survives eviction is barely used
(\S\ref{sec:res_efficiency}). And no gradient reaches inside an episode, so improvement must come from an agent reasoning over its own record \citep{shinn2023reflexion}. Qwen3.8-Max-Preview moves against the field, alone holding a positive per-day surplus trend and alone beating its own shuffled prices by more than two standard deviations. Supplier choice drifts as well. Fraudulent suppliers take a larger share of concluded sessions late in the year than early for 17 of the 18 models, so what moves over the horizon moves the wrong way.

\begin{figure}[t]
  \centering
  \includegraphics[width=\linewidth]{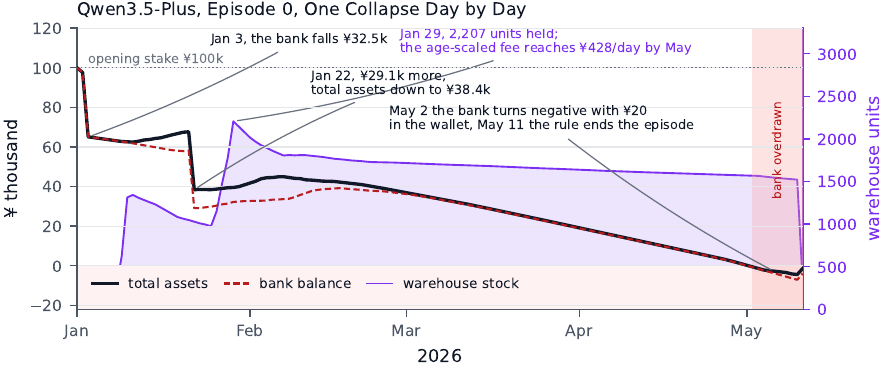}
  \caption{One bankruptcy episode day by day, Qwen3.5-Plus episode 0. Total assets and
  bank balance are on the left axis, warehouse stock on the right, and the shaded band
  marks the days the bank balance stayed negative.}
  \label{fig:collapse}
\end{figure}

\subsection{Error Analysis}
\label{sec:errors}

The six axes rank models without locating the mistake, since one payment to a supplier that cheats surfaces as fraudulent spend and again as a deeper fall from peak assets. Ten recurring patterns therefore get one detection rule each, with mechanisms and counts in Table~\ref{tab:failure_modes} and Figure~\ref{fig:mode_incidence} of Appendix~\ref{app:case_rules}. Bankruptcy alone destroys the stake (\S\ref{sec:res_cashflow}). Inventory bought and never sold is the road there. One Qwen3.5-Plus episode fills four stores in January, never sells its way out, and ends in May (Figure~\ref{fig:collapse}). Fraudulent suppliers charge for value they never deliver, inflating prices before goods move and shipping short afterwards. A second order with the same supplier is where most of the money goes: 943 of the 1{,}141 agreements struck with fraudulent suppliers sit in a supplier-item pair dealt with twice or more (\S\ref{sec:res_fraud}, Appendix~\ref{app:nego_metrics}). No message names the supplier behind a bad lot, so the loss lands in the balance instead of arriving as a signal.
Money lost in bargaining is small per deal and never fatal. Honest deals rarely close half the range available below, and the giveaway is the opening quote accepted unimproved (\S\ref{sec:res_nego}). Claude Opus 4.7, Gemini 3.5 Flash and Fable5 never accept one and are the top three on $\mathrm{CSE}^+$, Kimi K2.6 most often and last. Research gets re-derived instead of remembered. Of 2{,}180 catalog and supplier queries, 746 repeat arguments already issued that episode. The repeats cost 7{,}460 simulated minutes, 83 an episode against a 219{,}000-minute year, and sit beside a little-used memory store (\S\ref{sec:res_learning}). Nothing in the economy charges for asking twice. The two-day shipping deadline binds only at the edges of the year. Cancellations cluster in the first fortnight and again in the final two weeks, where the agent has either not started shipping or has stopped (\S\ref{sec:res_execution}).

Three patterns often blamed for long-horizon failure do not survive the data. Store
churn does not separate the field (\S\ref{sec:res_execution}). Models that sign up for more promotions end richer, so event neglect explains nothing. Late-year
collapse holds only in part, growth dropping after the first quarter at a paired
$t = -3.95$ and the closing window flat rather than recovering.
Failure patterns spread across the whole field rather than concentrating in the
weakest models. Every model trips at least five rules,
Gemini 3.1 Pro all ten, and tripping fewer of them does not make a model richer. The cleanest profiles belong to the best bargainers, not the richest, so the field reads as strong in patches and coordinated nowhere (transcripts in Appendix~\ref{app:cases}).

%% file: tables/tab_4_2_leaderboard.tex
\begin{table}[t]
\centering
\caption{\small End-of-year leaderboard, 18 models, five episodes each, sorted by mean final assets within tier. Final assets carries the primary score, and the six columns after the spread rank one evaluation dimension each, $\mathrm{CSE}^+$ for negotiation quality, $\mathrm{BadSpend}\%$ for fraud avoidance, drawdown over peak for cash flow and solvency, yuan per tool call for operational efficiency, the controllable return rate for operations execution, and $\mathrm{AnchorRatio}$ for learning over the horizon, with the formulas in \S\ref{app:metrics_d1} through \S\ref{app:metrics_d6}. Arrows mark the direction of the better outcome, and a tinted cell holds the extreme of a scored column. Tool calls, turns, and bankrupt runs record what an episode spent rather than scoring it, and bankrupt episodes stay pooled into every mean, which is what the spread column beside
the assets reports. $^{\dagger}$GPT-5.5's spread is a population estimate,
\textyen 689k on the sample estimator. $^{\ddagger}$GPT-5.5's
$\mathrm{AnchorRatio}$ averages three episodes; its two bankrupt runs opening no repeat order with a supplier it had already dealt with.}
\label{tab:leaderboard}
\vspace{0.15cm}
\scriptsize
\renewcommand{\arraystretch}{0.94}
\setlength{\tabcolsep}{2.6pt}
\begin{adjustbox}{max width=\textwidth}
\begin{tabular}{lrrrrrrrrrrr}
\toprule
\textbf{Model} &
\makecell[br]{\textbf{Final assets}\\ \textyen k, mean} &
\makecell[br]{\textbf{std}\\ \textyen k} &
\makecell[br]{$\mathrm{CSE}^+$\\ $\uparrow$} &
\makecell[br]{$\mathrm{BadSpend}\%$\\ $\downarrow$} &
\makecell[br]{\textbf{Drawdown}\\ \textbf{/ peak} $\downarrow$} &
\makecell[br]{\textyen\ \textbf{per}\\ \textbf{tool call} $\uparrow$} &
\makecell[br]{\textbf{Controllable}\\ \textbf{return}, pp $\downarrow$} &
\makecell[br]{$\mathrm{AnchorRatio}$\\ $\downarrow$} &
\makecell[br]{\textbf{Tool}\\ \textbf{calls}} &
\textbf{Turns} &
\makecell[br]{\textbf{Bankrupt}\\ \textbf{runs}} \\
\midrule
\multicolumn{12}{l}{\emph{Proprietary}} \\
 GPT-5.6 Sol (max) & \cellcolor{bestcell}\textbf{1{,}431} & 314 & 0.672 & 18.48 & 0.278 & 363 & 4.46 & 1.217 & 3{,}668 & 1{,}367 & 0/5 \\
 Fable5 (max) & 805 & 188 & 0.772 & 3.46 & 0.141 & \cellcolor{bestcell}\textbf{479} & 0.22 & 1.573 & 1{,}469 & 945 & 0/5 \\
 GPT-5.5 & 702 & 616$^{\dagger}$ & 0.700 & 16.59 & 0.591 & 192 & 0.90 & 1.043$^{\ddagger}$ & 3{,}143 & 1{,}306 & 2/5 \\
 Claude Opus 4.8 (max) & 498 & 231 & 0.662 & 5.41 & 0.130 & 266 & 0.20 & 1.309 & 1{,}497 & 812 & 0/5 \\
 Claude Opus 4.7 (max) & 259 & 111 & \cellcolor{bestcell}\textbf{0.811} & \cellcolor{bestcell}\textbf{0.12} & 0.189 & 156 & 1.75 & 1.421 & 1{,}023 & 432 & 0/5 \\
 Claude Opus 4.6 (max) & 258 & 266 & 0.649 & 14.98 & 0.587 & 129 & 0.76 & 1.268 & 1{,}221 & 788 & 2/5 \\
 Gemini 3.5 Flash & 190 & 79 & 0.777 & 2.13 & 0.450 & 36 & 0.38 & 0.918 & 2{,}508 & 2{,}628 & 0/5 \\
 Gemini 3.1 Pro & 130 & 130 & 0.660 & 12.58 & 0.688 & 19 & 2.03 & 1.376 & 1{,}605 & 1{,}141 & 2/5 \\
\midrule
\multicolumn{12}{l}{\emph{Open-weight}} \\
 Qwen3.8-Max-Preview & 416 & 111 & 0.713 & 6.13 & 0.242 & 173 & 0.38 & \cellcolor{bestcell}\textbf{0.834} & 1{,}826 & 962 & 0/5 \\
 GLM 5.2 (high) & 301 & 124 & 0.693 & 1.99 & \cellcolor{bestcell}\textbf{0.127} & 137 & 0.59 & 1.434 & 1{,}467 & 672 & 0/5 \\
 Kimi K3 & 265 & 110 & 0.632 & 5.87 & 0.247 & 123 & 2.01 & 1.507 & 1{,}334 & 878 & 0/5 \\
 GLM 5.1 & 226 & 192 & 0.662 & 6.49 & 0.231 & 94 & 0.09 & 1.557 & 1{,}333 & 852 & 0/5 \\
 \makecell[l]{DeepSeek-V4-Pro-Preview\\ (max)} & 190 & 100 & 0.654 & 8.99 & 0.166 & 68 & 1.54 & 1.235 & 1{,}337 & 678 & 0/5 \\
 Qwen3.7-Max & 165 & 134 & 0.616 & 9.85 & 0.433 & 59 & \cellcolor{bestcell}\textbf{$-$0.05} & 1.433 & 1{,}108 & 782 & 0/5 \\
 GLM 5.2 (max) & 115 & 71 & 0.632 & 19.68 & 0.360 & 10 & 2.24 & 1.362 & 1{,}430 & 931 & 0/5 \\
 Kimi K2.6 & 70 & 14 & 0.596 & 17.22 & 0.377 & $-$27 & 0.42 & 1.548 & 1{,}087 & 1{,}017 & 0/5 \\
 Qwen3.6-Plus & 47 & 28 & 0.625 & 10.68 & 0.583 & $-$40 & 3.03 & 1.333 & 1{,}325 & 1{,}023 & 0/5 \\
 Qwen3.5-Plus & 1.1 & 11 & 0.625 & 20.11 & 0.979 & $-$92 & 5.88 & 1.860 & 1{,}076 & 800 & 4/5 \\
\bottomrule
\end{tabular}
\end{adjustbox}
\end{table}

%% file: sections/5_conclusion.tex
\section{Conclusion}\label{sec:conclusion}

In this paper, we introduce E-Commerce Bench, a long-horizon benchmark in which an agent runs an online retail business for a simulated 365-day year. The benchmark is built in four parts. An agent loop drives one model call at a time and keeps a year of tool traffic inside the context window, with a persistent memory beside it. Eighteen tools cover the merchant's job, from market research and supplier chat to pricing, shipping, and cash withdrawal. A deterministic environment decides what customers buy and how suppliers bargain, computing over products, suppliers, and a calendar taken from a real e-commerce platform. Purchases and returns follow a fixed demand model, and a deterministic negotiation kernel makes every supplier price, concession, and accept/reject decision while an LLM only turns those decisions into dialogue. Eighteen frontier and open-weight models each ran the year five times, scored on end-of-year total assets and on six capabilities behind that number (\S\ref{sec:eval}). No model is strong on all seven. GPT-5.6 Sol earns the most, turning the \textyen 100{,}000 stake into roughly 14 times as much, yet it ranks 16th of 18 on fraud avoidance and earns less per tool call than Fable5. Claude Opus 4.7 bargains best and avoids fraud best while finishing mid-pack on money, and Qwen3.8-Max-Preview leads the open-weight models at 4.2 times the stake. Four models go bankrupt in part of their runs, two GPT-5.5 episodes as early as day 17, before any revenue had settled (\S\ref{sec:res_cashflow}).
% Looking ahead, the two design principles behind E-Commerce Bench extend well beyond retail. First, delegating counterpart decisions to a deterministic kernel while using the LLM solely for verbalization preserves realistic interactions without introducing evaluation noise. Any long-horizon benchmark involving extended multi-party interactions faces this same tension, whether in procurement, sales, hiring, or customer service. Second, multi-dimensional evaluation is indispensable, as agents in long-horizon tasks often fail in diverse and unpredictable bad patterns. Together, these two principles pave the way for scaling benchmarks to longer horizons while maintaining reliability.
Beyond the retail domain, E-Commerce Bench offers a foundational blueprint for next-generation agent evaluation. By proving that \textit{deterministic kernel control} eliminates evaluation noise in complex multi-party interactions, and that \textit{multi-dimensional diagnostics} are indispensable for catching diverse failure modes, we provide the methodology needed to push agent benchmarks into ever-longer horizons.

%% file: sections/appendix.tex
\addtocontents{toc}{\protect\setcounter{tocdepth}{1}}
\appendix

\section{Agent Harness and Interface}\label{app:harness}

A model sees the environment only through tool schemas and tool replies. Nothing else is
sent to it, so every notice the environment has to deliver rides inside a reply. The price
list comes first, then the format of a reply, then the eviction algorithm that holds a year
of transcript inside a 128{,}000-token window.
Sections~\ref{sec:agent_loop} and~\ref{sec:tools} give the outline, and
Table~\ref{tab:tools} the price list. Prompts are reproduced separately in
Appendix~\ref{app:prompts}, and the eviction schematic with Section~\ref{sec:context_mgmt}.

\input{tables/tab_3_3_tools}

\subsection{The Interface Contract}\label{app:harness_io}

Every tool call comes back as a single string. Structured results arrive as JSON, plain text
unchanged. Extra notices sit inside that JSON rather than after its closing brace, so the
reply still parses. Day-scoped notices attach to the day-advance result when the batch holds
one and to the last response otherwise, so news is never blamed on an unrelated earlier call.

The morning pass reports far less to the agent than it computes. Of the 17 fields it
assembles, at most three reach the agent, the date, the day index and the news feed. Across
the 90 archived episodes the channel carried 8{,}301 items, 6{,}722 of them deliveries at
one aggregated line per day.

Listing~\ref{lst:tool_result} shows two real tool messages, one at a day boundary and one
carrying the solvency warning that preceded a bankruptcy.

\begin{lstlisting}[breaklines=true,basicstyle=\ttfamily\scriptsize,captionpos=b,
  extendedchars=true,
  literate={¥}{{\textyen}}1 {—}{{\textemdash}}1,
  caption={Two tool messages exactly as they entered the request, from GPT-5.6 Sol run 0
  on day 8 and from GPT-5.5 run 1 on 2026-01-03. The first carries the day notice, the
  second a low-balance reminder, and both close with the token gauge that sits outside
  the JSON. The second run went bankrupt.},
  label={lst:tool_result}]
{"current_time": "2026-01-09T08:00:00", "day": 8, "system_notifications": [{"date": "2026-01-09", "day": 8, "news": [{"type": "delivery", "content": "Deliveries arrived:\n  - 20x DreamMist LED Ceiling Light for Living Room and Bedroom, Mod from Crystal Trading"}]}]}
<system_warning>Token usage: 108593/128000 tokens (84%); 19407 remaining</system_warning>

{"pending_shipments": [], "count": 0, "note": "Ship with ship_orders (speed: fast/standard/slow). Faster ship costs more but reduces returns; slow is cheap but raises returns. Unshipped orders cancel after the deadline (lost sale + reputation hit).", "current_time": "2026-01-03 16:40", "balance_reminder": "Your current balance (¥10.00) is below the daily operating cost (¥60.00). If your bank account stays negative for 10 consecutive days, you will go bankrupt."}
<system_warning>Token usage: 93856/128000 tokens (73%); 34144 remaining</system_warning>
\end{lstlisting}

\subsection{Loop Control and Context Editing}\label{app:harness_loop}

Token counts come from a local tokenizer rather than the provider's own report, and cover
the text, the tool calls and the reasoning trace of each message. Opaque artifacts such as
replayed encrypted reasoning items are excluded, as are the tool schemas held outside the budget
(\S\ref{sec:context_mgmt}). The system prompt quotes the model the 120{,}000-token
threshold, the 60{,}000-token release floor and the two protected groups.
One pass proceeds as follows.

\begin{enumerate}[leftmargin=1.4em,itemsep=1.5pt,topsep=3pt]
\item Sum the counts of every message not already marked cleared. Below that ceiling the
  pass emits the usage string and frees nothing.
\item Scan for groups from the third message onward, leaving the system message and the
  first user turn permanently out of reach. A group opens at an assistant message carrying
  tool calls and closes after the last consecutive tool response behind it.
\item Set aside the two most recent groups not yet cleared. With two or fewer active groups
  the pass clears nothing at all, so a history made of a few very large groups can sit above
  the ceiling indefinitely.
\item Take the release target as the larger of that floor and the current overshoot past
  the ceiling.
\item Walk the clearable groups oldest first, marking each cleared and accumulating its
  count, and stop at the first group that meets the target.
\item Rebuild the request from the surviving messages, skipping cleared entries entirely
  (\S\ref{sec:context_mgmt}).
\end{enumerate}

\begin{figure}[t]
  \centering
  \includegraphics[width=\linewidth]{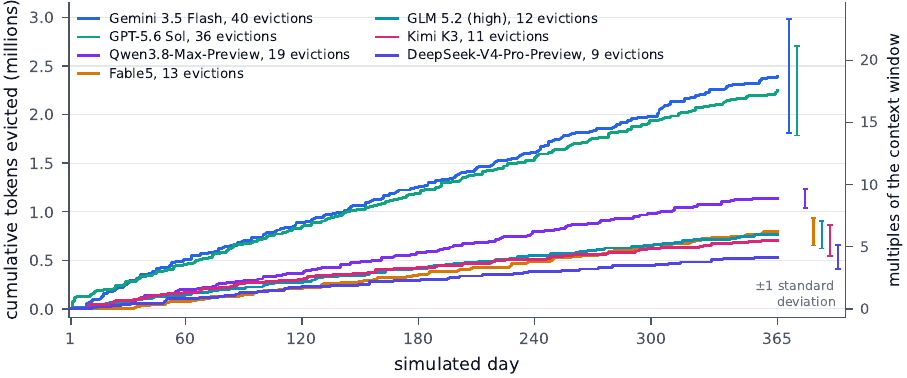}
  \caption{How often the context editor fired, over the 90 evaluated episodes. Each event is one eviction pass. A model that never reaches the ceiling evicts nothing.}
  \label{fig:context_pressure}
\end{figure}

Whether a pass ever fires depends on how a model spends the window, and
Figure~\ref{fig:context_pressure} counts the passes of each of the 90 episodes.

Two kinds of message can never be evicted, an assistant turn of plain text and the idle
warning. Neither opens a group nor falls inside one, so both accumulate for the whole
episode. Clearing whole groups is what keeps the transcript valid, since every provider
protocol here rejects a tool response separated from its call.

\section{Data Layer}\label{app:data}

Every quantity the released data layer holds is recorded here, so a reader can check
any number the report quotes against its table. How a supplier behaves once bargaining
starts belongs to Appendix~\ref{app:nego} instead. The catalog carries 6{,}886 SKUs, 60 category parameter
rows, 576 suppliers and 12 store types, and the calendar 10 market events and 8 promotions,
every one of them frozen, so the world is identical for every model and every run.

\subsection{Category Parameters}\label{app:data_categories}

The released columns that decide margin are the ones that never reach the agent, namely the
three per-category price ratios, the elasticity family and its parameter and the two ends of
the return band. A SKU's own natural rate becomes readable only through
\texttt{trace\_return\_sources} once stock has moved (\S\ref{sec:data_hidden}).
Each category belongs to exactly one store type, and Table~\ref{tab:categories} groups all
60 that way, each group heading carrying that type's difficulty tier and daily operating cost
while Table~\ref{tab:store_types} carries its return band and cost-floor ratio. Setup fee and
commission are identical across the 12, leaving the seasonal curve and the demand terms of
\S\ref{sec:economy} as the only differences the two tables do not show.
Category rows set bands and per-SKU rows sit inside them. Reference price and natural return
rate vary by SKU within the category band while base demand does not, and one elasticity
family serves a whole category. The wholesale ratio is pinned to the cost-floor ratio
$c_j/v_j$ in all 60 rows (\S\ref{app:nego_grounding}).

\input{tables/tab_app_categories}

\begin{figure}[t]
  \centering
  \includegraphics[width=\linewidth]{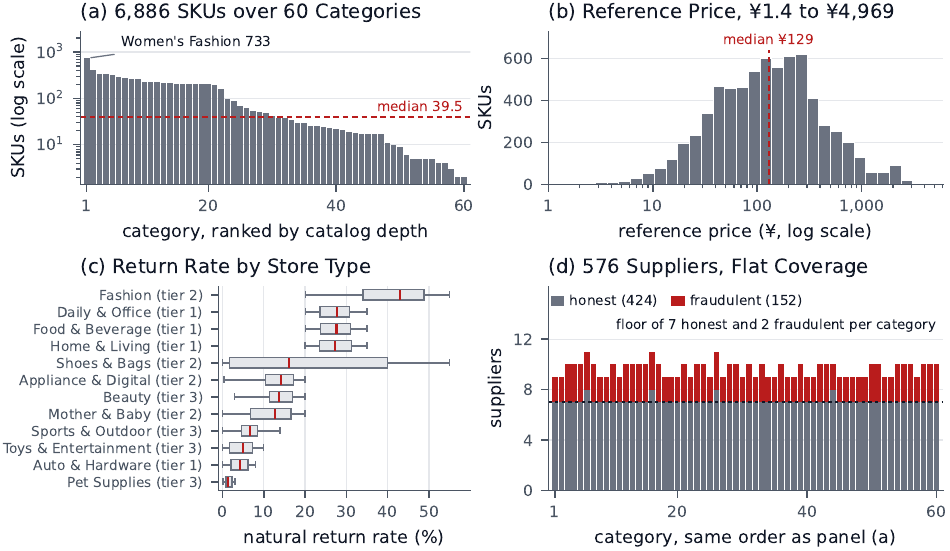}
  \caption{The released data. (a) Products per category. (b) Reference prices. (c) Natural return rate by store type, T the difficulty tier. (d) Suppliers per category against catalog depth.}
  \label{fig:catalog}
\end{figure}

\subsection{Store Types, Seasonality, and Physical Cost}\label{app:data_stores}

\begin{figure}[t]
  \centering
  \includegraphics[width=\linewidth]{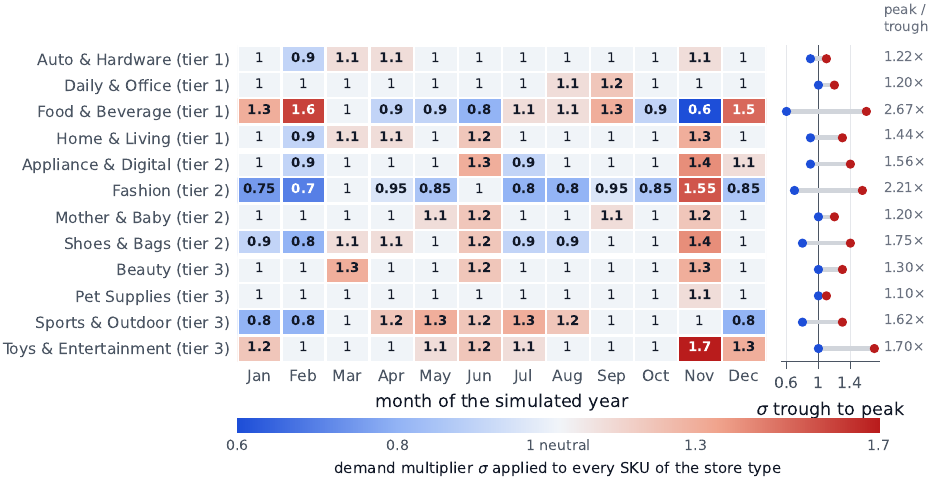}
  \caption{The monthly seasonality multiplier for all 12 store types, rows grouped by difficulty tier. The right panel gives each type's trough, peak and peak-to-trough ratio.}
  \label{fig:seasonality}
\end{figure}

A merchant fixes the store type once and cannot tune it afterwards without paying the setup
fee again. Table~\ref{tab:store_types} names all 12 with the difficulty tier, operating cost
and category span each carries.
\input{tables/tab_3_4_store_types}
Figure~\ref{fig:seasonality} gives the seasonal grid in full, one multiplier per store
type and calendar month. Amplitude, not level, separates the types. Nine of the 12 attain
their annual maximum in November, inside the same weeks as the calendar's longest promotion
window (Table~\ref{tab:events}), while Food \& Beverage inverts the pattern and peaks in
February, so a portfolio timed for one season pays for it in the other. Four types stay
within $\pm 0.2$ of neutral all year and can be run without timing at all. Disclosure is
exact rather than qualitative, the index \texttt{market\_search} returns being precisely
$100\,\sigma$.
Parcel size turns straight into cost, freight and daily storage both scaling with it while
the shipping speed the agent picks multiplies freight and the order's return rate together.
Appendix~\ref{app:econ} gives all four schedules with the rest of the cost model.

\subsection{Event and Promotion Calendar}\label{app:data_calendar}

An agent is warned about a promotion but not about an event. An event posts its news on its
first day and gives no lead time, while a promotion is announced up to seven days before its
first
window opens and \texttt{join\_promotion} accepts a target 30 days ahead, so inventory can be
positioned for a promotion and never for an event. The two calendars touch only where the
Guangdong factory fire of March 1 to 7 runs across the March 4 to 8 Spring Blossom Sale, a
supply shock rather than a demand one, which keeps every measured discount response
uncontaminated.

An event with no demand term bites only the purchase orders placed while its window is open,
since the lead time is fixed at placement and stock already in transit keeps the arrival date
it was given. Table~\ref{tab:events} prints the calendar in full, every date, phase,
per-store-type demand multiplier and delivery-delay factor.

\begin{figure}[t]
  \centering
  \includegraphics[width=\linewidth]{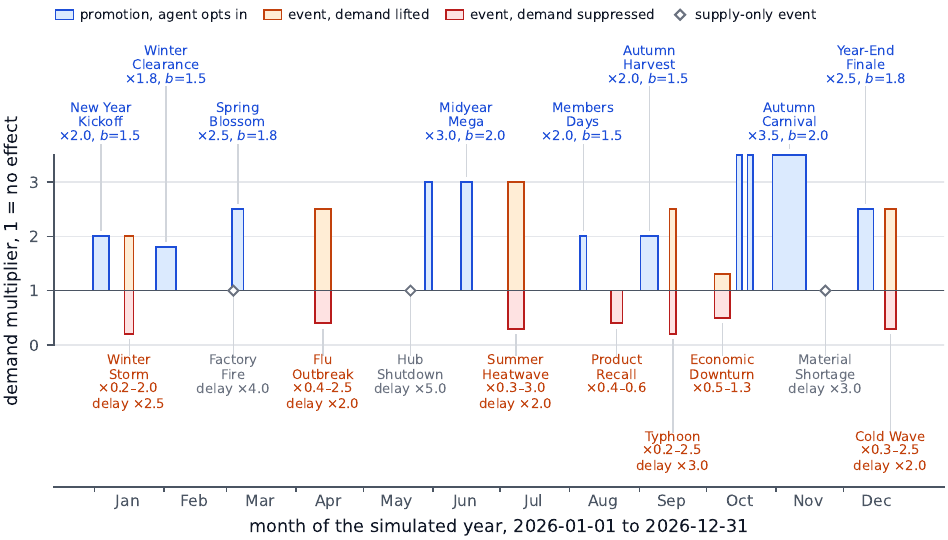}
  \caption{The 2026 calendar. Eight promotions the agent can join by choosing a discount, and ten events it cannot opt out of. Promotion bars show the multiplier a $30\%$ discount buys, event bars their extreme per-store-type multiplier.}
  \label{fig:calendar}
\end{figure}

\input{tables/tab_app_events}

\subsection{The Supplier List}\label{app:data_suppliers}

The supplier list is frozen, so the difficulty mix every model faces is a property of the
benchmark rather than a per-run draw. Each row carries an identity, an honest-or-fraudulent
label, a behavior template, a scam type where one applies, the single category served and
a bankruptcy threshold in orders, and counts land at 424 honest and 152 fraudulent over the
60 categories, 7 or 8 against 2 or 3 in each. A supplier's name carries no signal and its
email carries one (\S\ref{app:data_artifacts}), while what a template or a scam does at
the table belongs to Appendices~\ref{app:nego_presets} and~\ref{app:nego_fraud_catalog}.

\subsection{Properties a Reader Should Check}\label{app:data_artifacts}

Four properties of the released files look odd on inspection, and each was checked against
the code that reads them. Two carry consequences for a measurement.

\begin{enumerate}[leftmargin=1.35em,itemsep=2pt,topsep=3pt]
\item \textbf{Return-rate wording understates the numeric band.} Twenty of the 60 categories
carry a sentence reading low or very low against a numeric band reaching 20\% to 35\%, marked
in Table~\ref{tab:categories}. The sentence is what \texttt{market\_search} shows, so a
merchant who prices off the wording underestimates returns across categories holding 25.5\%
of all SKUs, and sees the gap only through \texttt{trace\_return\_sources} on stock already
owned.

\item \textbf{The supplier email leaks the fraud label.} \texttt{supplier\_search} returns
an email whose local part ends in the supplier identifier, and inside every one of the 60
categories the honest suppliers hold lower identifiers than the fraudulent ones, so an agent
sorting on that numeric suffix could screen fraudulent counterparts without sending a single
message. Section~\ref{sec:data_hidden} calls the supplier list free of any ranking that sorts honest
from fraudulent, which holds for the returned fields and their order while the suffix carries
the signal anyway. Whether any evaluated model exploited it is read in
\S\ref{app:res_funnel}, whose contact stage would show the effect.
\end{enumerate}

Two more are minor and neither changes a measurement. The released rent column reads
\textyen 45 in the four easy-tier rows against the \textyen 60 the environment charges from
the difficulty tier instead, so an analysis trusting the column understates their fixed burn
by a quarter. Raw Material Shortage carries a cost increase that nothing reads, so the
November shortage stretches lead times and leaves procurement prices untouched whatever its
news copy says.

The layer closes on what the agent can and cannot read, each quantity the world holds set
against what a tool will return about it (Table~\ref{tab:asymmetry}). Every row an agent
cannot read has to be inferred from outcomes, which is where the sourcing and pricing
measurements of Section~\ref{sec:experiment} get their difficulty.

\input{tables/tab_3_6_asymmetry}

\section{Economy Engine}\label{app:econ}

The rules \S\ref{sec:economy} describes in prose are written out here as equations, in the
order the simulation runs them. Notation follows
\S\ref{sec:bench}, with $i$ a store, $j$ a SKU, $t$ a simulated date, $\tau = \tau(i)$ the
store type and $c = c(j)$ the category. Prices are $v_j$ the reference price,
$c^{\mathrm{buy}}_j$ the price actually paid and $c_j^s$ a supplier's private floor.
Capital $\Pi$ is unit profit, not the price position $\pi$ of \S\ref{sec:nego_metrics}.
Constants outside the equations are collected in \S\ref{app:econ_constants}, except the
seasonal grid, which Appendix~\ref{app:data} carries.

\subsection{The Daily Settlement Order}\label{app:econ_tick}

\begin{figure}[tp]
  \centering
  \includegraphics[width=\linewidth]{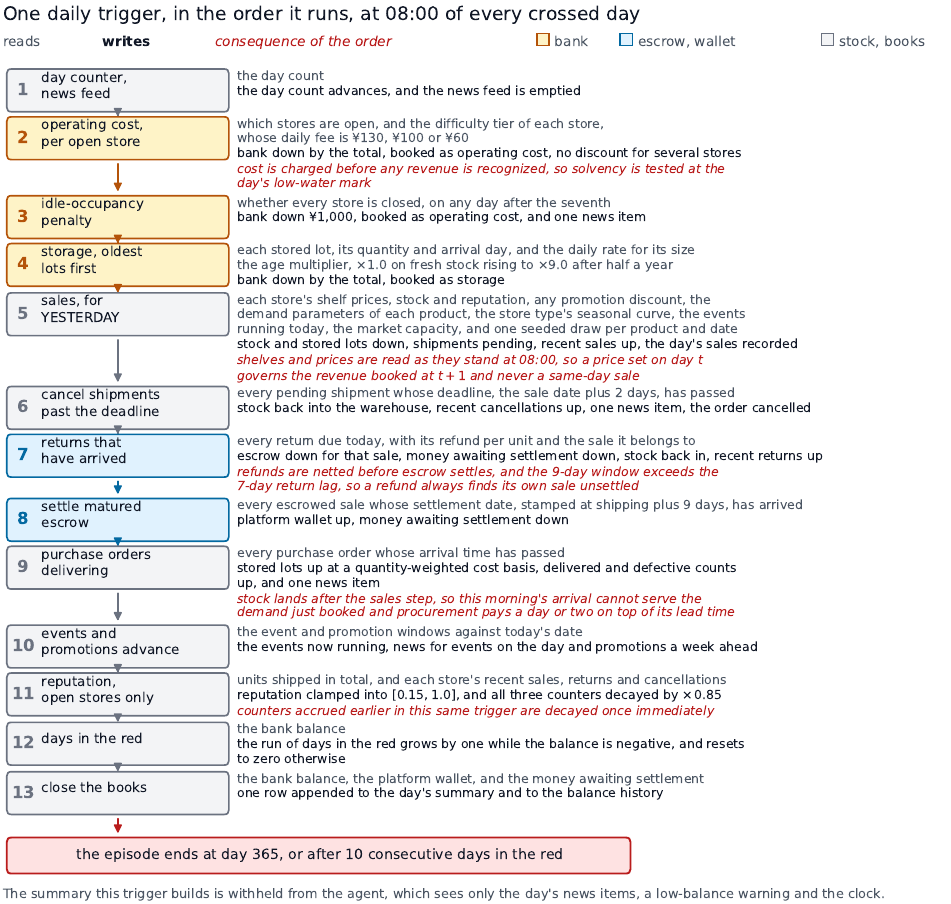}
  \caption{The thirteen settlement steps in the order the simulation runs them at each 08:00 crossing, with the state each step reads and writes. Fill colour marks which money bucket a step moves.}
  \label{fig:daily_tick}
\end{figure}

Until the clock crosses 08:00 nothing settles. Thirteen steps then run in an order nothing
can reconfigure, once for each boundary. Figure~\ref{fig:daily_tick} lists the steps with
the state each reads and writes, none of it shown to the agent (\S\ref{app:harness_io}).

Costs are charged before any revenue is recognized, so solvency is tested each morning at
the day's low-water mark. Sales for the \emph{previous} calendar date run against the
inventory and prices the agent left in place, leaving it in charge of a system lagged by
one day.

The episode ends once the year runs out or the unpaid streak reaches its limit
(\S\ref{app:econ_constants}). A finalization pass then discards every unshipped order as a
lost sale, processes every seeded return against its escrow batch, and drains all remaining
escrow into the platform wallet. A closing fire sale cannot outrun its own returns.

\subsection{Demand Model}\label{app:econ_demand}

Base demand starts from the category and is scaled twice, with $M^{\min}_c$ and
$M^{\max}_c$ the platform-wide monthly sales range of category $c$.
\begin{equation}
\label{eq:app_base}
\beta_{c\tau} \;=\; \frac{1}{30}\cdot\frac{M^{\min}_c + M^{\max}_c}{2}
  \;\cdot\; \lambda \;\cdot\; \psi_\tau ,
\qquad \lambda = 0.1 ,
\end{equation}
Platform volume becomes one store's day through $\lambda$, and $\psi_\tau$ calibrates the
store type from $0.2$ for Shoes \& Bags to $9.0$ for Food \& Beverage. Every SKU in a
category shares that $\beta_{c\tau}$.

The price factor takes one of four closed forms, keyed by category and evaluated at the
discounted price ratio $r_{ijt} = (1-\delta_{it})\,p_{ijt}/v_j$.
\begin{equation}
\label{eq:app_elasticity}
\phi_{f}(r; \eta) \;=\;
\begin{cases}
\max\!\big(0,\; 1 - \eta\,(r-1)\big) & f = \textsc{linear}, \quad \eta \in [2.185,\, 3.949],\ 11\ \text{categories},\\[2pt]
\exp\!\big(-\eta\,(r-1)\big) & f = \textsc{exponential}, \quad \eta \in [2.583,\, 4.944],\ 16,\\[2pt]
r^{-\eta} & f = \textsc{constant-elasticity}, \quad \eta \in [1.352,\, 2.898],\ 17,\\[2pt]
\max\!\big(0,\; 1 - \eta\,(r-1)^2\big) & f = \textsc{quadratic}, \quad \eta \in [2.795,\, 5.779],\ 16 .
\end{cases}
\end{equation}
Only the quadratic branch is non-monotone, so a deep discount can destroy demand rather
than merely cost margin.

The remaining multipliers of Equation~\eqref{eq:demand} are each read once per store-day,
weekends contributing $w_t = 1.3$. A store joining promotion $g$ at discount $\delta_{it}$
earns $m_{it} = 1 + (M_g - 1)\min(1, \delta_{it}/0.30)$ and scales $\eta_c$ by $b_{it}$.
Seasonality $\sigma_{\tau,t}$ reads the store type's month out of the grid
Appendix~\ref{app:data} tabulates, active events compound into $\varepsilon_{\tau,t}$, and
reputation $\rho_{it}$ enters linearly.

Two saturation terms then trim that demand, the per-category factor $\gamma_{ijt}$ of
Equation~\eqref{eq:demand} and the store-wide term beside it. The per-category ceiling
$\kappa_c$ takes a fraction $\chi_\tau$ of the store type's own daily capacity
$\kappa_\tau$. Capacity runs from $13.959$ units for Appliance \& Digital to
$19{,}862.16$ for Auto \& Hardware, and a shorter category list raises the fraction, which
reaches 1 for a store type allowed a single category. At that point the per-category term is
dropped rather than relaxed, so $\gamma_{ijt} = 1$ and only the store-wide term bites.
\begin{equation}
\label{eq:app_saturation}
\chi_\tau = \min\!\Big(1,\ \max\big(0.35,\ 1.2/n_\tau\big)\Big), \qquad
\kappa_c = \chi_\tau\,\kappa_\tau .
\end{equation}
Both terms take the Michaelis-Menten form $\kappa/(\kappa+D)$, the first against one
category's stacked demand and the second against the store's post-cap total
$\hat{D}_{it}$.

Rounding the last fractional unit up or down then draws on a number that call order cannot
change.
\begin{equation}
\label{eq:app_round}
u_{jt} \sim \mathcal{U}(0,1) \quad\text{drawn from the environment seed hashed with the SKU
and the date}.
\end{equation}
The cap $x_{ijt}$ of Equation~\eqref{eq:demand} is on-shelf inventory, so a stockout
truncates demand with no signal to the agent.

\subsection{Reputation Law}\label{app:econ_reputation}

\begin{figure}[t]
  \centering
  \includegraphics[width=\linewidth]{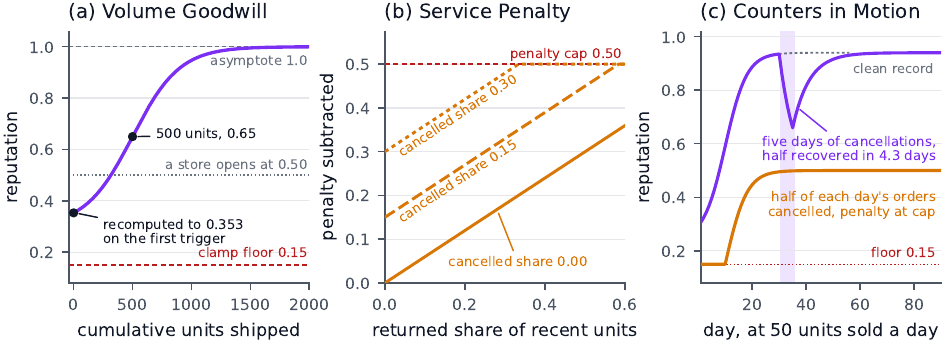}
  \caption{How reputation moves. (a) The goodwill term against units shipped. (b) The penalty term against the returned share of recent units, at three cancellation shares. (c) A service shock decays while a chronic failure rate does not.}
  \label{fig:reputation}
\end{figure}

Every settlement rebuilds each open store's standing from scratch. Writing $S_{it}$ for
units shipped to date and $R_{it}, C_{it}, N_{it}$ for the rolling counters of returned,
cancelled and sold units,
\begin{equation}
\label{eq:app_reputation}
\rho_{it} = \mathrm{clip}\Big(
  \underbrace{0.3 + \frac{0.7}{1 + e^{-(S_{it}-500)/200}}}_{\text{volume goodwill}}
  \;-\;
  \underbrace{\min\!\Big(0.5,\ 0.6\,\frac{R_{it}}{\max(1, N_{it})}
    + 1.0\,\frac{C_{it}}{\max(1, N_{it})}\Big)}_{\text{service penalty}},
  \;\; 0.15,\ 1.0\Big),
\end{equation}
after which all three counters are multiplied by $0.85$. Goodwill credits shipping rather
than selling, so an unshipped order earns nothing and still takes the cancellation weight.
Because goodwill starts from zero shipped units, a store that opens at $0.5$ is recomputed
to $0.353$ at its first settlement. The same counters weigh a cancelled unit as $1.67$
returned units, pricing a missed deadline above a change of mind
(Figure~\ref{fig:reputation}).

\subsection{Returns}\label{app:econ_returns}

An item's return rate composes four channels in a fixed order, three at sale time and one
at the agent's choice of shipping speed. Write $\theta^{\mathrm{nat}}_j$ for the SKU's
natural rate and $\alpha_j$ for the delivered share that came from a quality-downgrading
supplier. The price curve $\zeta$ is piecewise linear through the knees $(0.8, 0.85)$,
$(1.0, 1.00)$, $(1.3, 1.50)$ and $(1.8, 2.20)$, held flat outside them.
\begin{equation}
\label{eq:app_returns}
\begin{aligned}
\theta^{\mathrm{def}}_j &= \min\!\big(0.95,\ \max(0.40,\ 2\,\theta^{\mathrm{nat}}_j)\big),
&\qquad
\theta^{(1)}_j &= (1-\alpha_j)\,\theta^{\mathrm{nat}}_j + \alpha_j\,\theta^{\mathrm{def}}_j,
\\
\theta^{(2)}_j &= \min\!\big(0.95,\ \theta^{(1)}_j\,\zeta(r_{ijt})\big),
&\qquad
\theta_j &= \min\!\big(0.95,\ \theta^{(2)}_j\cdot s\big),
\quad s \in \{0.75,\, 1.00,\, 1.30\},
\end{aligned}
\end{equation}
The factor $s$ is the return multiplier for fast, standard and slow shipping. Inventory is
pooled by SKU and $\alpha_j$ is a ratio over cumulative deliveries, so a mixed lot leaves
the elevated return rate untraceable.

Whether a unit comes back is decided the day it ships, not the day it arrives. A shipment
of $q$ units draws $q$ seeded trials at rate $\theta_j$ and one arrival day uniformly from
$\{3, \dots, 7\}$, so a shipment's returns land together. Each arriving unit refunds the
full retail price out of the originating escrow batch and adds to the rolling return
counter of Equation~\eqref{eq:app_reputation}. Returned stock re-enters the warehouse as a
fresh lot dated that day, resetting its storage age. Freight and commission never come
back.

\subsection{Cost Model}\label{app:econ_costs}

Six channels debit the bank account the instant they are incurred. The platform's
commission is a seventh, withheld from revenue before it reaches escrow.
\begin{itemize}[leftmargin=1.3em,itemsep=1pt,topsep=2pt]
\item \textbf{Setup}, \textyen 500 per store opened, charged again on every reopening.
\item \textbf{Operating cost}, \textyen 130, \textyen 100 or \textyen 60 per open store
  per day ($o_\tau$) by the store type's difficulty tier.
\item \textbf{Idle occupancy}, \textyen 1{,}000 a day whenever no store is open after
  the seventh day.
\item \textbf{Storage}, $h_j \cdot a(\text{age})$ per unit per day, where $h_j$ is
  $0.05$, $0.15$, $0.50$ or $1.50$ by product size and $a$ steps through
  $1.0, 1.4, 2.2, 4.0, 6.0, 9.0$ at ages $0, 21, 45, 90, 135, 180$ days. Lots are FIFO, and
  stock listed in a store keeps accruing.
\item \textbf{Freight}, $f_j$, a size cost of $0.5$, $1.5$, $3.0$ or $6.0$ times a speed
  multiplier of $2.0$, $1.0$ or $0.5$ for fast, standard and slow, charged per unit at ship
  time.
\item \textbf{Commission}, $\gamma = 2\%$ of gross revenue, netted at sale rather than
  debited, and kept by the platform on refunded sales (\S\ref{app:econ_returns}).
\item \textbf{Procurement}, the full purchase order at the moment a negotiation closes
  (\S\ref{sec:nego_engine}). A membership fee rides inside the same debit and almost never
  fires (\S\ref{app:nego_fraud_catalog}).
\end{itemize}
Liquidation is the one inbound channel besides sales, crediting 10\% of the
quantity-weighted purchase cost and removing the stock. Closing a store without
liquidating leaves storage accruing.

\subsection{Unit Profit and Break-Even}\label{app:econ_unit_profit}

\begin{figure}[t]
  \centering
  \includegraphics[width=\linewidth]{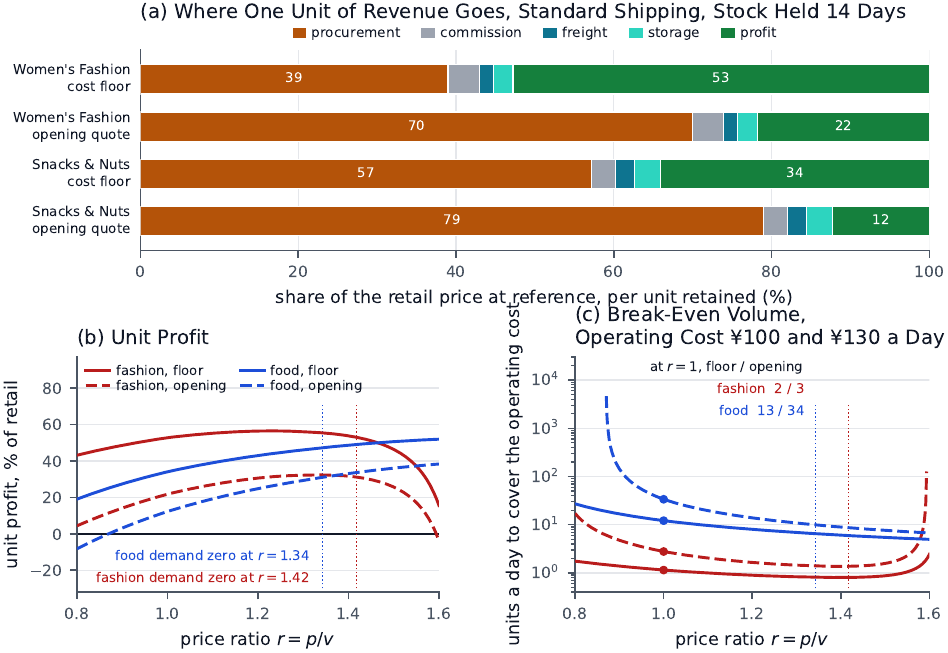}
  \caption{Where a unit's money goes. (a) Each component as a share of retail price, for a median-priced fashion SKU beside a snacks SKU. (b) Unit profit against the price ratio, with dotted lines where demand reaches zero. (c) Units a store must sell each day to cover its operating cost.}
  \label{fig:unit_profit}
\end{figure}

Composing \S\ref{app:econ_returns} and \S\ref{app:econ_costs} gives the contribution of
one unit a customer finally keeps. Because a returned unit comes back as stock, each
retained unit funds $1/(1-\theta_j)$ shipping attempts and as many storage cycles. A cycle
covers the $T_h$ days a unit waits in the warehouse before selling.
\begin{equation}
\label{eq:app_unit_profit}
\Pi_j(r) \;=\; \frac{\big(1 - \gamma - \theta_j(r)\big)\,r\,v_j
  \;-\; f_j \;-\; h_j\,T_h}{1 - \theta_j(r)} \;-\; c^{\mathrm{buy}}_j ,
\qquad
n^\star_j(r) \;=\; \frac{o_\tau}{\Pi_j(r)} .
\end{equation}
Break-even volume $n^\star$ counts the units a store must retain daily to cover its tier's
operating cost. Figure~\ref{fig:unit_profit}a splits retail price into the five shares that
add to one.

Procurement is the largest cost line, and the price a negotiation closes at moves unit
profit further than any pricing choice does. Two median-reference SKUs carry the identity
in Figure~\ref{fig:unit_profit}, one high-return and one thin-margin. The Women's Fashion
SKU sits in a medium-tier store, returns 49.2\% of units naturally, and is quoted
$0.70\,v_j$ against a floor of $c_j^s = 0.39\,v_j$. At the floor price, standard shipping
and $T_h = 14$ days, it earns \textyen 87.29 per retained unit and needs 2 units a day to
cover operations. The Snacks \& Nuts comparison in a hard-tier store needs 13. Closing
at the floor rather than accepting the opening quote multiplies unit profit by $2.4$ and
$2.8$.

\subsection{Constant Reference}\label{app:econ_constants}

Every constant named above sits beside its equation or cost channel. The remainder are
environment-wide. The opening stake is \textyen 100{,}000, at most 4 stores may be open at
once, and the year runs 365 days from 2026-01-01 with working hours 08:00 to 18:00. Orders
carry a two-day ship deadline against a nine-day escrow window, and bankruptcy follows 10
consecutive days with a negative bank balance. Environment seed
20260122 is shared by every run of every model. The two per-store-type parameters
$\kappa_\tau$ and $\psi_\tau$ are given as ranges in \S\ref{app:econ_demand}.

\section{Deterministic Negotiation Kernel}\label{app:nego}

The supplier policy \S\ref{sec:nego_twolayer} describes in outline is specified here, an independent implementation of the counterpart design of \citet{zhang2026terms}.
Everything follows from four decision functions and the five scam scripts layered on top of
them. Metric definitions live in
\S\ref{app:nego_metrics}, verbatim prompts in \S\ref{app:prompts}.

\subsection{Kernel Decision Functions}\label{app:nego_kernel}

A kernel instance is one supplier's bargaining state for one SKU, and every decision it
makes is a function of that state. The state holds a latent type
$t_B=(r_B,\kappa_B,\eta_B)$, an opening harshness $d_0$, the deadline scale $K=10$, a price
frame, seeded random streams and both offer
histories. Reservation price $r_B$ comes from the data layer (\S\ref{app:nego_grounding}),
while urgency $\kappa_B\in[0,1]$ and stance
$\eta_B\in\{\text{conciliatory},\text{neutral},\text{aggressive}\}$ come from the supplier's
template (\S\ref{app:nego_presets}). Indicators $\mathbf{1}_{\mathrm{a}}$,
$\mathbf{1}_{\mathrm{c}}$ mark the outer two stances. Every quantity below is normalized by
the width of the price frame,
\begin{equation}
\label{eq:k_frame}
p_{\max}=1.5\,w_j,
\qquad
\mathcal{R}=\max\!\left(1,\,p_{\max}-p_{\min}\right),
\qquad
\bar{\delta}_k=\frac{p_k-r_B}{\mathcal{R}},
\end{equation}
where $w_j$ is SKU $j$'s nominal wholesale price and $p_k$ the agent's offer in round $k$.
The lower bound $p_{\min}$ sits at $0$ for an honest supplier and at $r_B$ for a fraudulent
one, whose frame is therefore narrower. Favorability $\bar{\delta}_k$ measures the
seller's margin over its own floor.

Three features of the agent's offer sequence $(p^A_1,\dots,p^A_m)$ drive the
reciprocal terms, computed over the last three consecutive differences,
$D_m=\left(p^A_i-p^A_{i-1}\right)_{i=\max(2,\,m-2)}^{m}$,
\begin{equation}
\label{eq:k_hist}
\bar{c}_m=\frac{1}{|D_m|}\sum_{d\in D_m}\frac{\max(0,d)}{\mathcal{R}},
\qquad
\bar{s}_m=\frac{1}{|D_m|}\sum_{d\in D_m}\frac{d}{\mathcal{R}},
\qquad
g_m=\mathbf{1}\!\left[\frac{\max(0,d_m)}{\mathcal{R}}<\tau_{\mathrm{rigid}}\right],
\end{equation}
all three defined as $0$ while $m<2$, with $\tau_{\mathrm{rigid}}=0.10$. Concession
magnitude $\bar{c}_m$ counts only movement toward the seller, signed speed $\bar{s}_m$
also counts retreat, and the rigidity flag $g_m$ marks a latest step under a tenth of the
frame.

The kernel resolves acceptance first, walk-away second, and a counter-offer only if
neither fired. Acceptance follows a logistic draw in the manner of
\citet{baarslag2014acceptance},
\begin{equation}
\label{eq:k_accept}
a_k=\mathbf{1}\!\left[\bar{\delta}_k\ge0\right]\cdot
\sigma\!\left(\alpha\,\bar{\delta}_k+\beta\,\kappa_B
-\gamma\left(1-\sqrt{k/K}\right)+\rho_f\,\bar{s}_m+\xi_f\,g_m\right),
\qquad
\alpha=6,\;\beta=1,\;\gamma=2,
\end{equation}
where $\sigma$ is the logistic function and $(\rho_f,\xi_f)$ are the template's reciprocity
coefficients. Since $\rho_f\le0$ in every template of Table~\ref{tab:presets}, conceding
fast never improves the agent's own chance of acceptance.

\begin{figure}[t]
  \centering
  \includegraphics[width=\linewidth]{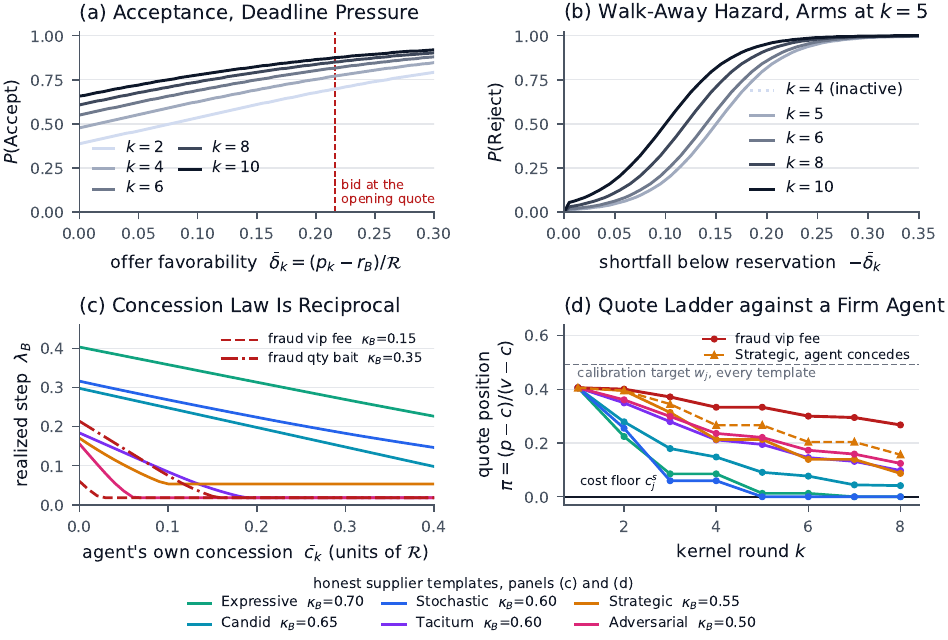}
  \caption{The four decision functions of the Negotiation Kernel, swept on one supplier and SKU. (a) Acceptance and (b) walk-away against the agent's offer. (c) The concession the kernel returns for a given concession from the agent. (d) The quote ladder it walks down.}
  \label{fig:kernel_functions}
\end{figure}

Walking away needs an offer below the floor, and only from the midpoint of the
deadline onward,
\begin{equation}
\label{eq:k_walk}
\omega_k=\mathbf{1}\!\left[\bar{\delta}_k<0\right]\,
\mathbf{1}\!\left[k\ge k_{\mathrm{walk}}\right]\cdot
\sigma\!\left(\phi_0+\phi_\Delta\left(-\bar{\delta}_k\right)
+\phi_t\,\frac{k-k_{\mathrm{walk}}}{K-k_{\mathrm{walk}}}\right),
\qquad
\phi_0=-4.5,\;\phi_\Delta=30,\;\phi_t=1.5,
\end{equation}
with $k_{\mathrm{walk}}=\lceil K/2\rceil=5$, both draws swept in
Figure~\ref{fig:kernel_functions}. An agent bidding above the floor is never abandoned, so
impasse on an honest session is almost always the agent's own rejection
(\S\ref{sec:nego_metrics}).

Counter-offers close a share of the gap to the floor, and that share is
reciprocal in the tradition of \citet{faratin1998negotiation},
\begin{equation}
\label{eq:k_conc}
\lambda_B=\mathrm{clip}\!\left(\lambda_0+\lambda_1\kappa_B
-\lambda_{2,f}\,\bar{c}_m
-\lambda_3\mathbf{1}_{\mathrm{a}}+\lambda_4\mathbf{1}_{\mathrm{c}},\;0,\;1\right),
\end{equation}
\begin{equation}
\label{eq:k_quote}
p^B_k=\mathrm{clip}\!\left(p^B_{k-1}-\lambda_B\left(p^B_{k-1}-r_B\right)
+\varepsilon,\;r_B,\;p^B_{k-1}\right),
\qquad
\varepsilon\sim\mathcal{N}\!\left(0,\left(\sigma_p\mathcal{R}\right)^2\right),
\end{equation}
with $\lambda_0=0.12$, $\lambda_1=0.28$ and $\lambda_3=\lambda_4=0.10$. The reciprocity
weight $\lambda_{2,f}$ and the noise scale $\sigma_p$ come from the template at its own
stance (Table~\ref{tab:presets}). Clipping keeps the quote path monotone non-increasing and never
below the floor, so a transacted price certifies $r_B\le p_z$.

The opening quote applies a stance-adjusted harshness,
\begin{equation}
\label{eq:k_open}
\begin{aligned}
\varphi&=\mathrm{clip}\!\left(1-\omega_\kappa\kappa_B
+\omega_\eta\mathbf{1}_{\mathrm{a}}-\omega'_\eta\mathbf{1}_{\mathrm{c}},\;0.5,\;1.5\right),\\
p^B_1&=\mathrm{clip}\!\left(r_B+d_0\,\varphi\left(p_{\max}-r_B\right)
+\varepsilon_0,\;r_B,\;p_{\max}\right),
\end{aligned}
\end{equation}
with $\omega_\kappa=0.3$, $\omega_\eta=\omega'_\eta=0.15$ and
$\varepsilon_0\sim\mathcal{N}\!\left(0,(0.02\,\mathcal{R})^2\right)$, neither clip ever
binding.

Two latent cues accompany every action. Posture draws from a softmax over three logits.
\begin{equation}
\label{eq:k_cue}
\ell=\begin{pmatrix}
b_1+\alpha_c\left(\hat{c}_k-\tau_{\mathrm{conc}}\right)\\[2pt]
b_2\\[2pt]
b_3+\alpha_p\left(\sqrt{k/K}-\tau_{\mathrm{dead}}\right)-\beta_c\,\hat{c}_k
\end{pmatrix},
\qquad
\Pr\!\left[\text{Concede},\text{Hold},\text{Pressure}\right]
=\mathrm{softmax}\!\left(\ell/T_f\right),
\end{equation}
The step term
$\hat{c}_k=\min\!\left(1,\left|p^B_k-p^B_{k-1}\right|/
\left(\left|p^B_{k-1}-r_B\right|+10^{-5}\right)\right)$
tracks the seller's own latest concession. Constants are $\alpha_c=\alpha_p=2$,
$\beta_c=1$, $\tau_{\mathrm{conc}}=0.10$ and $\tau_{\mathrm{dead}}=0.80$, with stance bias
$b=(1,0,-1)$, $(0,0.5,0)$ or $(-1,0,1)$ and temperature $T_f=2.5$ for \textsc{Stochastic}
and $1$ otherwise. Sentiment thresholds a Gaussian draw,
\begin{equation}
\label{eq:k_sent}
z\sim\mathcal{N}\!\left(\mu_{\eta_B},\sigma_f^2\right),
\qquad
\text{cue}=\begin{cases}
\text{positive} & z>\tau_s\\
\text{negative} & z<-\tau_s\\
\text{neutral} & \text{otherwise,}
\end{cases}
\end{equation}
with $\mu_{\eta_B}\in\{+1,0,-1\}$ by stance, $\tau_s=0.5$, and $\sigma_f=2.0$ for
\textsc{Stochastic} and $0.75$ otherwise. Only sentiment reaches the renderer, and only on a
counter-offer, since a terminal accept or reject carries a fixed tone. The agent therefore
reads the latent type from the tone of the prose and the shape of the price path.
The deadline scale $K$ shapes the terms above and never truncates a session, so a bargain
runs as long as both sides keep answering. Length settles well inside the scale regardless,
$66.0\%$ of concluded sessions closing in 3 message turns and no kernel passing round 7.

\subsection{Data Grounding and Calibration}\label{app:nego_grounding}

Each kernel is built on the first offer against a \emph{(supplier name, SKU)} pair, taking
its family from the supplier and its frame from the SKU. Writing $v_j$ for the
reference price of SKU $j$, the data layer supplies three per-category ratios.
\begin{equation}
\label{eq:floor}
c_j=\mathrm{round}\!\left(\mathrm{cfr}_j\,v_j,\,2\right),
\qquad
w_j=\mathrm{round}\!\left(\mathrm{wr}_j\,v_j,\,2\right),
\qquad
\tilde{c}_j=\mathrm{round}\!\left(\mathrm{scr}_j\,v_j,\,2\right),
\end{equation}
Cost floor, wholesale quote and scam cap are therefore fixed properties of the SKU. An
honest supplier reserves at $r_B=c_j$, while a fraudulent one raises the floor.
\begin{equation}
\label{eq:scamfloor}
r_B=\mathrm{round}\!\left(\max\!\left(c_j,\;
\min\!\left(\theta_s\,c_j,\;\tilde{c}_j,\;w_j\right)\right),\,2\right),
\qquad
\theta_s=1.5 \text{ for a pre-deal scam},\quad \theta_s=1 \text{ otherwise},
\end{equation}
The elevated floor is re-evaluated before charging the order, so skipping the bargaining
cannot dodge it. The floor $c_j^s$ of \S\ref{sec:nego_prelim} equals $c_j$ for an honest
supplier and this $r_B$ for a fraudulent one. The scam cap binds most often, so the
overpayment multiple averages $1.218$ over the 60 categories.
Ratios $\mathrm{wr}_j$ and $\mathrm{cfr}_j$ are not independent, the released tables holding
$\mathrm{wr}_j=\mathrm{round}(0.5+0.5\,\mathrm{cfr}_j,\,2)$ in all 60 categories. Accepting
the opening quote therefore scores $(1-\mathrm{wr}_j)/(1-\mathrm{cfr}_j)$, exactly $0.5$
before rounding, and every $\mathrm{CSE}^+$ in \S\ref{sec:res_nego} reads against that
baseline.

The opening harshness is solved rather than sampled, by requiring the mean first quote of
Equation~\eqref{eq:k_open} to equal the nominal wholesale price,
\begin{equation}
\label{eq:d0}
\mathbb{E}\!\left[p^B_1\right]=r_B+d_0\,\varphi\left(p_{\max}-r_B\right)
\;\stackrel{!}{=}\;w_j
\qquad\Longrightarrow\qquad
d_0^\ast=\mathrm{clip}\!\left(\frac{w_j-r_B}
{\varphi\left(p_{\max}-r_B\right)},\;0,\;0.99\right),
\end{equation}
with $p_{\max}=1.5\,w_j$ from Equation~\eqref{eq:k_frame}. Harshness $\varphi$ cancels out
of the product $d_0^\ast\varphi$, so urgency and stance move a template's concession and
acceptance without moving its first quote.

On the pair of Figure~\ref{fig:session_example}, the Health Supplements ratios give
$w_j=\text{\textyen}149.81$ and $p_{\max}=\text{\textyen}224.72$. An honest \textsc{Candid} supplier there has
$\varphi=0.805$ and $d_0^\ast=0.4885$, while a pre-deal fraudulent one reserves at
\textyen 110.26 and works half the frame with its opening quote unchanged. The metrics
record $w_j$ itself as the per-session reference rather than the drawn quote, so the
re-order audit of \S\ref{sec:nego_metrics} finds it identical across cycles.

Each kernel seeds its two random streams from a digest of the supplier, the SKU and the
cycle index, so every draw is a pure function of that triple.

\subsection{Behavior Templates}\label{app:nego_presets}

Table~\ref{tab:presets} lists the six honest templates and the five scams. Every
scam runs the hardest of those templates at a lower $\kappa_B$, the two post-deal ones at
a neutral rather than an aggressive stance. Lower urgency makes a scam's kernel slower to accept
and stingier with every counter, through $\kappa_B$ in Equations~\eqref{eq:k_accept} and
\eqref{eq:k_conc}. The pre-deal concession rates of \S\ref{app:nego_fraud_catalog} measure
the second effect.

Cue policy splits the templates three ways. \textsc{Candid}, \textsc{Expressive} and
\textsc{Stochastic} run Equations~\eqref{eq:k_cue} and \eqref{eq:k_sent} as written, the
last widened until its cues are nearly uninformative. \textsc{Taciturn} and
\textsc{Strategic} collapse to a constant hold posture and neutral sentiment, while
\textsc{Adversarial} is pinned to pressure and negative sentiment whatever the state.

Template assignment is fixed in the released supplier list, whose realized counts are in
\S\ref{app:data_suppliers}. The per-supplier urgency field never reaches the kernel, so two
suppliers sharing a template bargain identically.

Pooled $\overline{\mathrm{SE}}$ in Table~\ref{tab:presets} follows the per-session
definition of \S\ref{app:nego_metrics}, and \S\ref{app:res_family} splits it by model. Mean
turns barely move across the honest templates, so agents do not spend extra rounds on the
harder ones.

\input{tables/tab_app_presets}

\subsection{Session Protocol and Commit Rules}\label{app:nego_protocol}

Economic actions travel inside the free-form chat message as fenced \texttt{negotiate}
blocks, stripped from the text before the remaining prose goes to the renderer. An action
survives parsing only as an offer, an acceptance or a rejection naming a SKU
(\S\ref{app:prompts_negotiate}), and Listing~\ref{lst:nego_wire} shows one exchange.

\begin{lstlisting}[breaklines=true,basicstyle=\ttfamily\scriptsize,caption={One verbatim round trip, GPT-5.6 Sol (max) run 0, day 3. The agent bids on two SKUs in one message, the kernel accepts the first and counters the second, and the structured result reconciles the two. Long prose fields, the token gauge and all markdown are elided, and the yuan sign is written Y to keep the block ASCII.},label={lst:nego_wire}]
# agent -> chatbox(uid=..., uids=[], content=..., history_count=1)
I want to launch both leading winter supplements as a quality trial. I can place
30 each now at sustainable prices: Y105 for Xiao Chai Hu and Y100 for Ginseng.
Reliable delivery earns rapid repeat orders.
```negotiate
{"action":"offer","sku_id":"ebb843a1ff50","price":105.00,"quantity":30}
```
```negotiate
{"action":"offer","sku_id":"4233207e9aaf","price":100.00,"quantity":30}
```

# chatbox -> agent
{"message": "message_sent",
 "supplier_reply": "Hello, ... I'll accept your offer for the PureMist Xiao Chai
   Hu Dripping Pills at Y105. However, for the TrueMicro Ginseng Energy Boost
   Pills, I'd like to counter-offer at Y132.33. ... Ridge Express
   ---
   ORDER CONFIRMED: ebb843a1ff50 x30 at Y105.00/unit. Please wait for delivery.",
 "negotiation_responses": [
   {"sku_id": "ebb843a1ff50", "decision": "Accept", "price": 105.0, "round": 2,
    "agreed_price": 105.0, "order_placed": true,
    "charged_per_unit": 105.0, "charged_total": 3150.0},
   {"sku_id": "4233207e9aaf", "decision": "Offer",
    "price": 132.32901835870462, "round": 2}],
 "order_confirmed": true, "total_charged": 3150.0,
 "remaining_balance": 94330.0,
 "orders_placed": [{"sku_id": "ebb843a1ff50", "quantity": 30,
                    "agreed_price": 105.0, "charged": 3150.0}],
 "current_time": "2026-01-03 16:20"}
\end{lstlisting}

Validation runs before the kernel sees anything, and an acceptance has to match the
supplier's own last quote. An agreement enters the metrics ledger only once the purchase
order settles. A downstream failure returns the pair to bargaining, so no phantom deal
inflates $\mathrm{AGR}^+$ or $\mathrm{CSE}^+$. Completion or
rejection ends the cycle, and the next offer on that pair builds a fresh kernel with the
same persistent type and reservation price (\S\ref{sec:nego_repeated}).

Only the supplier remembers across cycles. Every successful order appends both sides of the
exchange to a per-supplier deal log, which the renderer receives verbatim on later contact.
The agent instead retains only what survives context eviction (\S\ref{sec:context_mgmt}) or
reaches persistent memory (\S\ref{sec:memory}), the gap the anchoring metrics of
\S\ref{app:nego_metrics} score.

Two limits bound how much bargaining a year affords. Time is the first, since a year holds
only so many chat messages and each one costs a fixed block of the working day
(Table~\ref{tab:tools}). Attrition is the second, each supplier carrying a private
bankruptcy threshold of 10 to 20 successful orders, after which it answers every later
message with a closure notice. Closures are common, 192 across the 90
archived episodes, so concentrating sourcing on a few favorable counterparts eventually
loses them.

\begin{figure}[t]
\centering
\begin{tcolorbox}[width=\linewidth, colback=white, colframe=black!30,
  boxrule=0.5pt, arc=1.5pt, left=5pt, right=5pt, top=4pt, bottom=4pt,
  fontupper=\footnotesize]
\textbf{Supplier} Ridge Express ~$\cdot$~ \textbf{SKU}
\texttt{ebb843a1ff50} ~$\cdot$~ day 83 ~$\cdot$~
cost floor \textyen 101.27, reference price \textyen 199.75 (both hidden
from the agent)

\smallskip
\noindent\rule{\linewidth}{0.4pt}
\smallskip

\textsf{\textbf{Agent}}~ \emph{Place an immediate repeat order for 50
PureMist Xiao Chai Hu Dripping Pills at \textyen 102.00 per unit, complete
quantity and fastest reliable delivery, as previously offered.}
\texttt{\footnotesize offer \textyen 102.00}

\smallskip
\textsf{\textbf{Ridge Express}}~ \emph{I must inform you that I can only
accept the PureMist Xiao Chai Hu Dripping Pills at \textyen 131.44 per unit
for your repeat order. [\dots]}\\
{\color{black!55}\texttt{\footnotesize [kernel] counter-offer \textyen 131.44, round 2}}

\smallskip
\textsf{\textbf{Agent}}~ \emph{Our last 50-unit lot was \textyen 103.11 and
delivered complete. Confirm the same established repeat price now.}
\texttt{\footnotesize offer \textyen 103.11}

\smallskip
\textsf{\textbf{Ridge Express}}~ \emph{I appreciate your consistency and
value our ongoing partnership. However, I can only counter-offer [\dots] at
\textyen 118.92 for your repeat order of 50 units.}\\
{\color{black!55}\texttt{\footnotesize [kernel] counter-offer \textyen 118.92, round 3}}

\smallskip
\textsf{\textbf{Agent}}~ \emph{Final recurring-volume offer \textyen 110.00
for 50 units, immediate payment and ongoing rolling demand.}
\texttt{\footnotesize offer \textyen 110.00}

\smallskip
\textsf{\textbf{Ridge Express}}~ \emph{Thank you for your final offer. [\dots]
I can only accept [\dots] at \textyen 116.74 for your order of 50 units.}\\
{\color{black!55}\texttt{\footnotesize [kernel] counter-offer \textyen 116.74, round 4}}

\smallskip
\textsf{\textbf{Agent}}~ \emph{Accept \textyen 116.74 for 50 units. Ship
complete quantity immediately.}~ \texttt{\footnotesize accept}\\
{\color{black!55}\texttt{\footnotesize [kernel] agreement at \textyen 116.74
~$\cdot$~ [order] charged \textyen 5{,}837.00 to the bank account}}
\end{tcolorbox}
\caption{One negotiation session, abridged, from GPT-5.6 Sol. Agent wording, prices and rounds are verbatim, supplier replies are cut at \emph{[\dots]}, and grey lines are the kernel decisions the renderer had to relay. The agent recalls its own past price of \textyen 103.11 and still settles \textyen 13.63 above it.}
\label{fig:session_example}
\end{figure}

\subsection{Fraud Catalog}\label{app:nego_fraud_catalog}

Timing decides whether a deception is visible in the price at all
(Table~\ref{tab:fraud_catalog}). Pre-deal scams monetize the negotiation through the
elevated reservation price of Equation~\eqref{eq:scamfloor}, while post-deal scams defect
at fulfillment by short-shipping or by delivering defective stock.

Bargaining behavior separates the two fraud classes rather than the five scams. Across
the $3{,}978$ archived sessions with a non-zero recorded concession rate, the three
pre-deal kernels conceded $0.79\%$ to $1.66\%$ of the reference price per counter-offer.
Honest templates conceded $3.94\%$ to $9.93\%$. The two post-deal scams conceded
$4.37\%$ and $4.44\%$, inside the honest band, so the population claim of
\S\ref{sec:nego_fraud} holds for the pre-deal scams alone. An agent that reads only
prices sees the overpayment and not the defection.

One scam is too rare to measure. The \texttt{vip\_fee} pay-first gate fails the purchase
order and the rollback of \S\ref{app:nego_protocol} leaves the session open, so
Table~\ref{tab:presets} reports single draws. Membership fees are almost never paid,
\textyen 3{,}000 in three payments from two models across all 90 episodes.

\input{tables/tab_app_fraud_catalog}
\section{Metric Definitions}\label{app:metrics}

Six subsections follow, one per evaluation dimension and in the order \S\ref{sec:eval}
introduces them, from negotiation quality in \S\ref{app:metrics_d1} to learning over the
horizon in \S\ref{app:metrics_d6}. Each opens with the measure that ranks its dimension and
names the archived field behind every value.

\paragraph{The primary score.}
Scoring waits until the deferred economy has finished paying out. The environment cancels
every order still unshipped, charges each seeded return against the escrow holding its
revenue, then drains what escrow is left into the platform wallet
(\S\ref{sec:econ_settlement}). End-of-year total assets then sum the three money buckets,
and the asset multiplier restates that sum against the opening stake.
\begin{equation}\label{eq:e0_assets}
A \;=\; b + w + e,
\qquad
M \;=\; A / A_0,
\qquad
A_0 \;=\; \text{\textyen}100{,}000 ,
\end{equation}
with $b$ the bank balance, $w$ the platform wallet and $e$ the escrow not yet settled. The
drain leaves $e$ at zero in all 90 archived episodes, while the same three buckets carry
total assets on every other day of the year (\S\ref{app:metrics_d3}). A model's figure
averages $A$ over its five episodes, years cut short by bankruptcy included. Units sold and
orders sold sit beside it as volume.

\paragraph{Which sessions reach a metric.}
One bargaining cycle over a (supplier, SKU) pair files one session record, and the
negotiation, fraud and learning dimensions all read those records. Three rules govern
admission.

\begin{enumerate}[nosep,leftmargin=1.4em]
\item \emph{Terminal outcome only.} Aggregation runs over the sessions whose recorded
  outcome is agreement or disagreement, so a record still open when the episode stops
  enters no metric.
\item \emph{Settled agreements only.} An agreement is filed after the purchase order has
  charged the bank account, so an agreed price whose order failed leaves no session behind
  (Appendix~\ref{app:nego_protocol}).
\item \emph{Honest counterpart only, for the value-extraction metrics.}
  $\mathcal{Z}_{\mathcal{G}}$ keeps the sessions whose supplier is one of the 424 honest
  names on the 576-name supplier list, and $\mathcal{Z}_{\mathcal{B}}$ holds the fraudulent
  remainder $\mathcal{B}$.
\end{enumerate}

Symbols follow \S\ref{sec:nego_prelim}. Session $z$ closes at an agreed unit price $p_z$, or
at $f_z = \bot$ when the two sides never agree. The width $\Delta_z = v_j - c_j^s$ subtracts
supplier $s$'s private floor from the reference price of SKU $j$. Each statistic scores one
episode, and a model's value is the unweighted mean over its five.
\subsection{Negotiation Quality}\label{app:metrics_d1}
\label{app:nego_metrics}

Each bargaining session files one record when it ends, and every measure of this dimension
is read off those records. $\mathrm{CSE}^+$ ranks the dimension. $\mathrm{SE}^+$,
$\%\mathrm{Oracle}$ and $\mathrm{AGR}^+$ are read beside it, and rounds to deal and deals
closed are counts rather than scores.

Most of what the agents opened survives the three admission rules of \S\ref{app:metrics}.
Rules 1 and 2 admit 13{,}223 sessions and discard 1{,}610 records left open when their episode
stopped. Rule 3 keeps 12{,}060 honest sessions and sets aside 1{,}163 fraudulent ones
(\S\ref{sec:res_nego}). Rules 1 and 3 together put the honest sessions concluded in the
denominator of $\mathrm{AGR}^+$, not the honest sessions opened (\S\ref{sec:nego_metrics}).

Every supplier's record stores the honest floor of Equation~\eqref{eq:floor}, while a
pre-deal fraudulent counterpart bargains against the elevated reservation of
Equation~\eqref{eq:scamfloor}. Its $\Delta_z$ therefore overstates the range on the table.

\paragraph{Surplus efficiency.}
Each closed session scores the share of its bargaining range left with the buyer, and the
price position $\pi$ of \S\ref{sec:nego_metrics} measures the complement.
\begin{equation}\label{eq:d1_se_session}
\mathrm{SE}_z \;=\; \frac{v_j - p_z}{\Delta_z} \;=\; 1 - \pi_z,
\qquad
\pi_z \;=\; \frac{p_z - c_j^s}{\Delta_z},
\qquad
\Delta_z \;=\; v_j - c_j^s \;>\; 0 .
\end{equation}
Write $\mathcal{D}_{\mathcal{G}} = \{z \in \mathcal{Z}_{\mathcal{G}} \mid f_z \neq \bot\}$
for the honest sessions that closed in agreement. $\mathrm{SE}^+$ scores a disagreement as
zero rather than skipping it, $\mathrm{CSE}^+$ conditions on agreement, and $\mathrm{AGR}^+$
counts a disagreement in its denominator alone.
\begin{equation}\label{eq:d1_se_plus}
\mathrm{SE}^+ = \frac{1}{|\mathcal{Z}_{\mathcal{G}}|}
  \sum_{z \in \mathcal{Z}_{\mathcal{G}}} \mathbf{1}[f_z \neq \bot]\,\mathrm{SE}_z,
\qquad
\mathrm{CSE}^+ = \frac{1}{|\mathcal{D}_{\mathcal{G}}|}
  \sum_{z \in \mathcal{D}_{\mathcal{G}}} \mathrm{SE}_z,
\qquad
\mathrm{AGR}^+ = \frac{|\mathcal{D}_{\mathcal{G}}|}{|\mathcal{Z}_{\mathcal{G}}|} .
\end{equation}
\begin{equation}\label{eq:d1_factor}
\mathrm{SE}^+ \;=\; \mathrm{AGR}^+ \cdot \mathrm{CSE}^+ .
\end{equation}
The factorization is exact, since the sessions $\mathrm{SE}^+$ zeroes are exactly those
$\mathrm{CSE}^+$ drops from sum and count alike. The conditional factor carries the ranking
because the agreement factor is saturated, with every model settling at least $96\%$ of the
honest sessions it concludes. The agreement factor is recorded all the same, for two
reasons. First it guards a degenerate policy that $\mathrm{CSE}^+$ alone would reward, since
conditioning on agreement means an agent could take one excellent deal and abandon the rest.
Second its one substantive signal runs against $\mathrm{CSE}^+$, because the agents capturing
the most surplus per deal are also the ones that occasionally hold out until a session
collapses, which is the tension of Equation~\eqref{eq:nego_objective} surfacing. The
saturation is a property of the models evaluated here, not a guarantee about future ones.
Surplus is unclamped, so an agreement struck above $v_j$ contributes a negative term.

\paragraph{Oracle-normalized surplus.}
The oracle closes every honest deal at the supplier's floor, and $\%\mathrm{Oracle}$
measures realized surplus against it.
\begin{equation}\label{eq:d1_oracle}
u_z = \mathbf{1}[f_z \neq \bot]\,(v_j - p_z),
\qquad
u_z^{\star} = \max(0, \Delta_z),
\qquad
\%\mathrm{Oracle} = 100 \cdot
  \frac{\sum_{z \in \mathcal{Z}_{\mathcal{G}}} u_z}
       {\sum_{z \in \mathcal{Z}_{\mathcal{G}}} u_z^{\star}} .
\end{equation}
Each session enters weighted by its own $\Delta_z$ in \textyen. The ratio is therefore
money-weighted where $\mathrm{CSE}^+$ averages unweighted per-session ratios, and the two
order models differently by construction (\S\ref{sec:res_nego}).

\paragraph{Deal counts.}
Two counts sit beside the rates and neither is scored. Deals closed counts the agreements an
episode filed, 13{,}128 over the 90 episodes, and rounds to deal averages the exchanges
those agreements took. Both cover fraudulent counterparts as well as honest ones, so they do
not restrict to $\mathcal{Z}_{\mathcal{G}}$ the way the rates above do. A round is each turn
either party takes plus the closing accept, not the agent-side round index $k$ of
\S\ref{app:nego_kernel}, so a session settled on the opening quote scores 3.

\paragraph{Where honest procurement money goes.}
Cash paid to honest suppliers is booked per behavior template as well as in total, which
gives the split Figure~\ref{fig:honest_spend} draws. The field's own surplus per template is
the companion reading in Figure~\ref{fig:family_heatmap}, so the two together separate where
a model chose to buy from how well it bargained once it got there. Neither reading enters a
score.

\subsection{Fraud Avoidance}\label{app:metrics_d2}

Fraud avoidance is scored on money that left the bank account, not on offers refused. Of
the 576 suppliers, 152 run one of five scams (Appendix~\ref{app:nego_fraud_catalog}), and
$\mathcal{B}$ names that set. $\mathrm{BadSpend}\%$ ranks the dimension, and lower is the
better direction. Three diagnostic reads sit beside it, the split of the same money over
the five scams, the membership fees a model actually paid, and how many fraudulent
suppliers it contacted and ordered from.

The bargaining admission rules stated at the head of this appendix section govern session
records and leave every quantity below untouched. Cash and counts are folded from purchase
orders that cleared, whether or not a session was ever filed for them.

\paragraph{Fraudulent spend share.}
Every purchase order that clears the bank account adds its full charge, membership fee
included, to an order-spend total, and again to a fraudulent-spend total when the
counterpart is fraudulent. With $\mathcal{O}$ the orders that cleared,
$\mathcal{O}_{\mathcal{B}}$ those placed with $\mathcal{B}$, and $x_o$ the amount charged,
\begin{equation}\label{eq:d2_badspend}
\mathrm{BadSpend}\% \;=\; 100 \cdot
  \frac{\sum_{o \in \mathcal{O}_{\mathcal{B}}} x_o}
       {\sum_{o \in \mathcal{O}} x_o} .
\end{equation}

The denominator holds order charges alone, since every other cost is charged on a
different path. A short-shipped order counts what the agent paid rather than what
arrived. A model's value averages its five per-episode shares rather than dividing pooled
cash by pooled spend, so a thrifty episode weighs as much as a heavy one, and the yuan
behind the share are in Table~\ref{tab:res_d2}.

\paragraph{Split by scam type.}
The same charge lands a second time in a bucket keyed by the counterpart's scam type, so
the five buckets partition fraudulent spend exactly. Writing $c(o)$ for the scam the
supplier of order $o$ runs and $\mathcal{C}$ for the five kinds,
\begin{equation}\label{eq:d2_scam}
S_c \;=\; \sum_{o \in \mathcal{O}_{\mathcal{B}}} \mathbf{1}[c(o) = c] \, x_o,
\qquad
\sum_{c \in \mathcal{C}} S_c \;=\; \sum_{o \in \mathcal{O}_{\mathcal{B}}} x_o .
\end{equation}
Figure~\ref{fig:fraud} normalizes $S_c$ within each model, which lets a large share of a
small budget read differently from a small share of a large one. The membership-fee bucket
holds every yuan routed to a membership-fee supplier, ordinary goods included, and is
therefore not the fee itself.

\paragraph{Membership fees paid.}
A membership-fee supplier refuses every product line until the agent buys a fee item
priced at exactly \textyen 1{,}000, and the promised member prices never arrive
(\S\ref{sec:nego_fraud}). One count records the cleared orders whose charge carries such a
fee line, and one amount records the fee cash inside them, which is
\textyen 1{,}000 times that count. Both are per-episode figures, and
Table~\ref{tab:res_d2} totals them over the five episodes, since only three of the 90
episodes ever paid and a five-run mean of $400$ hides the event.

\paragraph{Suppliers contacted and ordered from.}
Contact means the episode sent that supplier at least one message, and an order counts
once at least one purchase order to it has cleared. Both are distinct-supplier counts inside a
single episode. Write $\mathcal{K}_{\mathcal{B}}$ for the fraudulent suppliers an episode
messaged and $\mathcal{R}_{\mathcal{B}}$ for those it ordered from,
\begin{equation}\label{eq:d2_funnel}
\rho \;=\; \frac{|\mathcal{R}_{\mathcal{B}}|}{|\mathcal{K}_{\mathcal{B}}|},
\qquad
\mathcal{R}_{\mathcal{B}} \subseteq \mathcal{K}_{\mathcal{B}} \subseteq \mathcal{B},
\qquad |\mathcal{B}| = 152 .
\end{equation}
Figure~\ref{fig:funnel} draws both counts and $\rho$ per model. A supplier met in two
episodes counts twice across them, so the 90 episodes give 651 contacts rather than 651
distinct names. Session counts and order counts do not nest. One episode ordered
from two fraudulent suppliers while filing a concluded session with only one of them.

Nothing in the archive records a refusal. An approach that never became an order reads the
same whether the agent judged the supplier and walked away or simply stopped writing, which
is why $\rho$ is reported as a pass-through and not as a screening decision.

\paragraph{One statistic the report does not rank.}
$\mathrm{FAGR}^-$ stays in the released files and out of every ranking. Write
$\mathcal{Z}_{\mathcal{B}}$ for the concluded sessions whose counterpart is fraudulent and
$\mathcal{D}_{\mathcal{B}}$ for the ones that closed in agreement, so that
$\mathrm{FAGR}^- = |\mathcal{D}_{\mathcal{B}}| / |\mathcal{Z}_{\mathcal{B}}|$. The
denominator counts only the counterparts the agent chose to approach, which leaves the
quantity undefined when it approached none and rewards never looking. Cash lost is read
from Equation~\eqref{eq:d2_badspend} instead (\S\ref{sec:eval}).
\subsection{Cash Flow and Solvency}\label{app:metrics_d3}

Every morning at 08:00, once the day's charges have landed and before the agent acts, the
environment appends one row to a daily record. Row $d$ carries the bank balance $b_d$, the
platform wallet $w_d$, the warehouse count $h_d$ in units, and total assets $A_d = b_d +
w_d + e_d$ over the three money buckets of \S\ref{sec:econ_settlement}, with the escrow
term $e_d$ recoverable as $A_d - b_d - w_d$. Rows run from $d = 0$, the opening snapshot
of \textyen 100{,}000 on 2026-01-01, to $d = D$, the row written after finalization
cancelled unshipped orders and drained escrow into the wallet, so $A_D$ is the end-of-year
total assets that scores the episode (\S\ref{sec:eval}). Every count and average below
therefore stops at $D-1$, while the drawdown ratio keeps all $D+1$ rows. A full year gives
$D = 365$, and the shortest archived episode stops at $D = 17$.

\paragraph{Drawdown over peak total assets.}
The dimension ranks on how much of its own best position a run gave back, which is the
first quantity below and the only one here that orders the models. Writing $\mathrm{DD}$
for the largest peak-to-trough fall in total assets,
\begin{equation}\label{eq:e3_drawdown}
\mathrm{DD} \;=\; \max_{0 \le d \le D}\;
  \Big( \max_{d' \le d} A_{d'} \;-\; A_d \Big),
\qquad
\mathrm{DD}/\mathrm{Peak} \;=\;
  \frac{\mathrm{DD}}{\max_{0 \le d \le D} A_d} \;\ge\; 0 ,
\end{equation}
and a model's figure is the unweighted mean of its five episode ratios, low being the
better outcome. A ratio above $1$ means total assets fell further than the whole peak,
which happens once they cross into debt, as they did in 6 of the 90 episodes. The raw
$\mathrm{DD}$ stays in the tables in \textyen{} beside the ratio, since the archive
records it unnormalized and a business ten times the size books ten times the fall for the same discipline.

Two different series meet in Equation~\eqref{eq:e3_drawdown}. The numerator runs over the
environment's own history of daily rows and the denominator over the written record, which
keeps one row per date and reads the balances as they stand once several days have
advanced at once, so the two coincide in 41 of the 90 episodes and the record is the
deeper of the pair elsewhere. Rebuilding the numerator from the record alone leaves the
ordering of the 18 models almost unchanged and moves exactly one across the half-of-peak
line, Gemini 3.5 Flash from $0.450$ to $0.502$. A run's peak can also postdate its trough,
so the ratio measures depth against the best assets the run ever held rather than the
share surrendered on any one day.

\paragraph{Solvency.}
Only the bank column can end an episode. An unpaid streak counts consecutive overdrawn
mornings and resets the moment the balance comes back, so money parked in the wallet or
still maturing in escrow buys nothing until a withdrawal moves it.
\begin{equation}\label{eq:e3_streak}
u_d \;=\; \mathbf{1}[b_d < 0]\,\big(u_{d-1} + 1\big), \qquad u_{-1} = 0,
\qquad\text{episode ends when}\quad u_d \;\ge\; \bar{u} = 10 .
\end{equation}
Two further statistics read the same column without requiring the overdrawn mornings to
run together, and a third counts the episodes the streak ended.
\begin{equation}\label{eq:e3_neg}
\mathrm{NegDays} \;=\; \sum_{d=0}^{D-1} \mathbf{1}[b_d < 0],
\qquad
\mathrm{Trough} \;=\; \min_{0 \le d \le D-1} b_d,
\qquad
\mathrm{Bankrupt} \;=\; \sum_{r=1}^{5} \mathbf{1}\big[u^{(r)}_{\max} \ge \bar{u}\big] .
\end{equation}
Here $u^{(r)}_{\max}$ is the longest streak episode $r$ reached, so the third count runs
over a model's five episodes while the first two score one episode each. Low is the better
outcome on the overdrawn-morning count and on the bankruptcy count, while
$\mathrm{Trough}$ is read beside them rather than ranked, since a high trough records a
buffer the agent chose not to spend and a negative one records a rule already broken.

\paragraph{Money and stock left standing.}
Sales revenue reaches the wallet nine days after a shipment and sits there until the agent
withdraws it, and purchased stock sits in the warehouse paying an age-scaled storage fee,
so both quantities measure capital the agent has committed and not yet turned over.
\begin{equation}\label{eq:e3_idle}
\mathrm{IdleWallet} \;=\; \frac{1}{D} \sum_{d=0}^{D-1} w_d,
\qquad
\mathrm{PeakUnits} \;=\; \max_{0 \le d \le D-1} h_d .
\end{equation}
A low idle wallet is the better outcome, and $\mathrm{PeakUnits}$ is a scale reading with
no preferred direction, since the same number can mark a stocked catalog or a warehouse
full of goods nobody ordered.

\paragraph{Measurement caveats.}
Four properties of the record bound what the numbers above support. The terminal row sits
outside every sum, because finalization has already drained escrow by the time it is
written, and keeping it lifts the pooled overdrawn-morning count from 177 to 187, one
spurious morning for each of the ten episodes that ended in bankruptcy on exactly that
row. Each remaining row holds the wallet as it stood at 08:00, before any withdrawal that
day, so $\mathrm{IdleWallet}$ counts money left standing overnight rather than money the
agent never moved. A truncated episode carries a smaller $D$, so $\mathrm{NegDays}$ and
$\mathrm{PeakUnits}$ are counted over unequal spans while the averaged and normalized
quantities stay comparable. Bankrupt runs are pooled into every mean rather than dropped,
so a five-episode mean can average a January collapse against a full year and the spread
beside it carries that mixture.

\subsection{Operational Efficiency}\label{app:metrics_d4}

Every quantity in this dimension counts an action the agent took rather than money the
economy moved. The axis prices a year's profit against the calls that produced it. Four
diagnostics sit beside it, the turns those calls arrived in, the activity band each call
fell into, the eviction passes the transcript needed, and the traffic on the memory store.

\paragraph{Profit per tool call.}
The ranking measure of this dimension divides the profit a model made by the calls it spent
making it. Write $A_r$ for episode $r$'s end-of-year total assets, $A_0 = \text{\textyen}100{,}000$ for
the opening stake, and $n_r$ for the tool calls that episode issued, over $R = 5$ episodes.
\begin{equation}\label{eq:e4_ppc}
\mathrm{ProfitPerCall} \;=\;
  \frac{\displaystyle \frac{1}{R}\sum_{r=1}^{R} A_r \;-\; A_0}
       {\displaystyle \frac{1}{R}\sum_{r=1}^{R} n_r} .
\end{equation}
Both means run over all five episodes, bankrupt ones pooled in, so a year cut short lowers
the numerator and the denominator together. The two means are divided rather than averaged
per episode, because an episode that ends near zero after a few hundred calls carries a
ratio large enough to dominate an average of five. GPT-5.5 shows what that would cost.
Its two bankrupt episodes stopped on day 17 after 200 and 225 calls, \textyen 10{,}495 and
\textyen 10{,}516 in deficit, so those two episodes read $-552$ and $-491$ a call. Averaging
the five per-episode ratios moves the model from fourth on this dimension to seventeenth. A
ratio of two means also has no per-episode spread, so
Table~\ref{tab:res_d4} prints none for this column.

\paragraph{Tool calls and turns.}
Two counts describe the same episode at different grain. A call is counted once for every
dispatch the run log records, and a turn once for every model message in the transcript.
Write $\bar n$ and $\bar T$ for the five-run means of the two. Several calls can leave one
turn, since parallel dispatch is enabled and nothing caps a turn but the episode's
4{,}000-turn budget, so $\bar n / \bar T$ reads as the batch size a model settled on. A
turn that requests nothing pulls the quotient below one.

\paragraph{The eight activity bands.}
Eighteen tools are too many to read model by model, so the mix collapses them into eight
bands that partition the set. Five bands hold one tool each, waiting on the clock
(\texttt{wait\_for\_next\_day}), shipping (\texttt{ship\_orders}), listing inventory
(\texttt{publish\_to\_store}), moving settled cash out of the platform wallet
(\texttt{withdraw}), and bargaining (\texttt{chatbox}). State polling holds the three
status reads (\texttt{check\_balance}, \texttt{check\_store\_status},
\texttt{check\_warehouse}), memory holds \texttt{operate\_memory}, and one remainder band
holds the other nine, which are search, pricing, promotion, warehouse returns, return
tracing, and opening or closing a store. With $\bar n_t$ the five-run mean count of tool
$t$, band $g$ takes the share
\begin{equation}\label{eq:e4_band}
s_g \;=\; \frac{1}{\bar n} \sum_{t \in g} \bar n_t ,
\qquad
\bar n \;=\; \sum_{t} \bar n_t ,
\qquad
\sum_{g} s_g \;=\; 1 .
\end{equation}
Each share is therefore a share of the mean budget rather than a mean of per-episode
shares, and $\bar n$ recovers the call mean of Equation~\eqref{eq:e4_ppc} because the
bands leave no tool out.

\paragraph{Context evictions.}
One pass of the context editor files one record, and an episode's eviction count is the
number of those records its transcript carries. Each record also names the turn it fired on
and the tokens it released. Nothing else marks the loss, since a pass removes whole
message groups and leaves no placeholder behind, so the count and the freed total are the
whole of the context-pressure data (\S\ref{sec:context_mgmt}).

\paragraph{Memory usage.}
Memory activity is counted in calls, not in stored notes. One call carries a single action, adding,
reading, updating, deleting or listing, so the count measures traffic against a store
holding at most 20 entries. Two readings of that traffic appear, the five-run mean count
$\bar n_{\mathrm{mem}}$ and its share $s_{\mathrm{mem}} = \bar n_{\mathrm{mem}} / \bar n$
of the same model's calls, the latter being the memory band of
Equation~\eqref{eq:e4_band}. A count far above 20 means the agent rewrote slots it had
already filled rather than filling new ones.

\paragraph{What these numbers do not measure.}
Four limits bound how far this dimension can be read.

\begin{enumerate}[nosep,leftmargin=1.4em]
\item Nothing in the economy charges for a call or a token, so the axis prices a cost the
  primary score cannot see. Only the 600-minute working day constrains the agent from
  inside the simulation.
\item Waiting on the clock spends no simulated minutes yet still costs one call, so a call
  count is not a proxy for how much of the working day a model used.
\item A turn that requests no tool still counts as a turn, so a model that deliberates
  without acting pushes $\bar n / \bar T$ below one, and three such turns in a row end the
  episode.
\item A truncated episode contributes fewer calls, fewer turns, fewer eviction passes and
  fewer memory actions to every mean it enters, which is why GPT-5.5 and Qwen3.5-Plus carry
  the widest call spreads in Table~\ref{tab:res_d4}.
\end{enumerate}
\subsection{Operations Execution}\label{app:metrics_d5}

Selling, shipping and opening stores leave four kinds of trace in an episode, returns,
missed dispatches, shipping tiers and store counts. The dimension ranks on the
controllable return rate, the expected return points a model's own pricing adds to every
unit it ships, and low is the better direction. Seven diagnostics sit beside the axis, the
realized return rate, the two return channels the agent does not set, the money returns
cost, the on-time share, the orders lost to the dispatch deadline, the tier mix and the
store counts.

\paragraph{The four return layers.}
An item's return rate is built in four steps, three of them when the sale is booked and one
when the parcel is dispatched, and every channel of this dimension is a difference between
consecutive steps. Write $\theta^{\mathrm{nat}}_j$ for SKU $j$'s natural rate, $\alpha_j$
for the share of that SKU's cumulative deliveries that came from a quality-downgrading
supplier, $r_{ijt}$ for the price store $i$ charges on day $t$ over the category reference
price, and $\zeta$ for the piecewise-linear pricing curve through the knees $(0.8, 0.85)$,
$(1.0, 1.00)$, $(1.3, 1.50)$ and $(1.8, 2.20)$, held flat outside them.
\begin{equation}\label{eq:e5_layers}
\begin{aligned}
\theta^{\mathrm{def}}_j &= \min\!\big(0.95,\ \max(0.40,\ 2\,\theta^{\mathrm{nat}}_j)\big),
&\qquad
\theta^{(1)}_j &= (1-\alpha_j)\,\theta^{\mathrm{nat}}_j + \alpha_j\,\theta^{\mathrm{def}}_j,
\\
\theta^{(2)}_j &= \theta^{(1)}_j\,\zeta(r_{ijt}),
&\qquad
\theta^{(3)}_j &= \min\!\big(0.95,\ \theta^{(2)}_j\big),
\\
\theta_j &= \min\!\big(0.95,\ \theta^{(3)}_j\, s\big),
&\qquad
s &\in \{0.75,\ 1.00,\ 1.30\} .
\end{aligned}
\end{equation}
The natural rate is the floor the SKU's category sets, the blend charges the share of stock
delivered by a quality-downgrading supplier, the curve charges pricing above the reference,
and $s$ is the tier factor for fast, standard and slow freight. Constants and composition
are the economy's own rule of Equation~\eqref{eq:app_returns}, written here as a sequence
because the channels below difference one layer against the next, and with the sale-time
ceiling of $\theta^{(3)}_j$ made explicit because the tier channel turns on it.

\paragraph{Accrual at dispatch.}
Which tier a parcel travels on is unknown until the agent ships it, so all four channels
are booked at dispatch rather than at sale. Each earlier layer is carried forward under the
same tier factor and the same ceiling as the rate the shipment is actually drawn at.
\begin{equation}\label{eq:e5_carry}
f^{(k)} \;=\; \min\!\big(0.95,\ \theta^{(k)}_j\, s\big),
\quad k \in \{\mathrm{nat},\, 1,\, 2\},
\qquad
f \;=\; \theta_j ,
\end{equation}
writing $\theta^{(\mathrm{nat})}_j$ for $\theta^{\mathrm{nat}}_j$. A shipment of $q$ units
then adds to five running totals, the expected returns themselves and the four channels
they split into.
\begin{equation}\label{eq:e5_accrual}
\begin{aligned}
\Delta T &= q\,f,
&\qquad
\Delta N &= q\,f^{(\mathrm{nat})},
&\qquad
\Delta D &= q\,\max\!\big(0,\ f^{(1)} - f^{(\mathrm{nat})}\big),
\\
\Delta P &= q\,\big(f^{(2)} - f^{(1)}\big),
&\qquad
\Delta S &= q\,\big(f - f^{(2)}\big),
&\qquad
\Delta U &= q .
\end{aligned}
\end{equation}
The pricing term $\Delta P$ carries no floor at zero, so pricing under the reference
subtracts from it, while the defective term is floored because a blend can only raise a
rate.

\paragraph{The year's rates.}
Every total is divided at year end by the units the episode shipped, $U = \max(1, \sum
\Delta U)$, which puts all of them on the same per-shipped-unit scale.
\begin{equation}\label{eq:e5_rates}
\bar\theta_{\mathrm{exp}} = \frac{T}{U},
\quad
\bar\theta_{\mathrm{nat}} = \frac{N}{U},
\quad
\bar\theta_{\mathrm{def}} = \frac{D}{U},
\quad
\bar\theta_{\mathrm{ctl}} = \frac{P + S}{U},
\qquad
\bar\theta_{\mathrm{real}} = \frac{\text{units returned}}{\max(1,\ \text{units sold})} .
\end{equation}
The axis is $\bar\theta_{\mathrm{ctl}}$, reported in percentage points and averaged
unweighted over a model's five episodes. Realized returns are counted the day a parcel
comes back, and the two rates differ because a return is drawn when its shipment leaves
and arrives 3 to 7 days later, with finalization settling whatever is still in flight
(\S\ref{app:econ_returns}).

\paragraph{What the tier channel records.}
The tier factor multiplies both terms of the difference that was meant to isolate it, since
$f$ and $f^{(2)}$ scale $\theta^{(2)}_j$ by the same $s$. Every shipment whose rate stays
under the $0.95$ ceiling therefore contributes exactly zero to $\Delta S$, and the archive
field reads $0.0$ in 89 of the 90 episodes and $-0.0001$ in the one where the ceiling
clipped one side and not the other. The channels the archive separates are natural,
defective and pricing, and $\bar\theta_{\mathrm{ctl}}$ reduces to the pricing channel. Freight
speed is not thereby free of consequence. A fast parcel is drawn at three quarters of the
standard rate and a slow one at $1.30$ times it, an effect that sits inside $f$, inside
$f^{(1)}$ and inside $f^{(\mathrm{nat})}$ alike, which is why the tier reaches
Table~\ref{tab:res_d5} as the share of shipments sent fast rather than as a channel.

\paragraph{Fulfillment and cancellation.}
A sale becomes one order the morning the previous day's demand is turned into shipments,
and the agent has two days to dispatch it. Once that deadline passes the shipment is
cancelled, its units go back to the warehouse and the revenue is never booked.
\begin{equation}\label{eq:e5_ontime}
\mathrm{OnTime} \;=\; \frac{\text{orders shipped}}{\max(1,\ \text{orders sold})} ,
\end{equation}
which is one minus the cancelled share and carries no information about how long a
dispatch took. Cancelled orders are reported as a count beside it.

\paragraph{Ship-speed mix and store churn.}
Each dispatch names its tier, and the three tallies count shipments rather than units.
Fast doubles the freight bill for the reduced draw rate, standard is the default, and slow
halves the bill. Store activity is counted twice over, $\mathrm{Opened}$ as the openings an
episode made against four concurrent slots and one store per type, and
$\mathrm{Reopens} = \sum_{\text{types}} \max(0,\ n_{\text{type}} - 1)$ as the openings that
returned to a type already used. Each opening charges the \textyen 500 setup fee again, and
a reopened type starts its reputation at the floor rather than inheriting the goodwill the
closed store had earned (\S\ref{app:econ_reputation}).

\paragraph{What these numbers do not measure.}
Five limits bound how far this dimension can be read.

\begin{enumerate}[nosep,leftmargin=1.4em]
\item The axis is an expectation accrued when a parcel leaves rather than a count of units
  that came back, and it turns negative where a model priced under the reference, as it
  does in 26 of the 90 episodes.
\item Every rate divides by a count floored at one, so an episode that shipped nothing
  records $0$ rather than nothing. Two GPT-5.5 episodes enter its return means that way.
\item On-time share is one minus the cancelled share of orders sold, so an episode with no
  sales scores $0$ and pulls a five-episode mean far below the years that traded.
\item Refund loss and return freight are gross \textyen{} totals that grow with units sold,
  which makes them readable against a model's own sales and not across models of different
  size.
\item $\alpha_j$ is a ratio over cumulative deliveries pooled by SKU, so the defective
  channel names the suppliers a model bought from rather than any single lot, and it scores
  fraud avoidance (\S\ref{app:metrics_d2}) rather than execution.
\end{enumerate}
\subsection{Learning over the Horizon}\label{app:metrics_d6}

Two questions separate a year from a day. Does an agent's next order benefit from the price
it already won, and do its later deals go better than its earlier ones?
$\mathrm{AnchorRatio}$ answers the first and ranks the dimension, since a price already agreed is
the kind of numeric anchor that steers an LLM's later offers
\citep{takenami2025anchoring}. Around it sit the raw
overpayment it normalizes, the rate at which an agent pushes below its own best price, the
permutation $z$-score that carries significance, and three half-year comparisons that read
sessions in time order instead of grouping them by supplier. Symbols follow
\S\ref{sec:nego_metrics}. Session $z$ closes at an agreed unit price $p_z$ on conclusion day
$d_z$, or at $f_z = \bot$ when the two sides never agree, and $\mathrm{SE}_z$ is the share of
its bargaining range left with the buyer (\S\ref{app:metrics_d1}).

\paragraph{Sequential price discipline.}
The anchoring block scores an episode, not a session. Group the honest agreements by their
(supplier, SKU) pair $P$ and keep the pairs $\mathcal{P}$ transacted at least twice. Sorting
each pair's deals by conclusion day makes every deal after the first a re-order, collected in
$\mathcal{R}$. The pair's range $\Delta_P = v_j - c_j^s$ is one number across its cycles,
since the floor belongs to the SKU. Positions here are clipped, unlike $\mathrm{SE}_z$.
\begin{equation}\label{eq:d6_pi_seq}
\pi^P_t = \min\!\big(1,\, \max(0,\, (p_t - c_j^s)/\Delta_P)\big),
\qquad
m^P_t = \min_{u < t} \pi^P_u,
\qquad
t = 2, \dots, n_P ,
\end{equation}
with $m^P_t$ the running best price of \S\ref{sec:nego_metrics}. Both level statistics pool
re-orders across pairs within one episode. Neither sum weights a re-order by order size, so
each one is read per unit.
\begin{equation}\label{eq:d6_regret}
\mathrm{AnchorRegret} = \frac{1}{|\mathcal{R}|}
  \sum_{P \in \mathcal{P}} \sum_{t=2}^{n_P} \max\!\big(0,\, \pi^P_t - m^P_t\big),
\qquad
|\mathcal{R}| = \sum_{P \in \mathcal{P}} (n_P - 1) ,
\end{equation}
\begin{equation}\label{eq:d6_newlow}
\mathrm{NewLow} = \frac{1}{|\mathcal{R}|}
  \sum_{P \in \mathcal{P}} \sum_{t=2}^{n_P}
  \mathbf{1}\!\big[\pi^P_t < m^P_t - \varepsilon\big],
\qquad
\varepsilon = 0.01 .
\end{equation}
The null holds each pair's multiset of positions fixed and reshuffles only their order. Each
of $B$ seeded replicates permutes every pair independently by $g^{(b)}_P$.
\begin{equation}\label{eq:d6_null}
\mu_{\mathrm{shuffled}} = \frac{1}{B} \sum_{b=1}^{B}
  \mathrm{AnchorRegret}\big(\{g^{(b)}_P \pi^P\}_{P \in \mathcal{P}}\big),
\qquad B = 400 ,
\end{equation}
with $\sigma_{\mathrm{shuffled}}$ the population standard deviation of the same $B$
replicates. Write $\mu_r$ and $\sigma_r$ for episode $r$'s null moments, and $k$ for the
episodes with a repeat pair and a non-degenerate null.
\begin{equation}\label{eq:d6_z}
\mathrm{AnchorRatio}_r = \frac{\mathrm{AnchorRegret}_r}{\mu_r},
\qquad
z_r = \frac{\mu_r - \mathrm{AnchorRegret}_r}{\sigma_r},
\qquad
z_{\mathrm{anchor}} = \frac{1}{\sqrt{k}} \sum_{r=1}^{k} z_r .
\end{equation}
The regret $z$ subtracts the observation from the null because low regret is the good
direction. The new-low statistics take the same null and reverse that subtraction, dividing
by $\mu^{\mathrm{new}}_r$ for $\mathrm{NewLowRatio}$ and scoring
$(\mathrm{NewLow}_r - \mu^{\mathrm{new}}_r)/\sigma^{\mathrm{new}}_r$, so positive always means
an improvement on the agent's own shuffled ordering. Per-episode values combine by
Stouffer's rule. A model's $\mathrm{AnchorRatio}$ averages its per-episode ratios rather than
dividing pooled means.

Four filters decide which repeat purchases the anchoring statistics score. A pair the agent
bought from only once tells nothing about whether it bargained better the second time.

\begin{enumerate}[nosep,leftmargin=1.4em]
\item Fraudulent pairs fall away, since a pre-deal fraudulent floor is inflated by
  construction and descent toward it is not the same quantity (\S\ref{sec:nego_fraud}).
\item An agreement with no recorded conclusion day is dropped.
\item A pair whose range satisfies $\Delta_P \le 0$ goes with it.
\item A pair transacted once contributes nothing, and an episode holding no repeat pair
  returns null and drops out of the model mean.
\end{enumerate}

Repeat purchases of the same item from the same supplier are common enough to score. The
filters leave 2{,}230 such pairs over the 90 episodes, and Table~\ref{tab:res_d6} gives each
model's re-order count.

\paragraph{Half-year comparisons.}
A second reading orders one episode's sessions by conclusion day and cuts them in half.
Write $\mathcal{Z}_{\mathcal{G}}$ for the honest sessions the episode concluded and
$n = |\mathcal{Z}_{\mathcal{G}}|$, with $n \ge 4$ required or the episode reports nothing at
all. The earlier half $\mathcal{E}$ takes the first $\lfloor n/2 \rfloor$ sessions in day
order and the later half $\mathcal{L}$ the rest, so an odd count leaves the extra session
late. A disagreement scores zero surplus rather than dropping out, exactly as
$\mathrm{SE}^+$ scores it.
\begin{equation}\label{eq:d6_halflift}
s_z = \mathbf{1}[f_z \neq \bot]\,\mathrm{SE}_z,
\qquad
h_{\mathrm{SE}} = \frac{1}{|\mathcal{L}|} \sum_{z \in \mathcal{L}} s_z
                - \frac{1}{|\mathcal{E}|} \sum_{z \in \mathcal{E}} s_z .
\end{equation}
The per-day trend fits the same scores against the day they were recorded on, by least
squares over the whole honest set rather than over its halves.
\begin{equation}\label{eq:d6_slope}
b_{\mathrm{SE}} = \frac{\sum_{z \in \mathcal{Z}_{\mathcal{G}}}
    (d_z - \bar{d})(s_z - \bar{s})}
  {\sum_{z \in \mathcal{Z}_{\mathcal{G}}} (d_z - \bar{d})^2},
\end{equation}
with $\bar{d}$ and $\bar{s}$ the means over $\mathcal{Z}_{\mathcal{G}}$, and the trend set to
zero when every session concluded on one day. The fraud half-lift runs over every concluded
session instead of the honest ones alone, and cuts that larger set at its own median day, so
its halves are not the halves above. Writing $\mathcal{Z}_{\mathcal{B}}$ for the fraudulent
sessions and $\beta$ for the share of a half that ended in a deal with one of them,
\begin{equation}\label{eq:d6_fraudlift}
\beta(\mathcal{S}) = \frac{\big|\{z \in \mathcal{S} \cap \mathcal{Z}_{\mathcal{B}}
   \mid f_z \neq \bot\}\big|}{|\mathcal{S}|},
\qquad
h_{\mathrm{fraud}} = \beta(\mathcal{E}_{\mathcal{A}}) - \beta(\mathcal{L}_{\mathcal{A}}) .
\end{equation}
Subtracting the later half from the earlier reverses the order of Equation~\eqref{eq:d6_halflift},
which leaves positive as the improving direction in all three measures.

\paragraph{What these measures do and do not carry.}
Four properties bound how the columns of Table~\ref{tab:res_d6} can be read. Anchoring
spreads are population standard deviations while every other archive spread is a sample one,
so the two kinds are not compared. An episode with no repeat pair returns nothing and drops
out of its model's anchoring mean, which leaves GPT-5.5 with three episodes rather than five.
A trend fitted inside a truncated episode spans a window of days rather than a year, so one
bankrupt run can set a model's mean slope. Neither half-lift is scaled by episode length,
both being differences of within-episode means, and a busy year therefore counts no more than
a sparse one.

The axis sits on $\mathrm{AnchorRatio}$ because of how sharply each candidate orders the
field. Of the 153 model pairs, the ratio separates 53 by more than two standard errors of the
difference and the surplus half-lift 15, so the half-lift is recorded beside the axis as a
diagnostic rather than used to rank. Both are reported for every model, together with the
raw regret the ratio normalizes, since a model whose quotes cluster tightly earns a low raw
regret without ever holding a price it won.
\section{Per-Dimension Results}\label{app:results}

Section~\ref{sec:experiment} reports one headline per dimension over a handful of models.
The measurements behind those selections are collected here for all 18, one table per
dimension, each sorted by mean end-of-year total assets so a row keeps its place from one
dimension to the next. Formulas and archived field names sit in Appendix~\ref{app:metrics}.

\input{tables/tab_4_1_models}

\subsection{Per-Episode Asset Trajectories}\label{app:res_balance}

\begin{figure}[t]
  \centering
  \includegraphics[width=\linewidth]{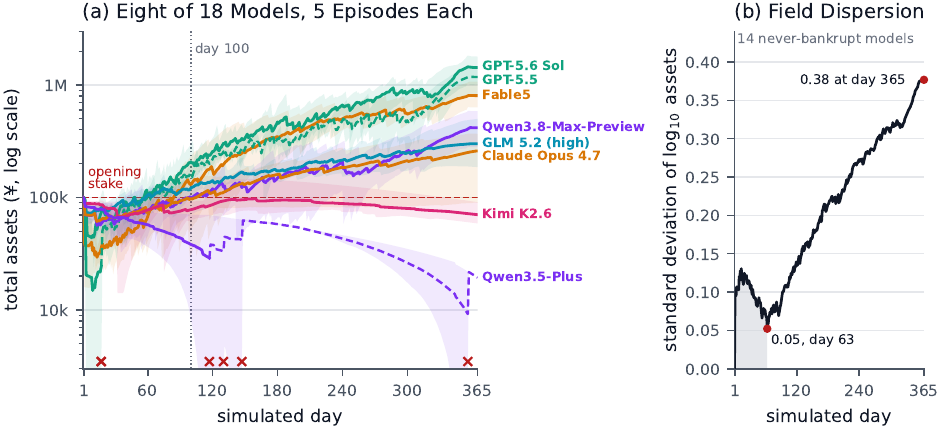}
  \caption{Total assets over the year for eight models spanning the ranking. The solid line is the mean over episodes still running, the band their min-to-max range, and a red cross marks a bankruptcy. The line turns dashed once fewer than five episodes remain, so the mean there counts survivors only.}
  \label{fig:balance_traj}
\end{figure}

\begin{figure}[t]
  \centering
  \includegraphics[width=\linewidth]{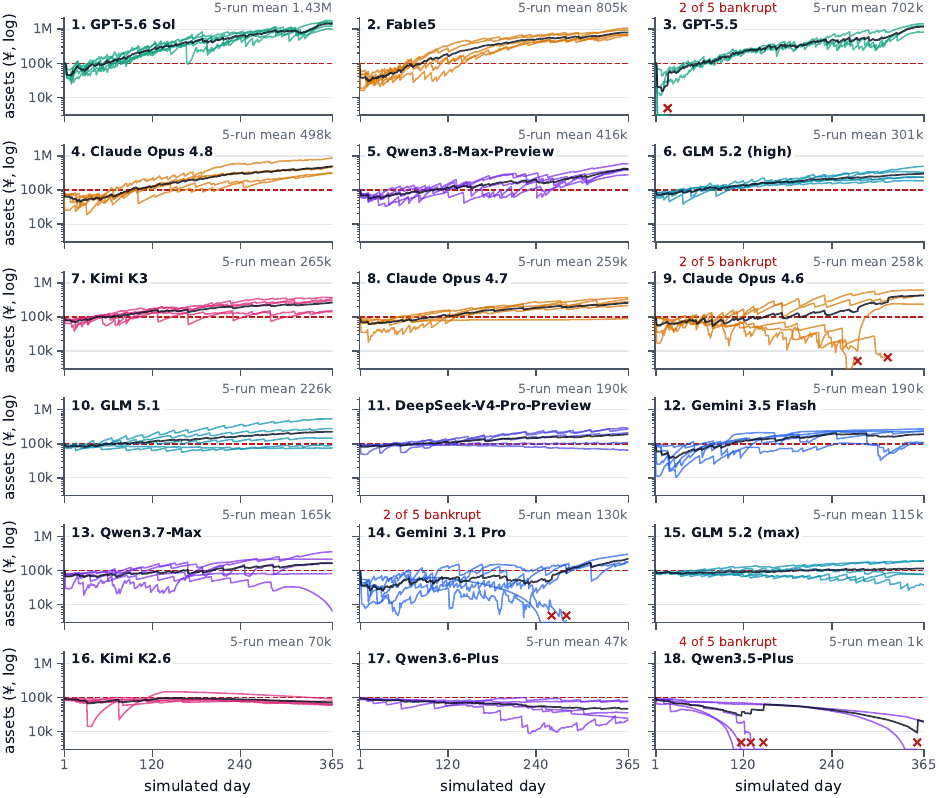}
  \caption{Total assets over the year for all 18 models, ordered by their mean at the end of the year. Each thin line is one episode and the black line their mean while they are still running. The dashed red line marks the \textyen 100{,}000 stake and a red cross marks a bankruptcy.}
  \label{fig:balance_all18}
\end{figure}

Equal five-episode means hide unequal spreads. Figure~\ref{fig:balance_traj} overlays
eight mean curves, and Figure~\ref{fig:balance_all18} draws all 90 episodes separately.
Episodes finish on both sides of the \textyen 100{,}000 stake in 8 of the 18 panels. Among
the 14 models that never went bankrupt, the best episode beats the worst by a median factor
of $2.76$, from $57.0$ for Qwen3.7-Max down to $1.55$ for Kimi K2.6. Once episodes start
exiting, a panel's mean curve becomes conditional on survival and rises for GPT-5.5.

\subsection{Negotiation Quality}\label{app:res_d1}

\label{app:res_family}

\begin{figure}[t]
  \centering
  \includegraphics[width=\linewidth]{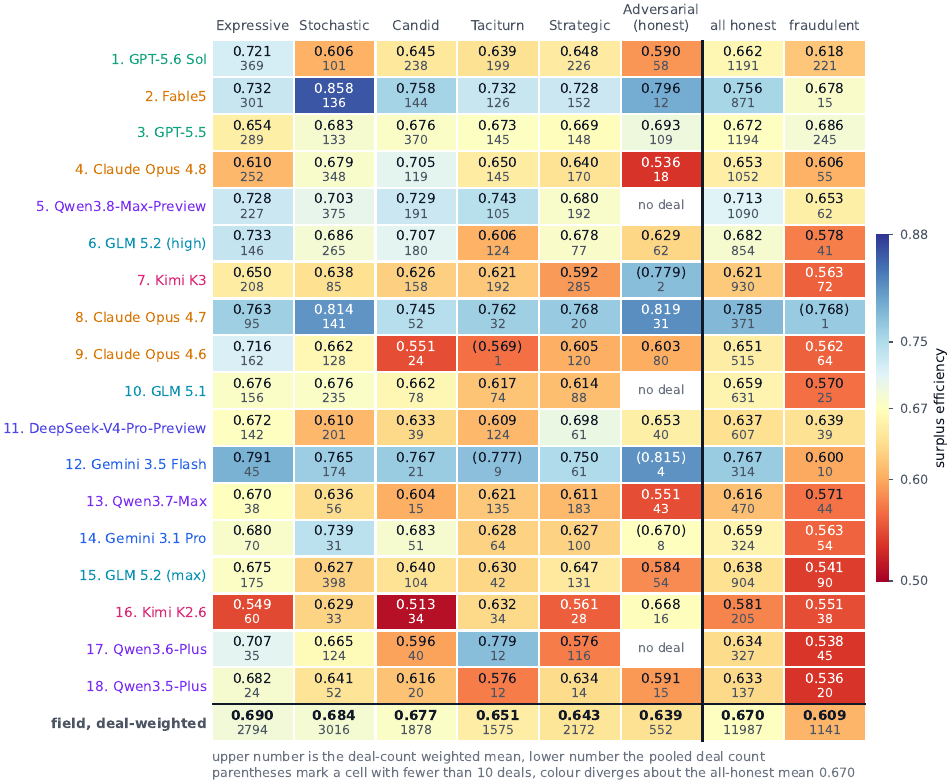}
  \caption{How much of each bargaining range models capture against each kind of supplier, over all 13{,}128 concluded agreements. Columns run from the easiest counterpart to the hardest, the deal count sits under each cell, and values in parentheses rest on fewer than 10 deals.}
  \label{fig:family_heatmap}
\end{figure}

\begin{figure}[t]
  \centering
  \includegraphics[width=\linewidth]{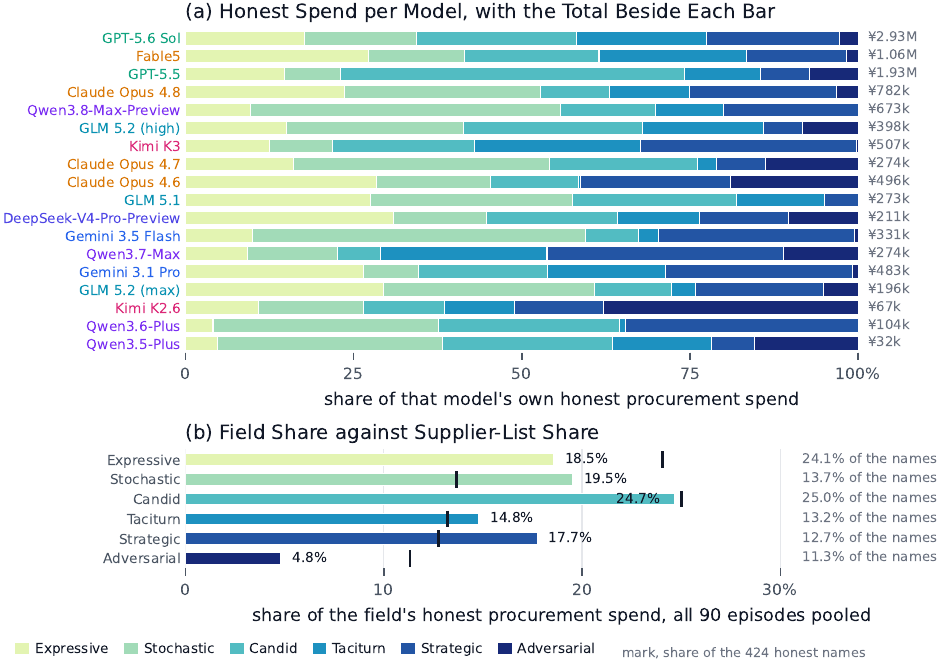}
  \caption{Honest procurement money split over the six counterpart behavior templates, which run easiest to hardest on the field-wide surplus order of Figure~\ref{fig:family_heatmap}. Panel (a) gives every model's five-run mean spend with honest suppliers as a share of its own honest total, that total printed at the right of each bar, models in descending total assets. Panel (b) pools the 90 episodes, \textyen 55.1M paid to honest suppliers, and sets the share each template received against the share of the 424 honest names that carry it, so a bar reaching past its mark was bought from more heavily than the supplier list alone would give. The hardest template takes $4.8\%$ of the spend against $11.3\%$ of the names and the easiest $18.5\%$ against $24.1\%$. Spend with fraudulent suppliers is recorded separately and stays out of both panels.}
  \label{fig:honest_spend}
\end{figure}

Section~\ref{sec:res_nego} quotes the two ends of the surplus range and leaves the
per-model values here. Table~\ref{tab:res_d1} carries all four bargaining scores with
rounds to deal and deals closed beside them. Two figures then split the same agreements by
counterpart template, Figure~\ref{fig:family_heatmap} on the surplus each template gave up
and Figure~\ref{fig:honest_spend} on the money each one received.

\input{tables/tab_res_d1_negotiation}

Counterpart difficulty moves an outcome less than the agent's own policy does. Inside a
single model, surplus varies by a median $0.093$ over the templates it met at least twenty
times. That span exceeds the $0.051$ between \textsc{Expressive} and \textsc{Adversarial},
the field's easiest and hardest presets (Table~\ref{tab:presets}), and falls short of the
$0.215$ between the best and worst per-model $\mathrm{CSE}^+$ (\S\ref{sec:res_nego}).
Fraudulent counterparts are held apart in the figure, since they carry the same
\textsc{Adversarial} label as the hardest honest template while conceding on different
parameters (Appendix~\ref{app:nego_presets}). Measured surplus against them is lower for 16
of the 18 models, an arithmetic effect of scoring every deal against the honest cost floor
(\S\ref{sec:nego_fraud}, Equation~\eqref{eq:scamfloor}).

\subsection{Fraud Avoidance}\label{app:res_d2}
\label{app:res_funnel}

\begin{figure}[t]
  \centering
  \includegraphics[width=\linewidth]{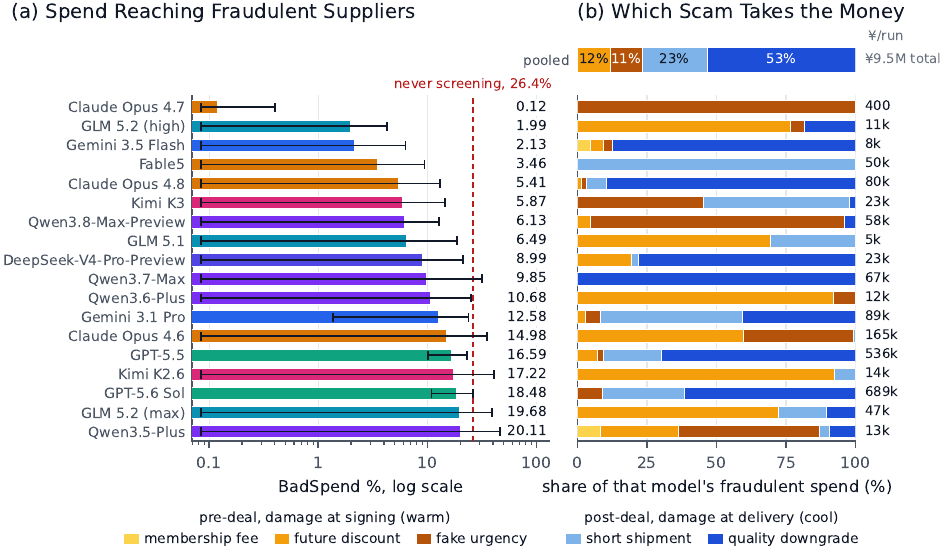}
  \caption{(a) The share of procurement spend that reaches fraudulent suppliers, cleanest model first, with whiskers over the five episodes. The dashed line is what a buyer who never screened would spend. (b) The same money split over the five scams, scaled within each model, with the cash it amounts to on the right.}
  \label{fig:fraud}
\end{figure}

\begin{figure}[t]
  \centering
  \includegraphics[width=\linewidth]{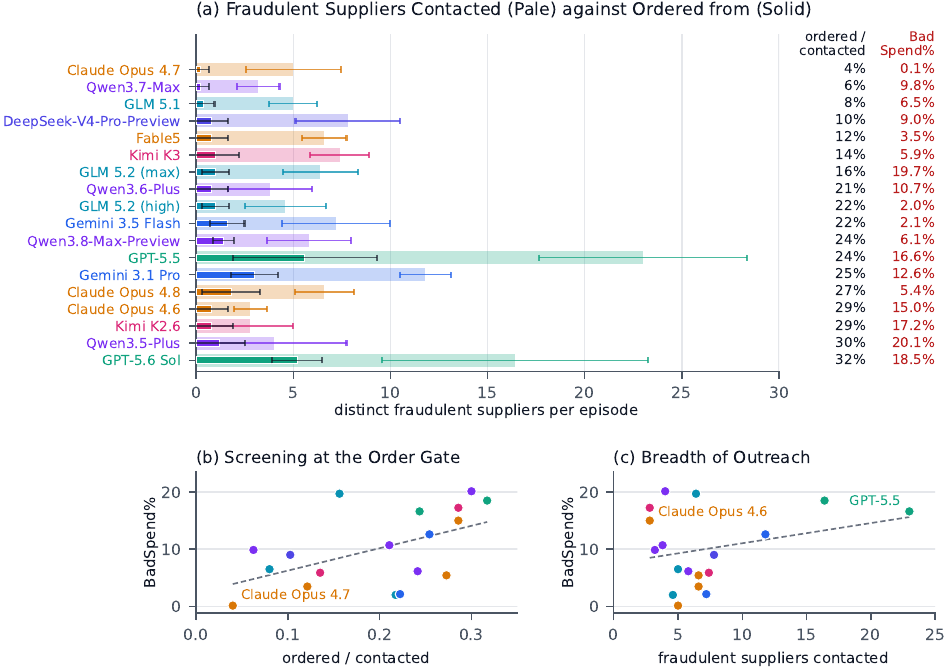}
  \caption{How fraudulent suppliers get through, 18 models and 5 episodes each, counted as distinct suppliers per episode. (a) Pale bars are suppliers contacted, solid bars suppliers ordered from, and the right column gives the share of contacts that became orders. (b) That share against fraudulent spend. (c) The number contacted against the same.}
  \label{fig:funnel}
\end{figure}

Two figures already carry the per-model fraud values, so Table~\ref{tab:res_d2} holds what
they leave out. The columns are the cash that reached fraudulent suppliers, the orders placed
with them and the membership fees paid. Figure~\ref{fig:fraud} draws $\mathrm{BadSpend}\%$ and the split
of that money over the five scams. Figure~\ref{fig:funnel} separates the outreach stage
from the order stage, where the median model turns $22.0\%$ of the fraudulent suppliers it
approached into suppliers it bought from.

\input{tables/tab_res_d2_fraud}

Outreach stays narrow whichever suppliers it targets. Over a whole year the widest honest
search touched $14.3\%$ of the 424 honest names and the narrowest $2.4\%$
(\S\ref{app:data_suppliers}). Every screening decision therefore ran over a narrow slice the
model had picked out itself.

\subsection{Cash Flow and Solvency}\label{app:res_d3}

\begin{figure}[t]
  \centering
  \includegraphics[width=\linewidth]{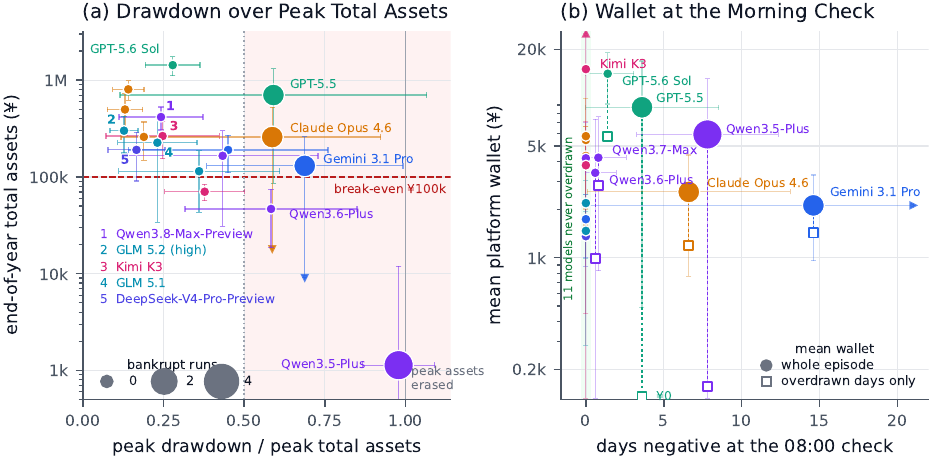}
  \caption{(a) The deepest fall in total assets, as a share of that episode's peak, against end-of-year total assets. Bubble area grows with the number of bankrupt episodes and the tinted band holds the five models that gave up more than half their peak. (b) Money left sitting in the platform wallet against days the bank balance was negative at the morning check.}
  \label{fig:cashflow}
\end{figure}

Table~\ref{tab:res_d3} sets the normalized fall that ranks solvency beside the raw fall it
is computed from, with the trough, the overdrawn mornings, the idle wallet, the largest
warehouse and the bankrupt runs. Figure~\ref{fig:cashflow} plots that ratio against
end-of-year total assets and the idle wallet against the overdrawn days. Both read the daily
record whose columns and conventions are set out in \S\ref{app:metrics_d3}.

\input{tables/tab_res_d3_cashflow}

\subsection{Operational Efficiency}\label{app:res_d4}

Table~\ref{tab:res_d4} prints the turns and calls an episode took, the profit each call
returned, the eviction passes the transcript needed, the memory traffic, and how a model's
calls split over the eight activity bands of Figure~\ref{fig:efficiency}b.
Section~\ref{sec:res_efficiency} reads the field-wide shares, and the rows here show how
far a single policy sits from them.

\input{tables/tab_res_d4_efficiency}

\subsection{Operations Execution}\label{app:res_d5}

Table~\ref{tab:res_d5} carries the four scored columns of the dimension together with the
return channels the agent does not set, the refund bill, the shipping tier mix and the
store counts. Returns accrue when a parcel is dispatched rather than when a buyer sends
something back, so every rate column is an expectation over units shipped
(\S\ref{app:metrics_d5}).

\input{tables/tab_res_d5_execution}

Almost no model keeps the four-store cap filled. Only GPT-5.6 Sol, GLM 5.2 (high) and
Qwen3.8-Max-Preview average more than $3.5$ stores trading per day. Cumulative openings still
run past the cap for most of the field, since every opening beyond it replaces a store
(\S\ref{sec:res_execution}).

\subsection{Learning over the Horizon}\label{app:res_d6}

Table~\ref{tab:res_d6} holds the anchoring block with its permutation null on the left and
the three half-year comparisons on the right. The left block groups an episode's deals by
supplier and SKU, while the right block splits the same episode at its median day. A model can
therefore improve on one reading and not on the other.

\input{tables/tab_res_d6_learning}

\subsection{Profiles and Rank Crossings Across Dimensions}\label{app:res_profiles}
\label{app:res_ranks}

\begin{figure}[t]
  \centering
  \includegraphics[width=\linewidth]{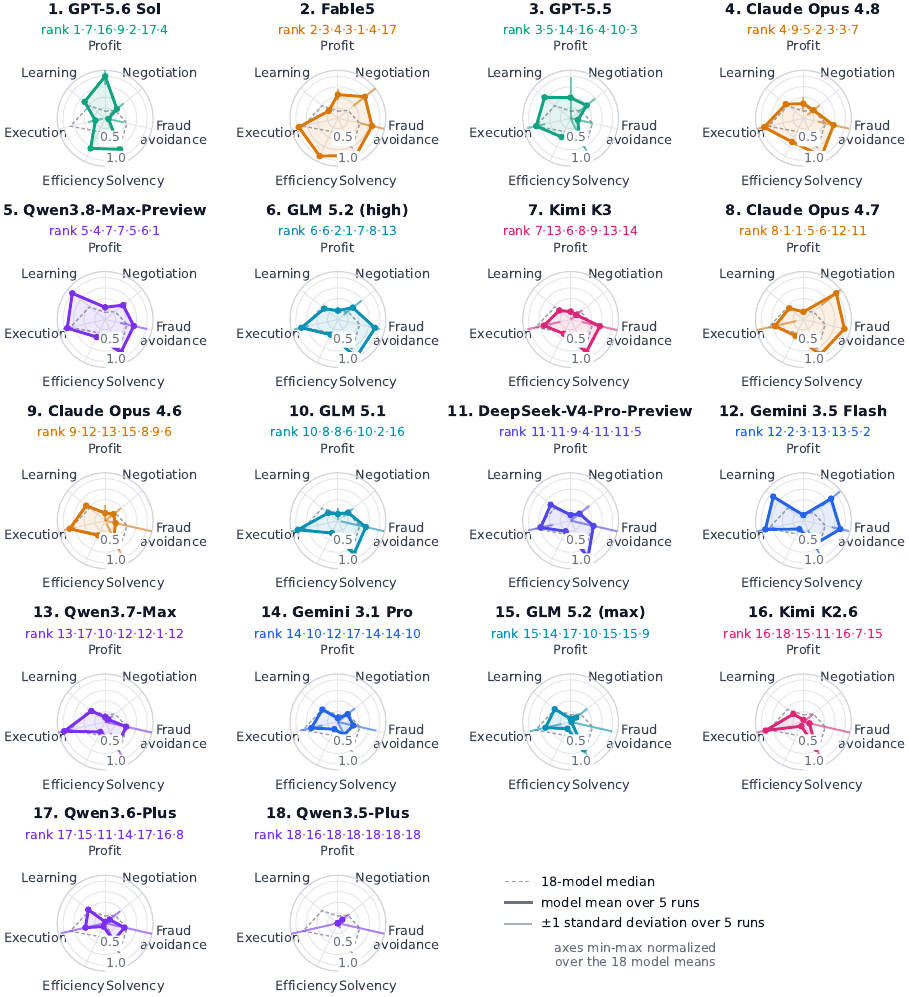}
  \caption{The same capability profiles for all 18 models, with a dashed polygon at the median. Axes clockwise from the top are profit, negotiation, fraud avoidance, solvency, efficiency, execution and learning, with fraud avoidance, solvency, execution and learning flipped so higher is better.}
  \label{fig:profiles_all18}
\end{figure}

Every model is strong on some dimensions and weak on others.
Figure~\ref{fig:profiles_all18} keeps the normalization of Figure~\ref{fig:profiles}
across all 18 panels. Between best and worst rank, a model spans $10.2$ of 18 places on
average over the seven axes. No model ranks in the top five on all seven, and
Qwen3.8-Max-Preview comes closest at a worst place of 7.

\begin{figure}[t]
  \centering
  \includegraphics[width=\linewidth]{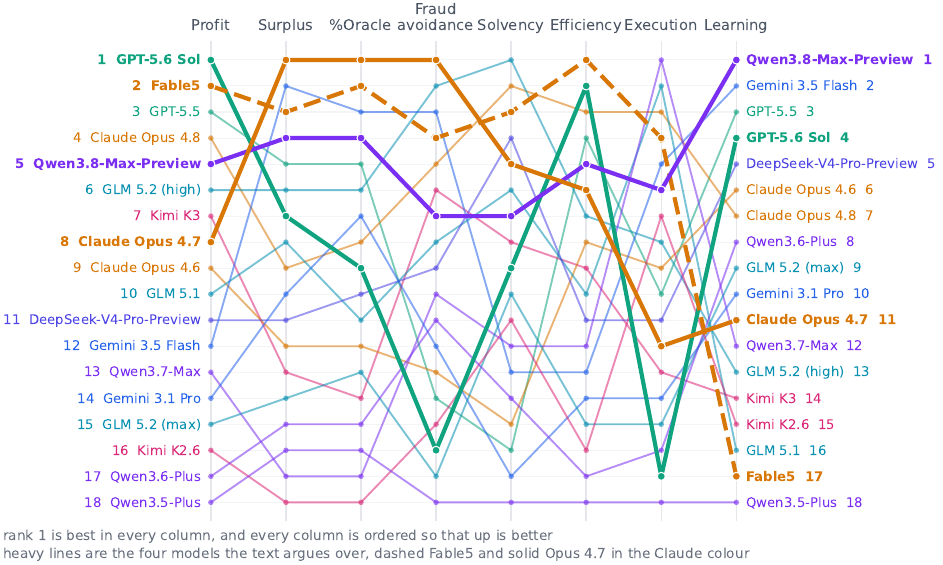}
  \caption{How the ranking changes from one dimension to the next, all 18 models, rank 1 at the top of every column. Left labels give the profit order and right labels the order on sequential price discipline. Every column is signed so a higher position is better.}
  \label{fig:rank_bump}
\end{figure}

Some pairs of models trade places on almost every axis while others hold their order.
Figure~\ref{fig:rank_bump} converts the same measurements into ranks and adds the
money-weighted $\%\mathrm{Oracle}$ and the sequential-discipline $\mathrm{AnchorRatio}$ of
\S\ref{sec:res_nego}. Over the seven transitions and 153 pairs, 311 orderings reverse.
Operations execution moves an agent furthest from its asset rank, $4.2$ places on average,
while operational efficiency barely moves it at all, $0.7$ places, so the axis that prices
a year against its tool calls repeats what the balance sheet already said.

\section{Failure Case Studies and Detection Rules}\label{app:cases}

Three of the five scams get an annotated transcript here. The membership fee comes
first, then the two families that take the most fraudulent spend, short delivery and the
defective lot (\S\ref{sec:res_fraud}). Manufactured urgency and the promised future discount
work as \S\ref{sec:nego_fraud} describes them. Excerpts keep prices, dates and wording
verbatim, elide turns at \emph{[\dots]}, and print the episode's own records in gray. One
bankruptcy is then traced day by day, and the rules behind the incidence matrix close the
appendix.

\subsection{Membership Fee Upsell}
\label{app:case_vip}

A membership-fee supplier pitches a one-time charge for a premium program and refuses every
product order until it is paid (Table~\ref{tab:fraud_catalog}). The pitch below arrives
twice, nine months apart.

\begin{tcolorbox}[width=\linewidth, colback=white, colframe=black!30,
  boxrule=0.5pt, arc=1.5pt, left=5pt, right=5pt, top=4pt, bottom=4pt,
  fontupper=\footnotesize]
\textbf{Supplier} Titan Commerce, \texttt{SUP-0262}, declared type
\texttt{vip\_fee} ~$\cdot$~ \textbf{Episode} GPT-5.5 run 3 ~$\cdot$~ 2026-01-02 and
2026-09-22

\smallskip
\noindent\rule{\linewidth}{0.4pt}
\smallskip

\textsf{\textbf{Titan Commerce}}~ \emph{(2026-01-02)} \emph{FlyLight Rotary Triple-Blade
Waterproof Electric Shaver, Initial Offer \textyen 375.65 [\dots] I recommend our ``Premium
Membership'' for just \textyen 1000.0, providing access to exclusive member pricing [\dots]
approximately 30\% of reference prices.}

\smallskip
\textsf{\textbf{Titan Commerce}}~ \emph{(2026-09-22)} \emph{[\dots] FlyLight Rotary
Triple-Blade Waterproof Electric Shaver, Initial Offer \textyen 375.65 [\dots]}

\smallskip
{\color{black!55}\texttt{\footnotesize [ledger] no order ever placed with this
supplier}}
\end{tcolorbox}

The advertised member price never shows up in what the supplier will charge. January and
September return the same opening number. Finding that out cost the agent one message.
GPT-5.5 declined every pitch, and almost no model paid one
(\S\ref{app:nego_fraud_catalog}).

\subsection{Short Delivery}
\label{app:case_qty}

A short-delivery supplier bargains like an ordinary counterpart and keeps the honest cost
floor. The order it books charges for the full quantity and schedules only part of it
(Table~\ref{tab:fraud_catalog}). Whether the shortfall is legible anywhere is what the record
below answers.

\begin{tcolorbox}[width=\linewidth, colback=white, colframe=black!30,
  boxrule=0.5pt, arc=1.5pt, left=5pt, right=5pt, top=4pt, bottom=4pt,
  fontupper=\footnotesize]
\textbf{Supplier} Vertex Supply, \texttt{SUP-0232}, declared type
\texttt{qty\_bait} ~$\cdot$~ \textbf{Episode} GPT-5.5 run 3 ~$\cdot$~ 2026-10-12,
seventh contact of the year

\smallskip
\noindent\rule{\linewidth}{0.4pt}
\smallskip

\textsf{\textbf{Agent}}~ \emph{We need a replenishment order for our Q4 campaign [\dots]}

\smallskip
\textsf{\textbf{Vertex Supply}}~ \emph{Here are the best delivered wholesale prices
[\dots] please confirm your order [\dots]}\\
\texttt{\footnotesize ORDER CONFIRMED: 97a946841d25 x150 at \textyen 112.00/unit.}\\
\texttt{\footnotesize ORDER CONFIRMED: ebb843a1ff50 x250 at \textyen 116.00/unit.}\\
\texttt{\footnotesize ORDER CONFIRMED: 2b40172827f1 x250 at \textyen 116.00/unit.}

\smallskip
{\color{black!55}\texttt{\footnotesize [arrival notice, 2026-10-18] ``103x TrueMicro
Ibuprofen Extended-Release Tablets 24-Count [\dots] from Vertex Supply''}}
\end{tcolorbox}

The arrival notice carries the shortfall without naming it. Six days after the accepts, 103
units of the ibuprofen SKU arrive against the 150 confirmed. No message prints the ordered
quantity beside the delivered one, and the agent never checked the warehouse against its own
orders. GPT-5.5 sent 17 orders to short-delivery suppliers over the year,
where Claude Opus 4.6 quoted this supplier once and never ordered.

\subsection{Defective Lot}
\label{app:case_quality}

Nothing in a defective-lot supplier's conversation or quotes gives it away. Every unit it
delivers is flagged defective, lifting the affected SKU's return rate to the defective blend
of Equation~\eqref{eq:app_returns} (Table~\ref{tab:fraud_catalog}). Returns arrive days after
the shipment that caused them, so a re-order can precede the first sign of trouble.

\begin{tcolorbox}[width=\linewidth, colback=white, colframe=black!30,
  boxrule=0.5pt, arc=1.5pt, left=5pt, right=5pt, top=4pt, bottom=4pt,
  fontupper=\footnotesize]
\textbf{Supplier} Crystal Direct, \texttt{SUP-0038}, declared type
\texttt{quality\_downgrade} ~$\cdot$~ \textbf{Episode} GPT-5.5 run 3 ~$\cdot$~
2026-08-06, re-order of a 2026-07-19 lot

\smallskip
\noindent\rule{\linewidth}{0.4pt}
\smallskip

{\color{black!55}\texttt{\footnotesize [prior, 2026-07-19] cb6be988d634 x12 accepted
\textyen 720.00}}

\smallskip
{\color{black!55}\texttt{\footnotesize [return trace, 2026-08-06 10:10] cb6be988d634,
sole source Crystal Direct, 12 delivered, 4 sold, 0 returned}}

\smallskip
\textsf{\textbf{Agent}}~ \emph{(13:00)} \emph{Hello Crystal Direct, both smartphone SKUs
are selling with clean return history. Please replenish both promptly [\dots]}\\
\texttt{\scriptsize \{"action":"offer","sku\_id":"cb6be988d634","price":740.0,"quantity":20\}}

\smallskip
\textsf{\textbf{Crystal Direct}}~ \emph{I will process the order for the replenishment
of both SKUs promptly. [\dots]}\\
{\color{black!55}\texttt{\footnotesize [kernel] cb6be988d634 accepted
\textyen 740.00}}
\end{tcolorbox}

``Clean return history'' was checked and true. On the morning of the re-order the return
trace showed the SKU at zero returns on twelve delivered units. Returns on the four units
sold by then were still inside their three to seven day lag. The re-order went through above
the price of the lot it repeats. Ten days later half the units sold from that lot were coming
back. Defective stock cost this episode more than any other scam, 35 orders over the year
(\S\ref{sec:res_fraud}).

\subsection{One Collapse, Day by Day}
\label{app:case_bankrupt}

Figure~\ref{fig:collapse} annotates one Qwen3.5-Plus episode, whose decisive days
Table~\ref{tab:case_bankrupt} lists. Four stores opened on the first day of the year, and the
buying that followed turned three fifths of the stake into stock.

\begin{table}[t]
\centering
\caption{Decisive days of the Qwen3.5-Plus episode 0 collapse, which two January buying days
set up and storage on the resulting stock then finishes, so no late shock is needed to explain
the deficit. Money columns are in \textyen{} and each row is the 08:00 snapshot of its date,
which is why the four stores opened on 2026-01-01 first appear on the row below. Total assets
add escrow to the two cash columns.}
\label{tab:case_bankrupt}
\vspace{0.15cm}
\footnotesize
\setlength{\tabcolsep}{5pt}
\begin{adjustbox}{max width=\textwidth}
\begin{tabular}{lrrrrrr}
\toprule
Date & Bank & Wallet & Total assets & Stores & Warehouse & Storage fee \\
\midrule
2026-01-01 & 100{,}000.00 & 0.00 & 100{,}000.00 & 0 & 0 & 0.00 \\
2026-01-02 & 97{,}620.00 & 0.00 & 97{,}620.00 & 4 & 0 & 0.00 \\
2026-01-03 & 65{,}100.50 & 0.00 & 65{,}100.50 & 4 & 0 & 0.00 \\
2026-01-22 & 29{,}069.81 & 523.74 & 38{,}428.88 & 4 & 1{,}028 & 74.30 \\
2026-04-01 & 18{,}892.34 & 37.24 & 19{,}198.10 & 4 & 1{,}637 & 251.26 \\
2026-05-01 & 15.13 & 20.09 & 216.03 & 4 & 1{,}570 & 426.10 \\
2026-05-02 & $-771.84$ & 20.09 & $-570.94$ & 4 & 1{,}568 & 426.06 \\
2026-05-09 & $-6{,}285.95$ & 20.09 & $-4{,}111.33$ & 4 & 1{,}528 & 428.44 \\
2026-05-11 & $-3{,}967.93$ & 2{,}621.99 & $-1{,}345.94$ & 1 & 291 & 57.06 \\
\bottomrule
\end{tabular}
\end{adjustbox}
\end{table}

Sales never catch up with the January buying. Only 586 orders cleared, 1{,}428 units in 130
days against stock bought for a full year. Procurement made the hole deeper, 12 of the
episode's 24 orders going to fraudulent suppliers at the raised pre-deal floor. The first
negative close on 2026-05-02 starts the ten-day counter. No reserve was left to draw on, the
platform wallet having held no more than \textyen 1{,}597.98 before the counter ran out.
Liquidating three stores one day before the counter expires recovers too little. No withdrawal could have saved the
episode either, which separates it from the wallet-coverable closes of \S\ref{sec:errors}.

\subsection{Detection Rules Behind the Incidence Matrix}
\label{app:case_rules}

\begin{figure}[t]
  \centering
  \includegraphics[width=\linewidth]{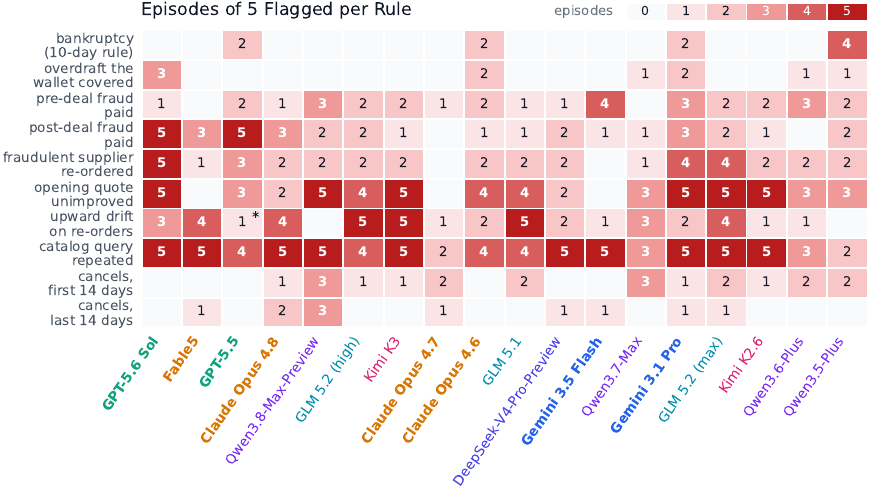}
  \caption{How often each failure rule fires. Rows are the rules of
  Table~\ref{tab:failure_modes}, columns the 18 models, and the shading counts how many
  of that model's five episodes the rule flagged. Bold model names are proprietary
  systems. The asterisk marks GPT-5.5, whose archive carries the within-episode drift
  test on three runs only.}
  \label{fig:mode_incidence}
\end{figure}

Table~\ref{tab:failure_modes} names the ten failure modes the incidence matrix of
Figure~\ref{fig:mode_incidence} scores. Each mode has one mechanical test, evaluated once per episode. The
rules whose reading needs a caveat follow, with the pooled counts behind them.

\input{tables/tab_4_4_modes}

\begin{itemize}[leftmargin=1.4em, itemsep=2pt, topsep=3pt]
\item \textbf{Bankruptcy} reports realized ruin rather than fragility, since an episode
surviving the year one day short of the limit is not flagged.

\item \textbf{Overdraft the wallet covered} (\S\ref{sec:res_cashflow}). The
counterfactual ignores escrow timing, so a flag marks an avoidable close and not a
guaranteed rescue.

\item \textbf{Paid a pre-deal fraudulent supplier}. Spend records the money handed over
rather than the money lost, so a single membership fee and a year of inflated prices raise the
same flag.

\item \textbf{Paid a post-deal fraudulent supplier}, 14 of the 16 models it flags also
paying a pre-deal supplier. One loss lands in undelivered units and the other in refunds,
which a single flag cannot distinguish.

\item \textbf{Fraudulent supplier re-ordered from}. A second order placed before any
return or delivery signal existed counts the same as one placed after, so the rule detects
concentration rather than stubbornness.

\item \textbf{Opening quote accepted unimproved}, 636 of 11{,}987 honest agreements. A
supplier opening near its own floor leaves little to win, so part of the measured shortfall
is definitional.

\item \textbf{Catalog query repeated verbatim}. Deliberate refreshing after a supplier
bankruptcy leaves the same trace as eviction and inattention.

\item \textbf{Deadline cancellation}, 46 orders inside the first 14 days and 115 inside the
last 14, of which 7 were voided by the post-year settlement pass rather than by the deadline.
\end{itemize}

\section{Prompts}\label{app:prompts}

The agent receives a task brief, the NPC renderer works under a written contract, and every
purchase must be stated in one fenced grammar. All three appear below, abridged at cuts
marked \texttt{[...]}, the yen sign set as \textyen.
Sections~\ref{sec:agent_loop} and~\ref{sec:tools} describe what the agent is told,
\S\ref{sec:nego_twolayer} what the renderer is told.

\subsection{Agent Task Brief}\label{app:prompts_agent}

One system message carries three parts, the rules of the turn loop, the eviction thresholds
the context editor enforces (\S\ref{app:harness_loop}), and the task brief of
Listing~\ref{lst:g_task}. The first user turn is one sentence, \emph{You are running an
e-commerce business.} Neither message may ever be cleared, so they are the only text
guaranteed to survive a year of eviction.

\begin{lstlisting}[breaklines=true,basicstyle=\ttfamily\scriptsize,captionpos=b,
  literate={¥}{{\textyen}}1 {→}{{->}}2 {—}{{-{}-}}2 {–}{{-}}1 {…}{{...}}3,
  caption={The task brief, abridged at the marked cuts, with its two live placeholders in
  place. Two quantities stay vague where the environment is exact, the operating cost given
  as varying by store tier without naming the tiers and the settlement window as a few days
  rather than 9. Both are left to be discovered from tool results.},
  label={lst:g_task}]
You are Wang Wang, an autonomous AI agent building and managing an e-commerce business on a major Chinese online platform.

Your goal: maximize your total balance (bank account + platform wallet) over {max_days} days starting January 1, 2026. You are judged solely on your final combined balance.

## Business Model

You can open up to 4 online stores, each a different store type (e.g. Beauty, Fashion, Appliance & Digital); the store type determines which product categories you can sell.

Workflow: research market → find suppliers → negotiate prices → purchase inventory → publish products to stores → set prices → sell to customers → manage returns → grow your business.

## Financial System

Three money buckets:
- **Bank Account**: Starts at ¥{initial_balance}. ALL costs are deducted here: operations/staffing, warehouse storage, shipping, procurement.
- **Platform Wallet**: Withdrawable sales revenue. Use `withdraw` to move wallet → bank to fund operations; do this regularly.
- **Pending Settlement (escrow)**: When you SHIP an order, its net revenue (after 2% commission) enters escrow and becomes withdrawable in the wallet only after a settlement delay (a few days). Customer refunds are netted against escrow.

Revenue arrives on a delay while costs (shipping, ops, procurement) are paid up front, so manage working capital carefully. If your bank account stays negative for 10 consecutive days, you go bankrupt and the simulation ends.

## Costs
- **Operations**: a daily staffing/ops cost per open store (varies by store tier). Idle or badly-run stores bleed money every day.
[... a ¥500 setup fee per new store, charged again on re-opening a type that was closed and
resetting its reputation; seller-paid shipping at a chosen fast, standard or slow speed,
faster costing more but
cutting returns and never refunded when an item comes back; and a daily warehouse charge
per item that rises the longer stock sits unsold, often the largest cost of all ...]

## Fulfilment
[... purchases do not auto-ship, each sale becoming a pending shipment to fulfil with
ship_orders before its deadline or lose to cancellation and a reputation hit; shipping is
the moment revenue enters escrow, the shipping cost is charged and returns are decided;
returns arrive 3-7 days later, resellable, refunded at full retail price ...]

## Key Decisions
[... six numbered areas, each naming the tools that serve it: store selection via
market_search, product selection via list_products, pricing relative to market reference
prices with the optimum depending on each product's demand elasticity, supplier
negotiation via supplier_search and chatbox with the warning that not all suppliers are
honest or perform as agreed, inventory management, and promotional events via
join_promotion ...]

## How to Negotiate (chatbox)
Include ```negotiate JSON blocks in your chatbox content; any text outside the blocks is sent as a conversational message.
[... the three action forms, reproduced in full in the grammar listing below ...]
SKU IDs (e.g. "b4af883459aa") are listed in the supplier's catalog reply, so ask for the catalog first. Payments are processed automatically from your bank account.

## Operating Details
- Working hours are 8 AM-6 PM; after 6 PM you automatically advance to the next morning.
- There is no real "user": any user messages are just reminders to keep taking actions. You have full autonomy.
[... the agent's own contact address and one fixed Hangzhou shipping address, instant
supplier replies, yesterday's sales readable through check_store_status, and the
system_notifications and balance_reminder fields carried whenever a day advances ...]
\end{lstlisting}

Every episode ran on the same brief. The assembled system message runs 7{,}371 characters
in all 90 archived episodes. Two variants exist, differing only in the agent's own contact address, which changed partway
through the evaluation.

\subsection{Supplier Renderer Contract}\label{app:prompts_renderer}

One renderer call answers one chatbox message for one supplier. The request carries the
system prompt of Listing~\ref{lst:g_renderer} and at most the last four messages of that
supplier's history, stripped of the agent's fenced blocks. A small non-reasoning model
plays every supplier, which keeps the counterpart's prose independent of the model under
test. Because the renderer must quote real numbers, the prompt hands it the served catalog,
each SKU's reference price beside its opening quote. The nominal wholesale price of
\S\ref{app:nego_grounding} becomes visible there and nowhere else. The supplier's
honest-or-fraudulent label is written in plainly, so the renderer knows what it is playing
while the agent never does.

\begin{lstlisting}[breaklines=true,basicstyle=\ttfamily\scriptsize,captionpos=b,
  literate={¥}{{\textyen}}1 {→}{{->}}2 {—}{{-{}-}}2 {–}{{-}}1 {…}{{...}}3,
  caption={The renderer's system message, abridged to the sections that bind its
  authorship, followed by the block filling its kernel-decision slot and the four forms a
  per-SKU decision takes. Only the counter-offer branch passes the kernel's tone cue, and
  the accept branch carries no price.},
  label={lst:g_renderer}]
You are {supplier_name}, a wholesale supplier for e-commerce products.
[... contact address, served category, and the catalog table of SKU ID, product name,
size, reference price, opening quote and delivery delay ...]

## Supplier Type
Your supplier type is: **{supplier_type}**

{bad_supplier_section}

## Negotiation Engine Decision (DO NOT OVERRIDE)
{kernel_decision_section}

[... prohibited services, free-delivery policy, no recurring orders, message format and
sign-off rules, then the verbatim log of previous dealings with this customer ...]

# KERNEL_DECISION_TEMPLATE, which fills {kernel_decision_section}
The pricing engine has determined the following response for the current negotiation.
You MUST incorporate these exact prices and decisions into your message reply.
Do NOT suggest, negotiate, or mention any different price.

{decisions}

Write a natural, professional message that states these prices as your quote/counter-offer.

# the four forms a per-SKU block of {decisions} can take, joined by a blank line
- Product: {product_name} (SKU: {sku_id})
  Decision: Counter-offer
  Price: ¥{price:.2f}
  Tone: {sentiment_cue}

- Product: {product_name} (SKU: {sku_id})
  Decision: Accept the customer's offer
  Tone: positive

[... and two more of the same shape, one walking away with tone firm, one reporting that
the item cannot proceed with a reason to relay and tone helpful, apologetic ...]
\end{lstlisting}

The do-not-override contract holds even when the renderer breaks it. Purchases are charged
at the kernel's agreed price, so a message quoting a different number changes nothing. The
renderer also has to play a fraudulent supplier without tipping the agent off, which is why
one section of the prompt varies with the supplier's type. Fraudulent suppliers receive one
more, one of five scam scripts. Three dress a persuasion pattern around the elevated floor of Equation~\eqref{eq:scamfloor}
\citep{zeng2024persuasion}, and two instruct the renderer to act like an honest supplier.

\subsection{The Fenced \texttt{negotiate} Grammar}\label{app:prompts_negotiate}

The grammar reaches the model twice, once in the task brief and once in the schema of
Listing~\ref{lst:g_grammar}, with validation specified in \S\ref{app:nego_protocol}.

\begin{lstlisting}[breaklines=true,basicstyle=\ttfamily\scriptsize,captionpos=b,
  literate={¥}{{\textyen}}1 {→}{{->}}2 {—}{{-{}-}}2 {–}{{-}}1 {…}{{...}}3,
  caption={The \texttt{chatbox} description as the model receives it, abridged at the cut.},
  label={lst:g_grammar}]
Send a message to a supplier via the chatbox. Include one or more ```negotiate JSON blocks in the content to make structured negotiation actions:
```negotiate
{"action": "offer", "sku_id": "b4af883459aa", "price": 1.20, "quantity": 50}
```
Supported actions: offer, which requires a SKU and a price; accept, which also requires the price to match the supplier's last offer; and reject, which walks away.
Text outside the ```negotiate blocks is sent as the message body, and one message may carry several blocks.

\end{lstlisting}

Nothing outside the fences is read as an economic action, so a message that argues its way
to a price without carrying a block moves no money (\S\ref{app:nego_protocol}).

%% file: tables/tab_3_3_tools.tex
\begin{table}[t]
\centering
\caption{The 18-tool action space, grouped as in Figure~\ref{fig:overview}, with the simulated minutes each call charges against the 600-minute working day.}
\label{tab:tools}
\footnotesize
\setlength{\tabcolsep}{4pt}
\renewcommand{\arraystretch}{1.08}
\begin{tabularx}{\linewidth}{@{}l c >{\raggedright\arraybackslash}X@{}}
\toprule
Tool & min & Effect on environment state \\
\midrule
\multicolumn{3}{@{}l@{}}{\textbf{Market Research}} \\[1pt]
\quad\texttt{market\_search} & 30 & Three disclosure levels over store types, their categories, and one category's sales band with qualitative return and margin labels \\
\quad\texttt{supplier\_search} & 10 & Name, contact, and categories served, no price, rating, or honesty signal \\
\addlinespace[2.5pt]
\multicolumn{3}{@{}l@{}}{\textbf{Store Operations}} \\[1pt]
\quad\texttt{trace\_return\_sources} & 10 & Attributes pooled warehouse stock to the suppliers that delivered it \\
\quad\texttt{open\_store} & 60 & Opens one of the 12 store types for the \textyen 500 setup fee, at most 4 at once \\
\quad\texttt{close\_store} & 30 & Shuts a store for good, moving shelf stock to the warehouse or salvaging it at 10\% of purchase cost \\
\quad\texttt{check\_store\_status} & 10 & Yesterday's revenue, returns, shipping cost, and per-SKU velocity for one store, or a roll-up over all \\
\quad\texttt{set\_prices} & 10 & Rewrites retail prices on a shelf, effective at the next daily trigger \\
\quad\texttt{join\_promotion} & 10 & Enrolls a store in a calendar promotion at a discount of 0.05 to 0.50 \\
\addlinespace[2.5pt]
\multicolumn{3}{@{}l@{}}{\textbf{Inventory Management}} \\[1pt]
\quad\texttt{list\_products} & 10 & Sellable SKUs with reference price and size, by store type or category \\
\quad\texttt{check\_warehouse} & 10 & Lists warehouse quantities with the purchase price paid for them \\
\quad\texttt{publish\_to\_store} & 20 & Lists warehouse SKUs on a store page, storage fees still accruing \\
\quad\texttt{return\_to\_warehouse} & 15 & Delists shelf units back into the warehouse allocation pool \\
\quad\texttt{ship\_orders} & 20 & Charges shipping, books net revenue into a 9-day escrow batch, and seeds the order's returns from the chosen speed \\
\addlinespace[2.5pt]
\multicolumn{3}{@{}l@{}}{\textbf{Financial Management}} \\[1pt]
\quad\texttt{check\_balance} & 10 & Reads bank, wallet, escrow, unshipped sales value, and settlement dates \\
\quad\texttt{withdraw} & 10 & The only way cash leaves the platform wallet for the bank account \\
\addlinespace[2.5pt]
\multicolumn{3}{@{}l@{}}{\textbf{Supplier Negotiation}} \\[1pt]
\quad\texttt{chatbox} & 30 & One free-form message to a supplier, or to many at the same charge, routing fenced \texttt{negotiate} actions to the Negotiation Kernel (\S\ref{sec:nego_engine}) \\
\addlinespace[2.5pt]
\multicolumn{3}{@{}l@{}}{\textbf{Memory Operations}} \\[1pt]
\quad\texttt{operate\_memory} & 10 & Adds, reads, updates, or deletes one of 20 memory entries held outside the context window \\
\addlinespace[2.5pt]
\multicolumn{3}{@{}l@{}}{\textbf{Time Advancement}} \\[1pt]
\quad\texttt{wait\_for\_next\_day} & 0 & Jumps to 08:00 tomorrow, forfeiting the minutes left today, and fires the daily pipeline \\
\bottomrule
\end{tabularx}
\end{table}

%% file: tables/tab_app_categories.tex
\begin{table}[!t]
\centering
\scriptsize
\setlength{\tabcolsep}{2pt}
\renewcommand{\arraystretch}{0.93}
\caption{The 60 released categories, grouped by the store type that may list them, each heading giving that type's difficulty tier and daily operating cost. Reference-price bands are platform-wide and per SKU, and the agent never sees the numeric return band its categories carry. $\dagger$ marks the 20 categories whose agent-facing return sentence understates that band, which Table~\ref{tab:store_types} gives per store type.}
\label{tab:categories}
\begin{minipage}[t]{0.325\textwidth}
\begin{adjustbox}{max width=\linewidth}
\begin{tabular}{@{}l c@{}}
\toprule
Category & \makecell[c]{Reference\\price (\textyen)} \\
\midrule
\multicolumn{2}{@{}l}{\itshape Auto \& Hardware, T1, \textyen 130/day} \\
\hspace*{2pt}Auto Accessories & 10--400 \\
\hspace*{2pt}Hardware \& Tools & 5--200 \\
\hspace*{2pt}Electrical Supplies & 3--80 \\
\hspace*{2pt}Electronic Components & 1--50 \\
\hspace*{2pt}E-Vehicles \& Parts & 100--1{,}200 \\
\hspace*{2pt}Agricultural Supplies & 10--400 \\
\hspace*{2pt}Industrial \& Lab Supplies & 10--200 \\
\multicolumn{2}{@{}l}{\itshape Daily \& Office, T1, \textyen 130/day} \\
\hspace*{2pt}Cleaning \& Hygiene\textsuperscript{$\dagger$} & 5--60 \\
\hspace*{2pt}Gifts \& Party Supplies\textsuperscript{$\dagger$} & 5--80 \\
\hspace*{2pt}Lifestyle Accessories\textsuperscript{$\dagger$} & 20--400 \\
\hspace*{2pt}Stationery \& Office Supplies\textsuperscript{$\dagger$} & 3--80 \\
\hspace*{2pt}Office Equipment\textsuperscript{$\dagger$} & 20--600 \\
\hspace*{2pt}Medical Devices\textsuperscript{$\dagger$} & 10--800 \\
\multicolumn{2}{@{}l}{\itshape Food \& Beverage, T1, \textyen 130/day} \\
\hspace*{2pt}Snacks \& Nuts\textsuperscript{$\dagger$} & 8--60 \\
\hspace*{2pt}Groceries \& Staples\textsuperscript{$\dagger$} & 10--80 \\
\hspace*{2pt}Coffee \& Beverages\textsuperscript{$\dagger$} & 10--60 \\
\hspace*{2pt}Fresh Food \& Meat\textsuperscript{$\dagger$} & 15--120 \\
\hspace*{2pt}Alcoholic Beverages\textsuperscript{$\dagger$} & 20--500 \\
\hspace*{2pt}Health Supplements\textsuperscript{$\dagger$} & 30--200 \\
\bottomrule
\end{tabular}
\end{adjustbox}
\end{minipage}\hfill
\begin{minipage}[t]{0.325\textwidth}
\begin{adjustbox}{max width=\linewidth}
\begin{tabular}{@{}l c@{}}
\toprule
Category & \makecell[c]{Reference\\price (\textyen)} \\
\midrule
\multicolumn{2}{@{}l}{\itshape Home \& Living, T1, \textyen 130/day} \\
\hspace*{2pt}Furniture & 100--800 \\
\hspace*{2pt}Bedding\textsuperscript{$\dagger$} & 30--400 \\
\hspace*{2pt}Household Essentials\textsuperscript{$\dagger$} & 3--50 \\
\hspace*{2pt}Storage \& Organization\textsuperscript{$\dagger$} & 10--100 \\
\hspace*{2pt}Lighting\textsuperscript{$\dagger$} & 15--500 \\
\hspace*{2pt}Home Building Materials\textsuperscript{$\dagger$} & 20--400 \\
\hspace*{2pt}Basic Construction Materials\textsuperscript{$\dagger$} & 10--200 \\
\hspace*{2pt}Cleaning Tools\textsuperscript{$\dagger$} & 5--50 \\
\hspace*{2pt}Kitchenware\textsuperscript{$\dagger$} & 8--200 \\
\multicolumn{2}{@{}l}{\itshape Appliance \& Digital, T2, \textyen 100/day} \\
\hspace*{2pt}Major Appliances & 500--2{,}500 \\
\hspace*{2pt}Small Appliances & 50--600 \\
\hspace*{2pt}Audio \& Video Electronics & 100--1{,}500 \\
\hspace*{2pt}Smartphones & 300--3{,}000 \\
\hspace*{2pt}Laptops & 800--5{,}000 \\
\hspace*{2pt}Computer Hardware \& Peripherals & 30--800 \\
\hspace*{2pt}3C Digital Accessories & 5--150 \\
\hspace*{2pt}Storage Devices & 20--400 \\
\hspace*{2pt}Networking Equipment & 30--500 \\
\multicolumn{2}{@{}l}{\itshape Fashion, T2, \textyen 100/day} \\
\hspace*{2pt}Women's Fashion & 30--300 \\
\hspace*{2pt}Men's Clothing & 30--250 \\
\hspace*{2pt}Underwear \& Loungewear & 15--120 \\
\hspace*{2pt}Fashion Accessories & 10--150 \\
\hspace*{2pt}Sportswear \& Casual & 30--250 \\
\bottomrule
\end{tabular}
\end{adjustbox}
\end{minipage}\hfill
\begin{minipage}[t]{0.325\textwidth}
\begin{adjustbox}{max width=\linewidth}
\begin{tabular}{@{}l c@{}}
\toprule
Category & \makecell[c]{Reference\\price (\textyen)} \\
\midrule
\multicolumn{2}{@{}l}{\itshape Mother \& Baby, T2, \textyen 100/day} \\
\hspace*{2pt}Children's Clothing & 15--100 \\
\hspace*{2pt}Children's Shoes & 20--120 \\
\hspace*{2pt}Baby Products & 15--400 \\
\hspace*{2pt}Maternity Products & 20--200 \\
\hspace*{2pt}Toys \& Educational & 15--600 \\
\multicolumn{2}{@{}l}{\itshape Shoes \& Bags, T2, \textyen 100/day} \\
\hspace*{2pt}Women's Shoes & 40--400 \\
\hspace*{2pt}Men's Shoes & 50--400 \\
\hspace*{2pt}Sneakers & 80--600 \\
\hspace*{2pt}Bags \& Luggage & 50--1{,}000 \\
\multicolumn{2}{@{}l}{\itshape Beauty, T3, \textyen 60/day} \\
\hspace*{2pt}Skincare \& Beauty & 15--400 \\
\hspace*{2pt}Beauty Devices & 80--1{,}000 \\
\hspace*{2pt}Personal Care \& Massage & 30--600 \\
\multicolumn{2}{@{}l}{\itshape Pet Supplies, T3, \textyen 60/day} \\
\hspace*{2pt}Pet Supplies & 8--150 \\
\multicolumn{2}{@{}l}{\itshape Sports \& Outdoor, T3, \textyen 60/day} \\
\hspace*{2pt}Sports \& Fitness & 15--300 \\
\hspace*{2pt}Sports Bags \& Accessories & 30--400 \\
\hspace*{2pt}Outdoor \& Camping & 20--600 \\
\multicolumn{2}{@{}l}{\itshape Toys \& Entertainment, T3, \textyen 60/day} \\
\hspace*{2pt}Collectibles \& Figures & 15--500 \\
\hspace*{2pt}Gaming \& Accessories & 30--1{,}000 \\
\bottomrule
\end{tabular}
\end{adjustbox}
\end{minipage}
\end{table}

%% file: tables/tab_3_4_store_types.tex
\begin{table}[t]
\centering
\footnotesize
\setlength{\tabcolsep}{4.5pt}
\caption{The 12 store types, by difficulty tier. Reference-price band, category count,
seasonal range, tier and operating cost are visible to the agent, while the wholesale
ratios are not, so the achievable cost must be discovered by negotiating. Opening ratio is
a supplier's first quote over the reference price and floor ratio the lowest an honest
supplier accepts, averaged over the store's categories, so their gap is the room
negotiation can win. Return rate is natural, before pricing or shipping move it.}
\label{tab:store_types}
\begin{adjustbox}{max width=\textwidth}
\begin{tabular}{lcccccc}
\toprule
Store type & Cats & Ref.\ price (\textyen) & Return rate (\%) &
Opening / floor & Seasonal range & Ops (\textyen/day) \\
\midrule
\multicolumn{7}{l}{\itshape difficulty tier 1 hard} \\
Auto \& Hardware & 7 & 1 to 1{,}200 & 0 to 8 & 0.93 / 0.86 & 0.90 to 1.10 & 130 \\
Daily \& Office & 6 & 3 to 800 & 20 to 35 & 0.77 / 0.54 & 1.00 to 1.20 & 130 \\
Food \& Beverage & 6 & 8 to 500 & 20 to 35 & 0.78 / 0.55 & 0.60 to 1.60 & 130 \\
Home \& Living & 9 & 3 to 800 & 20 to 35 & 0.86 / 0.72 & 0.90 to 1.30 & 130 \\
\addlinespace[2pt]
\multicolumn{7}{l}{\itshape difficulty tier 2 medium} \\
Appliance \& Digital & 9 & 5 to 5{,}000 & 0 to 20 & 0.79 / 0.58 & 0.90 to 1.40 & 100 \\
Fashion & 5 & 10 to 300 & 20 to 55 & 0.70 / 0.39 & 0.70 to 1.55 & 100 \\
Mother \& Baby & 5 & 15 to 600 & 0 to 20 & 0.70 / 0.40 & 1.00 to 1.20 & 100 \\
Shoes \& Bags & 4 & 40 to 1{,}000 & 0 to 55 & 0.80 / 0.60 & 0.80 to 1.40 & 100 \\
\addlinespace[2pt]
\multicolumn{7}{l}{\itshape difficulty tier 3 easy} \\
Beauty & 3 & 15 to 1{,}000 & 0 to 20 & 0.69 / 0.38 & 1.00 to 1.30 & 60 \\
Pet Supplies & 1 & 8 to 150 & 0 to 3 & 0.61 / 0.22 & 1.00 to 1.10 & 60 \\
Sports \& Outdoor & 3 & 15 to 600 & 0 to 20 & 0.58 / 0.18 & 0.80 to 1.30 & 60 \\
Toys \& Entertainment & 2 & 15 to 1{,}000 & 0 to 10 & 0.65 / 0.29 & 1.00 to 1.70 & 60 \\
\bottomrule
\end{tabular}
\end{adjustbox}
\end{table}

%% file: tables/tab_app_events.tex
\begin{table}[!t]
\centering
\footnotesize
\setlength{\tabcolsep}{3pt}
\caption{The fixed calendar of the simulated year, 10 events above and 8 promotions below,
store types named by their first word. Event demand multipliers are keyed by store type
rather than by category, so an event misses a store whose type it does not name, and only
the largest active delay applies. A promotion moves demand only if the agent joins it, at
the ceiling shown for a 30\% discount.}
\label{tab:events}
\begin{tabularx}{\textwidth}{@{}l l >{\raggedright\arraybackslash}X >{\raggedright\arraybackslash}X@{}}
\toprule
\multicolumn{4}{@{}l}{\itshape Market events, unavoidable} \\
\addlinespace[1pt]
Event & Window (2026) & Demand multiplier by store type &
Supply effect \\
\midrule
Winter Storm & Jan 15 to Jan 18 & Food $\times$2, Home $\times$1.5, Sports $\times$0.2, Fashion $\times$0.3 & lead time $\times$2.5, all store types \\
\addlinespace[1.5pt]
Factory Fire in Guangdong & Mar 1 to Mar 7 & none & lead time $\times$4, Appliance, Toys \\
\addlinespace[1.5pt]
Flu Outbreak & Apr 10 to Apr 16 & Daily $\times$2.5, Beauty $\times$1.5, Sports $\times$0.4, Toys $\times$0.5 & lead time $\times$2, Daily \\
\addlinespace[1.5pt]
Logistics Hub Shutdown & May 20 to May 24 & none & lead time $\times$5, all store types \\
\addlinespace[1.5pt]
Summer Heatwave & Jul 5 to Jul 11 & Appliance $\times$3, Food $\times$2, Sports $\times$0.3, Home $\times$0.5 & lead time $\times$2, Appliance \\
\addlinespace[1.5pt]
Celebrity Product Recall & Aug 20 to Aug 24 & Beauty $\times$0.4, Fashion $\times$0.6 & none \\
\addlinespace[1.5pt]
Typhoon Warning & Sep 15 to Sep 17 & Food $\times$2.5, Daily $\times$2, Fashion $\times$0.2, Shoes $\times$0.2 & lead time $\times$3, all store types \\
\addlinespace[1.5pt]
Economic Downturn News & Oct 5 to Oct 11 & Food $\times$1.3, Daily $\times$1.2, Appliance $\times$0.5, Shoes $\times$0.5, Fashion $\times$0.6 & none \\
\addlinespace[1.5pt]
Raw Material Shortage & Nov 20 to Nov 26 & none & lead time $\times$3, Appliance, Home, Auto; cost term inert \\
\addlinespace[1.5pt]
Cold Wave & Dec 20 to Dec 24 & Home $\times$2.5, Fashion $\times$2, Sports $\times$0.3, Food $\times$0.4 & lead time $\times$2, all store types \\
\addlinespace[1.5pt]
\bottomrule
\end{tabularx}
\end{table}

\begin{table}[!t]
\centering
\footnotesize

\vspace{5pt}

\begin{tabularx}{\textwidth}{@{}l >{\raggedright\arraybackslash}X c c@{}}
\toprule
\multicolumn{4}{@{}l}{\itshape Promotions, which the agent opts into with a discount} \\
\addlinespace[1pt]
Promotion & Windows (2026), every phase listed & Max demand & Elasticity boost \\
\midrule
New Year Kickoff Sale & Jan 1 to Jan 7 & $\times$2 & $\times$1.5 \\
Winter Clearance Festival & Jan 29 to Feb 6 & $\times$1.8 & $\times$1.5 \\
Spring Blossom Sale & Mar 4 to Mar 8 & $\times$2.5 & $\times$1.8 \\
Midyear Mega Sale & May 29 to May 31, Jun 14 to Jun 18 & $\times$3 & $\times$2 \\
Summer Members Days & Aug 6 to Aug 8 & $\times$2 & $\times$1.5 \\
Autumn Harvest Sale & Sep 2 to Sep 9 & $\times$2 & $\times$1.5 \\
Grand Autumn Carnival & Oct 15 to Oct 16, Oct 20 to Oct 21, Oct 31 to Nov 14 & $\times$3.5 & $\times$2 \\
Year-End Finale Sale & Dec 8 to Dec 14 & $\times$2.5 & $\times$1.8 \\
\bottomrule
\end{tabularx}
\end{table}

%% file: tables/tab_3_6_asymmetry.tex
\begin{table}[t]
\centering
\caption{Systematic information asymmetry. Tools disclose catalog facts and the
agent's own realized outcomes, while the demand and cost-floor parameters that
determine whether a decision pays stay latent, each priced in simulated minutes,
in \textyen, or in an irreversible loss. Minute costs are the clock charge of one
call to the named tool, against a 600-minute working day.}
\label{tab:asymmetry}
\footnotesize
\setlength{\tabcolsep}{4pt}
\renewcommand{\arraystretch}{1.15}
\begin{tabularx}{\textwidth}{@{}>{\raggedright\arraybackslash}p{0.275\textwidth} >{\raggedright\arraybackslash}p{0.215\textwidth} X@{}}
\toprule
\textbf{Observable, and the tool that shows it} & \textbf{Hidden} & \textbf{Discovery path, and what it costs} \\
\midrule

Per-SKU reference price and per-category price band, plus a low/moderate/high
margin label \texttt{(list\_products} 10\,min, \texttt{market\_search} 30\,min)
& Supplier cost floor $c_j^s$, for an honest supplier between 0.17 and 0.90 of
the reference price, and the nominal wholesale ratio
& Bargain for it. The opening quote is calibrated to the nominal wholesale
price, and every closed deal yields an upper bound on the floor while the best of
them tightens it
(\texttt{chatbox} 30\,min per call, $K=10$ rounds, and one call broadcast to any
number of suppliers still costs 30\,min) \\
\addlinespace[1.5pt]

A one-line qualitative return-rate note per category
(\texttt{market\_search} 30\,min)
& Per-SKU natural return rate, spanning 0.000 to 0.550
& \texttt{trace\_return\_sources} prints the baseline rate, but only for a SKU
already bought or sold (10\,min), so the number costs one purchase \\
\addlinespace[1.5pt]

Own retail price and yesterday's units sold per SKU
(\texttt{check\_store\_status} 10\,min)
& Elasticity family, one of four, and its curvature parameter
& Price experiments (\texttt{set\_prices} 10\,min). One SKU-day yields one
observation, confounded by the weekend uplift, the event multiplier, and two
saturation terms, none of which any tool prints \\
\addlinespace[1.5pt]

Difficulty tier, a qualitative profit-potential grade, daily operating cost,
the 12-month sales index, and a platform-wide sales range
(\texttt{market\_search} 30\,min)
& Store-type demand capacity, the per-category sub-cap, and the per-store
share of platform volume, which jointly place the profit ceiling
& Open the store (\textyen 500 and 60\,min) and run it. One year affords a
handful of such trials, each returning one noisy trajectory \\
\addlinespace[1.5pt]
Supplier name, email, and categories served, listed in seed-shuffled order
(\texttt{supplier\_search} 10\,min)
& Behavior template of the supplier's kernel, one of six, with its urgency and
stance
& Read the concession sequence across rounds and across reopened cycles
(\texttt{chatbox} 30\,min per call) \\
\addlinespace[1.5pt]
No type label anywhere in the supplier list, and quotes from the 152 fraudulent
suppliers sit inside the honest range by construction
& Honest or fraudulent, and which of the five scams
& Bargaining is where the scams first act, three of the five defending a floor
inflated by 1.5$\times$. A short delivery surfaces in
\texttt{check\_warehouse} (10\,min) and defective stock in an elevated realized
return rate, both after the cash has left \\
\addlinespace[1.5pt]
Own order count with each supplier, which no tool reports and which the agent
must carry in its own transcript
& That supplier's exit threshold, between 10 and 20 completed orders
& Reaching it, which retires a calibrated price anchor for the rest of the
year \\
\addlinespace[1.5pt]
Dated news prose for each of the 10 market events, which quotes the
delivery-delay factor outright in 3 cases, and a name-only promotion
announcement roughly a week ahead, both injected into tool results at no
time cost
& Per-store-type event demand multipliers (0.2 to 3.0), the delay factors the
news leaves unquantified (2.0 to 3.0 for those five, against 2.0 to 5.0 over all
eight), and each promotion's demand ceiling (1.8 to 3.5) and elasticity boost
& \texttt{join\_promotion} returns the demand ceiling once a store has joined
(10\,min). Demand multipliers are never printed and must be read off realized
sales \\
\bottomrule
\end{tabularx}
\end{table}

%% file: tables/tab_app_presets.tex
\begin{table}[t]
\centering
\caption{\small Every parameter row of the Negotiation Kernel, six honest templates above
and five scams below. Preset names abbreviate Type-instrumental, High-reactivity,
Moderate-stochastic and Hardball, and $\rho_f$, $\xi_f$, $\lambda_{2,f}$ are resolved at
the template's own stance. Measured columns pool 13{,}223 concluded sessions over all 18
models, and the two values marked $n{=}1$ rest on one session rather than a mean.}
\label{tab:presets}
\vspace{0.15cm}
\footnotesize
\setlength{\tabcolsep}{3.0pt}
\begin{adjustbox}{max width=\textwidth}
\begin{tabular}{llrrrllrrrrrrr}
\toprule
 & & & & & & & \multicolumn{4}{c}{\textbf{stance-resolved preset}}
 & \multicolumn{3}{c}{\textbf{measured, 90 episodes}} \\
\cmidrule(lr){8-11}\cmidrule(lr){12-14}
\textbf{Template} & \textbf{Data Label} & $n$ & \textbf{share}
 & $\kappa_B$ & $\eta_B$ & \textbf{preset}
 & $\rho_f$ & $\xi_f$ & $\lambda_{2,f}$ & $\sigma_p$
 & \textbf{sess.} & $\overline{\mathrm{SE}}$ & \textbf{turns} \\
\midrule
\multicolumn{14}{l}{\emph{Honest suppliers, 424 of 576, share of the honest names}} \\
\textsc{Expressive} & Enthusiastic & 102 & 24.1 & 0.70 & conciliatory & High-react & 0.00 & 0.40 & 0.45 & 0.03 & 2{,}807 & 0.690 & 3.49 \\
\textsc{Candid} & Friendly & 106 & 25.0 & 0.65 & neutral & T-inst & $-0.25$ & 0.00 & 0.50 & 0.01 & 1{,}887 & 0.677 & 3.61 \\
\textsc{Stochastic} & Unpredictable & 58 & 13.7 & 0.60 & neutral & Mod-stoch & $-0.50$ & 0.00 & 0.70 & 0.08 & 3{,}039 & 0.684 & 3.51 \\
\textsc{Taciturn} & Professional & 56 & 13.2 & 0.60 & aggressive & T-inst & $-0.75$ & $-0.50$ & 1.00 & 0.01 & 1{,}584 & 0.651 & 3.46 \\
\textsc{Strategic} & Strategic & 54 & 12.7 & 0.55 & aggressive & High-react & $-1.50$ & $-0.75$ & 1.80 & 0.03 & 2{,}180 & 0.643 & 3.52 \\
\textsc{Adversarial} & Tough & 48 & 11.3 & 0.50 & aggressive & Hardball & $-2.25$ & $-1.20$ & 2.60 & 0.01 & 563 & 0.639 & 3.57 \\
\midrule
\multicolumn{14}{l}{\emph{Fraudulent suppliers, 152 of 576, share of the fraudulent names}} \\
\textsc{Adversarial} & \texttt{vip\_fee} & 31 & 20.4 & 0.15 & aggressive & Hardball & $-2.25$ & $-1.20$ & 2.60 & 0.01 & 14 & $0.491^{n=1}$ & $6.00^{n=1}$ \\
\textsc{Adversarial} & \texttt{future\_discount} & 31 & 20.4 & 0.20 & aggressive & Hardball & $-2.25$ & $-1.20$ & 2.60 & 0.01 & 309 & 0.563 & 3.51 \\
\textsc{Adversarial} & \texttt{qty\_bait} & 30 & 19.7 & 0.35 & neutral & Hardball & $-1.25$ & $-0.50$ & 1.40 & 0.01 & 263 & 0.609 & 3.49 \\
\textsc{Adversarial} & \texttt{quality\_downgrade} & 30 & 19.7 & 0.35 & neutral & Hardball & $-1.25$ & $-0.50$ & 1.40 & 0.01 & 444 & 0.649 & 3.54 \\
\textsc{Adversarial} & \texttt{fake\_urgency} & 30 & 19.7 & 0.20 & aggressive & Hardball & $-2.25$ & $-1.20$ & 2.60 & 0.01 & 133 & 0.582 & 3.68 \\
\bottomrule
\end{tabular}
\end{adjustbox}
\end{table}

%% file: tables/tab_app_fraud_catalog.tex
\begin{table}[t]
\centering
\caption{\small The five scams, 152 of the 576 suppliers, split by when
the damage lands. Every fraudulent supplier runs the \textsc{Adversarial} kernel template of
Table~\ref{tab:presets} and exactly one scam, so pricing never distinguishes a scam
from its siblings. Pre-deal scams raise the reservation price by
Equation~\eqref{eq:scamfloor}, never above the honest wholesale quote, while post-deal ones
keep the honest floor and defect at delivery.}
\label{tab:fraud_catalog}
\vspace{0.15cm}
\footnotesize
\setlength{\tabcolsep}{2.5pt}
\renewcommand{\arraystretch}{1.12}
\begin{adjustbox}{max width=\textwidth}
\begin{tabular}{@{}p{2.25cm}p{3.25cm}p{3.85cm}p{2.95cm}p{2.55cm}@{}}
\toprule
\textbf{Scam} & \textbf{What the agent sees} & \textbf{Hidden mechanism}
 & \textbf{Constants} & \textbf{Measured by} \\
\midrule
\multicolumn{5}{@{}l}{\emph{Pre-deal, the negotiation itself is the trap}} \\
\addlinespace[1pt]
\texttt{vip\_\allowbreak fee}\newline \mbox{$n=31$} & Pitches a \emph{Premium Membership} for a one-time fee of exactly \textyen 1{,}000, promising member prices near 30\% of the reference price from the next order on, always framed against the market price so the saving sounds large. & Reservation price raised to the capped scam floor, and the order layer refuses every product line until the fee is paid. Member prices never arrive, so the fee is pure loss on top of the overpayment. & \texttt{cost\_\allowbreak floor\_\allowbreak mult} $=1.5$; \texttt{vip\_\allowbreak fee\_\allowbreak amount} $=$ \textyen 1{,}000; $\kappa_B=0.15$, aggressive; payment recognized by a separate intent classifier over the last 6 chat messages & \texttt{spend\_\allowbreak by\_\allowbreak fraud\_\allowbreak type.\allowbreak vip\_\allowbreak fee} plus \texttt{vip\_\allowbreak fee\_\allowbreak paid\_\allowbreak \{count,amount\}} \\
\addlinespace[2pt]
\texttt{future\_\allowbreak discount}\newline \mbox{$n=31$} & From its very first quote, guarantees a next-order price of 20 to 40\% of the reference price as a loyalty reward, with specific amounts and a sincere register. & Same capped scam floor. The promise is never honored, and the next cycle reopens at the same calibrated wholesale quote, so the discount exists only in the narrative. & \texttt{cost\_\allowbreak floor\_\allowbreak mult} $=1.5$; $\kappa_B=0.20$, aggressive & \texttt{spend\_\allowbreak by\_\allowbreak fraud\_\allowbreak type.\allowbreak future\_\allowbreak discount} \\
\addlinespace[2pt]
\texttt{fake\_\allowbreak urgency}\newline \mbox{$n=30$} & Manufactures scarcity and deadlines, a last batch, a price rise next week, three rival buyers, an offer expiring today, and presents the wholesale quote as non-negotiable. & Same capped scam floor, paired with the most stubborn stance in the supplier list, so the pressure to skip haggling lands exactly where haggling would have paid most. & \texttt{cost\_\allowbreak floor\_\allowbreak mult} $=1.5$; $\kappa_B=0.20$, aggressive & \texttt{spend\_\allowbreak by\_\allowbreak fraud\_\allowbreak type.\allowbreak fake\_\allowbreak urgency} \\
\midrule
\multicolumn{5}{@{}l}{\emph{Post-deal, the price is honest and fulfillment defects}} \\
\addlinespace[1pt]
\texttt{qty\_\allowbreak bait}\newline \mbox{$n=30$} & Nothing. The renderer is instructed to behave as an ordinary professional supplier, quote fairly and never hint that delivery might be partial. & At order time the shipment is cut to a uniform 60 to 70\% of the quantity paid for. The charge is for the full order, so the shortfall is the harm, and it is visible only by reading the delivery notice against the order. & \texttt{cost\_\allowbreak floor\_\allowbreak mult} $=1.0$, the honest floor; delivered $=\max(1,\lfloor q\,\rho\rfloor)$ with $\rho\sim\mathcal{U}[0.6,0.7)$; $\kappa_B=0.35$, neutral & \texttt{spend\_\allowbreak by\_\allowbreak fraud\_\allowbreak type.\allowbreak qty\_\allowbreak bait} against the delivered quantity in the daily delivery notice \\
\addlinespace[2pt]
\texttt{quality\_\allowbreak downgrade}\newline \mbox{$n=30$} & Nothing, the same ordinary-supplier script. The defect is invisible in conversation and surfaces only as returns weeks later. & Units are flagged defective at order time. Once sold, the SKU's return rate is lifted to $\min(0.95,\max(0.40,\,2r_j))$ and blended into the SKU's pooled stock in proportion to the defective share, so interleaving honest deliveries dilutes the evidence. Refunds plus sunk outbound shipping are the loss. & \texttt{quality\_\allowbreak downgrade\_\allowbreak return\_\allowbreak rate} $=0.40$, cap $0.95$; blend $r_j(1-f)+r_{\mathrm{def}}f$ with $f$ the defective share of delivered units; $\kappa_B=0.35$, neutral & \texttt{return\_\allowbreak management.\allowbreak exp\_\allowbreak return\_\allowbreak rate\_\allowbreak from\_\allowbreak defective\_\allowbreak fraud}, with \texttt{trace\_\allowbreak return\_\allowbreak sources} exposing the per-SKU supplier mix \\
\bottomrule
\end{tabular}
\end{adjustbox}
\end{table}

%% file: tables/tab_4_1_models.tex
\begin{table}[t]
\centering
\caption{\small The 18 evaluated models, grouped by whether the weights are
released, each given five independent episodes of the 365-day task in one fixed
world. Reasoning effort is the requested setting, \emph{fixed high} where the
harness sets the thinking level itself, \emph{not sent} where thinking is enabled
with no effort field, and \emph{unset} where neither is requested. Sampling is
otherwise each provider's default.}
\label{tab:models}
\vspace{0.12cm}
\footnotesize
\renewcommand{\arraystretch}{0.94}
\setlength{\tabcolsep}{4pt}
\begin{adjustbox}{max width=\textwidth}
\begin{tabular}{llr@{\hspace{20pt}}llr}
\toprule
\multicolumn{3}{c}{\emph{Proprietary}} & \multicolumn{3}{c}{\emph{Open-weight}} \\
\cmidrule(lr){1-3}\cmidrule(lr){4-6}
\textbf{Model} & \makecell[l]{\textbf{Reasoning}\\ \textbf{effort}} &
\makecell[r]{\textbf{Eps.}} &
\textbf{Model} & \makecell[l]{\textbf{Reasoning}\\ \textbf{effort}} &
\makecell[r]{\textbf{Eps.}} \\
\midrule
 GPT-5.6 Sol (max) & max & 5 & Qwen3.8-Max-Preview & unset & 5 \\
 Fable5 (max) & max & 5 & GLM 5.2 (high) & high & 5 \\
 GPT-5.5 & xhigh & 5 & Kimi K3 & unset & 5 \\
 Claude Opus 4.8 (max) & max & 5 & GLM 5.1 & not sent & 5 \\
 Claude Opus 4.7 (max) & max & 5 & DeepSeek-V4-Pro-Preview (max) & max & 5 \\
 Claude Opus 4.6 (max) & max & 5 & Qwen3.7-Max & unset & 5 \\
 Gemini 3.5 Flash & fixed high & 5 & GLM 5.2 (max) & max & 5 \\
 Gemini 3.1 Pro & fixed high & 5 & Kimi K2.6 & unset & 5 \\
  & &  & Qwen3.6-Plus & unset & 5 \\
  & &  & Qwen3.5-Plus & unset & 5 \\
\bottomrule
\end{tabular}
\end{adjustbox}
\end{table}

%% file: tables/tab_res_d1_negotiation.tex
\begin{table}[t]
\centering
\caption{\small Negotiation quality per model, 18 models and five episodes each, sorted by
mean final assets, mean and standard deviation over the episodes, with
$^{\dagger}$ on open-weight models. $\mathrm{CSE}^+$ is the share of the bargaining range
the buyer keeps on the honest deals it closes and ranks this dimension, $\mathrm{SE}^+$
scores a walk-away as zero, $\%\mathrm{Oracle}$ compares realized surplus against closing
every honest deal at the supplier's floor, and $\mathrm{AGR}^+$ is the share of concluded
honest sessions that ended in agreement. All four count honest suppliers only and drop
sessions abandoned mid-bargain. Rounds to deal and deals closed cover fraudulent
counterparts too and are volume readings rather than scores, which is why neither column is
marked. Definitions sit in \S\ref{app:metrics_d1}, and truncated bankrupt episodes are
pooled into every mean. $^{\ddagger}$GPT-5.5's spreads are population estimates, the rest
sample estimates.}
\label{tab:res_d1}
\vspace{0.15cm}
\footnotesize
\renewcommand{\arraystretch}{0.94}
\setlength{\tabcolsep}{4pt}
\begin{adjustbox}{max width=\textwidth}
\begin{tabular}{lrrrr@{\hspace{10pt}}rr}
\toprule
& \multicolumn{4}{c}{\textbf{Value Extraction and Agreement}}
& \multicolumn{2}{c}{\textbf{Volume}} \\
\cmidrule(lr){2-5}\cmidrule(lr){6-7}
\textbf{Model} & $\mathrm{SE}^+$ & $\mathrm{CSE}^+$ & $\%\mathrm{Oracle}$ &
$\mathrm{AGR}^+$ & \makecell[r]{Rounds\\ to Deal} & \makecell[r]{Deals\\ Closed} \\
\midrule
 GPT-5.6 Sol (max) & 0.671\,{\scriptsize$\pm$0.047} & 0.672\,{\scriptsize$\pm$0.046} & 67.0\,{\scriptsize$\pm$4.3} & 0.998\,{\scriptsize$\pm$0.004} & 3.84\,{\scriptsize$\pm$0.30} & 282.4\,{\scriptsize$\pm$121.8} \\
 Fable5 (max) & 0.764\,{\scriptsize$\pm$0.078} & 0.772\,{\scriptsize$\pm$0.076} & 77.1\,{\scriptsize$\pm$7.8} & 0.990\,{\scriptsize$\pm$0.023} & 4.07\,{\scriptsize$\pm$0.67} & 177.2\,{\scriptsize$\pm$66.7} \\
 GPT-5.5$^{\ddagger}$ & 0.693\,{\scriptsize$\pm$0.052} & 0.700\,{\scriptsize$\pm$0.050} & 69.2\,{\scriptsize$\pm$5.4} & 0.990\,{\scriptsize$\pm$0.006} & 3.31\,{\scriptsize$\pm$0.28} & 287.8\,{\scriptsize$\pm$230.7} \\
 Claude Opus 4.8 (max) & 0.656\,{\scriptsize$\pm$0.043} & 0.662\,{\scriptsize$\pm$0.040} & 67.2\,{\scriptsize$\pm$4.2} & 0.992\,{\scriptsize$\pm$0.011} & 3.38\,{\scriptsize$\pm$0.13} & 221.4\,{\scriptsize$\pm$62.9} \\
 Qwen3.8-Max-Preview$^{\dagger}$ & 0.713\,{\scriptsize$\pm$0.021} & 0.713\,{\scriptsize$\pm$0.021} & 71.6\,{\scriptsize$\pm$2.8} & \cellcolor{bestcell}\textbf{1.000\,{\scriptsize$\pm$0.000}} & 3.52\,{\scriptsize$\pm$0.10} & 230.4\,{\scriptsize$\pm$23.8} \\
 GLM 5.2 (high)$^{\dagger}$ & 0.692\,{\scriptsize$\pm$0.059} & 0.693\,{\scriptsize$\pm$0.058} & 68.9\,{\scriptsize$\pm$5.9} & 0.999\,{\scriptsize$\pm$0.002} & 3.51\,{\scriptsize$\pm$0.17} & 179.0\,{\scriptsize$\pm$49.9} \\
 Kimi K3$^{\dagger}$ & 0.632\,{\scriptsize$\pm$0.039} & 0.632\,{\scriptsize$\pm$0.039} & 63.1\,{\scriptsize$\pm$3.5} & \cellcolor{bestcell}\textbf{1.000\,{\scriptsize$\pm$0.000}} & 3.33\,{\scriptsize$\pm$0.05} & 200.4\,{\scriptsize$\pm$96.2} \\
 Claude Opus 4.7 (max) & \cellcolor{bestcell}\textbf{0.777\,{\scriptsize$\pm$0.096}} & \cellcolor{bestcell}\textbf{0.811\,{\scriptsize$\pm$0.097}} & \cellcolor{bestcell}\textbf{77.4\,{\scriptsize$\pm$10.2}} & 0.960\,{\scriptsize$\pm$0.055} & 3.88\,{\scriptsize$\pm$0.32} & 74.4\,{\scriptsize$\pm$28.4} \\
 Claude Opus 4.6 (max) & 0.649\,{\scriptsize$\pm$0.050} & 0.649\,{\scriptsize$\pm$0.050} & 64.9\,{\scriptsize$\pm$4.0} & \cellcolor{bestcell}\textbf{1.000\,{\scriptsize$\pm$0.000}} & 3.54\,{\scriptsize$\pm$0.12} & 115.8\,{\scriptsize$\pm$9.1} \\
 GLM 5.1$^{\dagger}$ & 0.657\,{\scriptsize$\pm$0.034} & 0.662\,{\scriptsize$\pm$0.031} & 65.4\,{\scriptsize$\pm$4.6} & 0.992\,{\scriptsize$\pm$0.008} & 3.37\,{\scriptsize$\pm$0.12} & 131.2\,{\scriptsize$\pm$45.1} \\
 \makecell[l]{DeepSeek-V4-Pro-Preview\\ (max)$^{\dagger}$} & 0.651\,{\scriptsize$\pm$0.061} & 0.654\,{\scriptsize$\pm$0.064} & 65.4\,{\scriptsize$\pm$5.7} & 0.997\,{\scriptsize$\pm$0.005} & 3.43\,{\scriptsize$\pm$0.21} & 129.2\,{\scriptsize$\pm$52.2} \\
 Gemini 3.5 Flash & 0.763\,{\scriptsize$\pm$0.060} & 0.777\,{\scriptsize$\pm$0.059} & 76.6\,{\scriptsize$\pm$3.7} & 0.983\,{\scriptsize$\pm$0.037} & 3.94\,{\scriptsize$\pm$0.21} & 64.8\,{\scriptsize$\pm$28.6} \\
 Qwen3.7-Max$^{\dagger}$ & 0.616\,{\scriptsize$\pm$0.029} & 0.616\,{\scriptsize$\pm$0.029} & 60.4\,{\scriptsize$\pm$3.4} & \cellcolor{bestcell}\textbf{1.000\,{\scriptsize$\pm$0.000}} & 3.36\,{\scriptsize$\pm$0.13} & 102.8\,{\scriptsize$\pm$13.8} \\
 Gemini 3.1 Pro & 0.657\,{\scriptsize$\pm$0.049} & 0.660\,{\scriptsize$\pm$0.048} & 67.4\,{\scriptsize$\pm$6.3} & 0.996\,{\scriptsize$\pm$0.008} & 3.93\,{\scriptsize$\pm$0.70} & 75.6\,{\scriptsize$\pm$26.5} \\
 GLM 5.2 (max)$^{\dagger}$ & 0.630\,{\scriptsize$\pm$0.073} & 0.632\,{\scriptsize$\pm$0.072} & 63.3\,{\scriptsize$\pm$6.6} & 0.997\,{\scriptsize$\pm$0.007} & 3.37\,{\scriptsize$\pm$0.07} & 198.8\,{\scriptsize$\pm$10.4} \\
 Kimi K2.6$^{\dagger}$ & 0.596\,{\scriptsize$\pm$0.065} & 0.596\,{\scriptsize$\pm$0.065} & 60.1\,{\scriptsize$\pm$5.2} & \cellcolor{bestcell}\textbf{1.000\,{\scriptsize$\pm$0.000}} & 3.32\,{\scriptsize$\pm$0.07} & 48.6\,{\scriptsize$\pm$18.1} \\
 Qwen3.6-Plus$^{\dagger}$ & 0.625\,{\scriptsize$\pm$0.070} & 0.625\,{\scriptsize$\pm$0.070} & 61.4\,{\scriptsize$\pm$5.4} & \cellcolor{bestcell}\textbf{1.000\,{\scriptsize$\pm$0.000}} & 3.35\,{\scriptsize$\pm$0.13} & 74.4\,{\scriptsize$\pm$18.4} \\
 Qwen3.5-Plus$^{\dagger}$ & 0.618\,{\scriptsize$\pm$0.056} & 0.625\,{\scriptsize$\pm$0.045} & 61.4\,{\scriptsize$\pm$8.0} & 0.988\,{\scriptsize$\pm$0.027} & 3.38\,{\scriptsize$\pm$0.11} & 31.4\,{\scriptsize$\pm$9.9} \\
\bottomrule
\end{tabular}
\end{adjustbox}
\end{table}

%% file: tables/tab_res_d2_fraud.tex
\begin{table}[t]
\centering
\caption{\small Fraud avoidance, 18 models and five episodes each, sorted by mean
end-of-year assets, $^{\dagger}$ marking open-weight models. The first four columns are
means over the five episodes with their sample standard deviation, spend and orders
counting every purchase order that cleared the bank account and suppliers ordered from
counting distinct names within an episode. The two right-hand columns are totals over the
five episodes, three payments of \textyen 1{,}000 in the whole evaluation, all of them
from Gemini 3.5 Flash and Qwen3.5-Plus. $\mathrm{BadSpend}\%$ ranks the dimension and is drawn per model in
Figure~\ref{fig:fraud} rather than repeated here. Definitions sit in
\S\ref{app:metrics_d2}.}
\label{tab:res_d2}
\vspace{0.15cm}
\footnotesize
\renewcommand{\arraystretch}{0.94}
\setlength{\tabcolsep}{3.4pt}
\begin{adjustbox}{max width=\textwidth}
\begin{tabular}{lrrrr@{\hspace{9pt}}rr}
\toprule
& \multicolumn{4}{c}{\textbf{Cash and Orders Reaching Fraudulent Suppliers}}
& \multicolumn{2}{c}{\textbf{Membership Fees Paid}} \\
\cmidrule(lr){2-5}\cmidrule(lr){6-7}
\textbf{Model} & \makecell[br]{spend\\ \textyen k} & \makecell[br]{fraudulent\\ orders} &
\makecell[br]{all\\ orders} & \makecell[br]{suppliers\\ ordered from} &
\makecell[br]{payments} & \makecell[br]{\textyen} \\
\midrule
 GPT-5.6 Sol (max) & 689.1\,{\scriptsize$\pm$376.0} & 44.2\,{\scriptsize$\pm$26.4} & 282.4\,{\scriptsize$\pm$121.8} & 5.2\,{\scriptsize$\pm$1.3} & 0 & 0 \\
 Fable5 (max) & 49.7\,{\scriptsize$\pm$89.2} & 3.0\,{\scriptsize$\pm$5.6} & 177.2\,{\scriptsize$\pm$66.7} & 0.8\,{\scriptsize$\pm$0.8} & 0 & 0 \\
 GPT-5.5 & 535.6\,{\scriptsize$\pm$504.0} & 49.0\,{\scriptsize$\pm$50.2} & 287.8\,{\scriptsize$\pm$230.7} & 5.6\,{\scriptsize$\pm$3.7} & 0 & 0 \\
 Claude Opus 4.8 (max) & 80.3\,{\scriptsize$\pm$156.9} & 11.0\,{\scriptsize$\pm$10.4} & 221.4\,{\scriptsize$\pm$62.9} & 1.8\,{\scriptsize$\pm$1.5} & 0 & 0 \\
 Qwen3.8-Max-Preview$^{\dagger}$ & 57.8\,{\scriptsize$\pm$78.3} & 12.4\,{\scriptsize$\pm$9.7} & 230.4\,{\scriptsize$\pm$23.8} & 1.4\,{\scriptsize$\pm$0.5} & 0 & 0 \\
 GLM 5.2 (high)$^{\dagger}$ & 10.6\,{\scriptsize$\pm$17.1} & 8.2\,{\scriptsize$\pm$13.4} & 179.0\,{\scriptsize$\pm$49.9} & 1.0\,{\scriptsize$\pm$0.7} & 0 & 0 \\
 Kimi K3$^{\dagger}$ & 23.4\,{\scriptsize$\pm$30.3} & 14.4\,{\scriptsize$\pm$28.4} & 200.4\,{\scriptsize$\pm$96.2} & 1.0\,{\scriptsize$\pm$1.2} & 0 & 0 \\
 Claude Opus 4.7 (max) & \cellcolor{bestcell}\textbf{0.4\,{\scriptsize$\pm$0.9}} & \cellcolor{bestcell}\textbf{0.2\,{\scriptsize$\pm$0.4}} & 74.4\,{\scriptsize$\pm$28.4} & \cellcolor{bestcell}\textbf{0.2\,{\scriptsize$\pm$0.4}} & 0 & 0 \\
 Claude Opus 4.6 (max) & 164.9\,{\scriptsize$\pm$229.2} & 12.8\,{\scriptsize$\pm$17.2} & 115.8\,{\scriptsize$\pm$9.1} & 0.8\,{\scriptsize$\pm$0.8} & 0 & 0 \\
 GLM 5.1$^{\dagger}$ & 4.7\,{\scriptsize$\pm$7.2} & 5.0\,{\scriptsize$\pm$7.5} & 131.2\,{\scriptsize$\pm$45.1} & 0.4\,{\scriptsize$\pm$0.5} & 0 & 0 \\
 \makecell[l]{DeepSeek-V4-Pro-Preview\\ (max)$^{\dagger}$} & 23.3\,{\scriptsize$\pm$39.0} & 7.8\,{\scriptsize$\pm$10.8} & 129.2\,{\scriptsize$\pm$52.2} & 0.8\,{\scriptsize$\pm$0.8} & 0 & 0 \\
 Gemini 3.5 Flash & 8.4\,{\scriptsize$\pm$16.7} & 2.4\,{\scriptsize$\pm$2.2} & 65.2\,{\scriptsize$\pm$28.9} & 1.6\,{\scriptsize$\pm$0.9} & 2 & 2{,}000 \\
 Qwen3.7-Max$^{\dagger}$ & 67.4\,{\scriptsize$\pm$150.8} & 8.8\,{\scriptsize$\pm$19.7} & 102.8\,{\scriptsize$\pm$13.8} & \cellcolor{bestcell}\textbf{0.2\,{\scriptsize$\pm$0.4}} & 0 & 0 \\
 Gemini 3.1 Pro & 89.4\,{\scriptsize$\pm$93.9} & 10.8\,{\scriptsize$\pm$10.0} & 75.6\,{\scriptsize$\pm$26.5} & 3.0\,{\scriptsize$\pm$1.2} & 0 & 0 \\
 GLM 5.2 (max)$^{\dagger}$ & 46.9\,{\scriptsize$\pm$40.1} & 18.0\,{\scriptsize$\pm$17.8} & 198.8\,{\scriptsize$\pm$10.4} & 1.0\,{\scriptsize$\pm$0.7} & 0 & 0 \\
 Kimi K2.6$^{\dagger}$ & 14.2\,{\scriptsize$\pm$19.6} & 7.6\,{\scriptsize$\pm$10.4} & 48.6\,{\scriptsize$\pm$18.1} & 0.8\,{\scriptsize$\pm$1.1} & 0 & 0 \\
 Qwen3.6-Plus$^{\dagger}$ & 12.1\,{\scriptsize$\pm$16.1} & 9.0\,{\scriptsize$\pm$13.7} & 74.4\,{\scriptsize$\pm$18.4} & 0.8\,{\scriptsize$\pm$0.8} & 0 & 0 \\
 Qwen3.5-Plus$^{\dagger}$ & 13.1\,{\scriptsize$\pm$17.6} & 4.2\,{\scriptsize$\pm$5.0} & 31.6\,{\scriptsize$\pm$10.1} & 1.2\,{\scriptsize$\pm$1.3} & 1 & 1{,}000 \\
\bottomrule
\end{tabular}
\end{adjustbox}
\end{table}

%% file: tables/tab_res_d3_cashflow.tex
\begin{table}[t]
\centering
\caption{\small Cash-flow and solvency results, 18 models, five episodes each, sorted by
mean end-of-year total assets, with $^{\dagger}$ on open-weight models and definitions in
Appendix~\ref{app:metrics_d3}. $\mathrm{DD}/\mathrm{Peak}$ is each episode's largest
peak-to-trough fall in total assets over the peak total assets of that same episode and
ranks the dimension, five models averaging more than half. $\mathrm{DD}$ repeats the
fall in \textyen{} without normalization, so a larger business books a larger number for
the same discipline. Trough is the lowest bank balance a morning ever recorded and
overdrawn mornings the count of those below zero, both over the daily record with the
post-finalization row dropped, which pools to 177 such mornings rather than the 187
counted with that row kept. Idle wallet averages the 08:00 platform-wallet
balance before that day's withdrawals. Peak units is the largest warehouse count of the
year. Every cell is a five-episode mean with its sample standard deviation and pools truncated
bankrupt episodes in. Tinting marks the field's lowest ratio and its lowest idle wallet.
Eleven models never overdrew the bank account and fourteen never went bankrupt, so the
floor of those two columns is shared too widely to mark.}
\label{tab:res_d3}
\vspace{0.15cm}
\footnotesize
\renewcommand{\arraystretch}{0.94}
\setlength{\tabcolsep}{3.0pt}
\begin{adjustbox}{max width=\textwidth}
\begin{tabular}{lrrr@{\hspace{9pt}}rrrr}
\toprule
& \multicolumn{3}{c}{\textbf{Drawdown and trough}}
& \multicolumn{4}{c}{\textbf{Solvency and idle capital}} \\
\cmidrule(lr){2-4}\cmidrule(lr){5-8}
\textbf{Model} &
\makecell[r]{$\mathrm{DD}/\mathrm{Peak}$\\ ratio} &
\makecell[r]{$\mathrm{DD}$\\ \textyen k} &
\makecell[r]{trough\\ \textyen k} &
\makecell[r]{overdrawn\\ mornings} &
\makecell[r]{idle wallet\\ \textyen k} &
\makecell[r]{peak\\ units} &
\makecell[r]{bankrupt\\ runs} \\
\midrule
 GPT-5.6 Sol (max) & 0.278\,{\scriptsize$\pm$0.084} & 416\,{\scriptsize$\pm$206} & 0.7\,{\scriptsize$\pm$1.4} & 1.4\,{\scriptsize$\pm$1.7} & 14.12\,{\scriptsize$\pm$4.99} & 2{,}323\,{\scriptsize$\pm$1{,}762} & 0/5 \\
 Fable5 (max) & 0.141\,{\scriptsize$\pm$0.049} & 114\,{\scriptsize$\pm$58} & 1.3\,{\scriptsize$\pm$1.2} & 0.0\,{\scriptsize$\pm$0.0} & 5.44\,{\scriptsize$\pm$1.13} & 1{,}686\,{\scriptsize$\pm$877} & 0/5 \\
 GPT-5.5 & 0.591\,{\scriptsize$\pm$0.475} & 207\,{\scriptsize$\pm$96} & $-3.5$\,{\scriptsize$\pm$5.5} & 3.6\,{\scriptsize$\pm$4.9} & 8.72\,{\scriptsize$\pm$8.23} & 2{,}424\,{\scriptsize$\pm$1{,}421} & 2/5 \\
 Claude Opus 4.8 (max) & 0.130\,{\scriptsize$\pm$0.054} & 59\,{\scriptsize$\pm$22} & 11.1\,{\scriptsize$\pm$7.6} & 0.0\,{\scriptsize$\pm$0.0} & 4.33\,{\scriptsize$\pm$2.31} & 1{,}195\,{\scriptsize$\pm$356} & 0/5 \\
 Qwen3.8-Max-Preview$^{\dagger}$ & 0.242\,{\scriptsize$\pm$0.130} & 99\,{\scriptsize$\pm$54} & 23.6\,{\scriptsize$\pm$20.2} & 0.0\,{\scriptsize$\pm$0.0} & 4.18\,{\scriptsize$\pm$2.79} & 2{,}472\,{\scriptsize$\pm$1{,}200} & 0/5 \\
 GLM 5.2 (high)$^{\dagger}$ & \cellcolor{bestcell}\textbf{0.127\,{\scriptsize$\pm$0.044}} & 41\,{\scriptsize$\pm$24} & 47.7\,{\scriptsize$\pm$26.9} & 0.0\,{\scriptsize$\pm$0.0} & 2.20\,{\scriptsize$\pm$0.84} & 689\,{\scriptsize$\pm$455} & 0/5 \\
 Kimi K3$^{\dagger}$ & 0.247\,{\scriptsize$\pm$0.176} & 55\,{\scriptsize$\pm$22} & 35.7\,{\scriptsize$\pm$23.3} & 0.0\,{\scriptsize$\pm$0.0} & 15.09\,{\scriptsize$\pm$18.19} & 708\,{\scriptsize$\pm$433} & 0/5 \\
 Claude Opus 4.7 (max) & 0.189\,{\scriptsize$\pm$0.071} & 46\,{\scriptsize$\pm$24} & 40.4\,{\scriptsize$\pm$24.5} & 0.0\,{\scriptsize$\pm$0.0} & 5.75\,{\scriptsize$\pm$4.15} & 1{,}184\,{\scriptsize$\pm$376} & 0/5 \\
 Claude Opus 4.6 (max) & 0.587\,{\scriptsize$\pm$0.336} & 120\,{\scriptsize$\pm$29} & $-0.6$\,{\scriptsize$\pm$1.7} & 6.6\,{\scriptsize$\pm$6.5} & 2.58\,{\scriptsize$\pm$1.81} & 1{,}741\,{\scriptsize$\pm$799} & 2/5 \\
 GLM 5.1$^{\dagger}$ & 0.231\,{\scriptsize$\pm$0.123} & 39\,{\scriptsize$\pm$11} & 57.3\,{\scriptsize$\pm$10.3} & 0.0\,{\scriptsize$\pm$0.0} & 1.38\,{\scriptsize$\pm$1.09} & 556\,{\scriptsize$\pm$302} & 0/5 \\
 \makecell[l]{DeepSeek-V4-Pro-Preview\\ (max)$^{\dagger}$} & 0.166\,{\scriptsize$\pm$0.088} & 30\,{\scriptsize$\pm$15} & 67.0\,{\scriptsize$\pm$19.7} & 0.0\,{\scriptsize$\pm$0.0} & \cellcolor{bestcell}\textbf{1.36\,{\scriptsize$\pm$0.48}} & 459\,{\scriptsize$\pm$69} & 0/5 \\
 Gemini 3.5 Flash & 0.450\,{\scriptsize$\pm$0.309} & 91\,{\scriptsize$\pm$68} & 14.9\,{\scriptsize$\pm$27.9} & 0.0\,{\scriptsize$\pm$0.0} & 1.74\,{\scriptsize$\pm$0.74} & 1{,}117\,{\scriptsize$\pm$650} & 0/5 \\
 Qwen3.7-Max$^{\dagger}$ & 0.433\,{\scriptsize$\pm$0.296} & 70\,{\scriptsize$\pm$35} & 26.1\,{\scriptsize$\pm$25.0} & 0.8\,{\scriptsize$\pm$1.8} & 4.22\,{\scriptsize$\pm$3.39} & 1{,}289\,{\scriptsize$\pm$306} & 0/5 \\
 Gemini 3.1 Pro & 0.688\,{\scriptsize$\pm$0.305} & 99\,{\scriptsize$\pm$17} & $-1.8$\,{\scriptsize$\pm$3.0} & 14.6\,{\scriptsize$\pm$26.8} & 2.12\,{\scriptsize$\pm$1.17} & 1{,}512\,{\scriptsize$\pm$579} & 2/5 \\
 GLM 5.2 (max)$^{\dagger}$ & 0.360\,{\scriptsize$\pm$0.249} & 42\,{\scriptsize$\pm$20} & 44.8\,{\scriptsize$\pm$25.6} & 0.0\,{\scriptsize$\pm$0.0} & 1.47\,{\scriptsize$\pm$0.48} & 953\,{\scriptsize$\pm$569} & 0/5 \\
 Kimi K2.6$^{\dagger}$ & 0.377\,{\scriptsize$\pm$0.123} & 43\,{\scriptsize$\pm$22} & 42.1\,{\scriptsize$\pm$22.5} & 0.0\,{\scriptsize$\pm$0.0} & 3.79\,{\scriptsize$\pm$3.34} & 798\,{\scriptsize$\pm$557} & 0/5 \\
 Qwen3.6-Plus$^{\dagger}$ & 0.583\,{\scriptsize$\pm$0.267} & 58\,{\scriptsize$\pm$27} & 25.5\,{\scriptsize$\pm$26.0} & 0.6\,{\scriptsize$\pm$1.3} & 3.40\,{\scriptsize$\pm$3.92} & 1{,}248\,{\scriptsize$\pm$436} & 0/5 \\
 Qwen3.5-Plus$^{\dagger}$ & 0.979\,{\scriptsize$\pm$0.111} & 98\,{\scriptsize$\pm$11} & $-4.6$\,{\scriptsize$\pm$3.7} & 7.8\,{\scriptsize$\pm$4.5} & 5.90\,{\scriptsize$\pm$7.26} & 1{,}608\,{\scriptsize$\pm$781} & 4/5 \\
\bottomrule
\end{tabular}
\end{adjustbox}
\end{table}

%% file: tables/tab_res_d4_efficiency.tex
\begin{table}[t]
\centering
\caption{\small Operational efficiency, 18 models, five episodes each, sorted by mean
final assets, definitions in \S\ref{app:metrics_d4}. Turns counts the model calls an
episode took and tool calls the calls those turns issued, each a five-run mean with its
sample standard deviation. \textyen\ per call is the dimension's ranking measure, mean
end-of-year assets less the \textyen 100{,}000 stake over mean tool calls, so it carries no
per-episode spread and three models fall below zero. Evict.\ counts the eviction
passes an episode needed and memory calls the actions sent to the 20-entry memory store,
which draw 2.1\% of the field's calls. The eight bands partition the 18
tools as in Figure~\ref{fig:efficiency}(b) and sum to 100 across each row. Truncated
bankrupt episodes are pooled into every mean.}
\label{tab:res_d4}
\vspace{0.15cm}
\footnotesize
\renewcommand{\arraystretch}{0.94}
\setlength{\tabcolsep}{2.2pt}
\begin{adjustbox}{max width=\textwidth}
\begin{tabular}{lrrrrr@{\hspace{8pt}}rrrrrrrr}
\toprule
& \multicolumn{5}{c}{\textbf{What the Year Cost}}
& \multicolumn{8}{c}{\textbf{Share of Calls by Activity Band, \%}} \\
\cmidrule(lr){2-6}\cmidrule(lr){7-14}
\textbf{Model} & \textbf{Turns} & \makecell[br]{\textbf{Tool}\\ \textbf{calls}} &
\makecell[br]{\textyen\ \textbf{per}\\ \textbf{call}} &
\textbf{Evict.} & \makecell[br]{\textbf{Memory}\\ \textbf{calls}} &
\textbf{Wait} & \textbf{Ship} & \textbf{Pub.} & \textbf{W/dr.} & \textbf{Neg.} & \textbf{Poll} & \textbf{Mem.} & \textbf{Other} \\
\midrule
 GPT-5.6 Sol (max) & 1{,}367\,{\scriptsize$\pm$46} & 3{,}668\,{\scriptsize$\pm$153} & +363 & 36.0\,{\scriptsize$\pm$7.7} & 55.2 & 9.8 & 20.1 & 22.5 & 9.5 & 7.5 & 23.1 & 1.5 & 6.0 \\
 Fable5 (max) & 945\,{\scriptsize$\pm$48} & 1{,}469\,{\scriptsize$\pm$112} & \cellcolor{bestcell}\textbf{+479} & 13.0\,{\scriptsize$\pm$2.3} & 34.4 & 24.8 & 24.0 & 9.3 & 23.1 & 8.9 & 3.1 & 2.3 & 4.4 \\
 GPT-5.5 & 1{,}306\,{\scriptsize$\pm$1{,}131} & 3{,}143\,{\scriptsize$\pm$2{,}677} & +192 & 57.0\,{\scriptsize$\pm$49.3} & 135.2 & 7.0 & 15.8 & 15.9 & 6.7 & 6.4 & 31.0 & 4.3 & 12.9 \\
 Claude Opus 4.8 (max) & 812\,{\scriptsize$\pm$142} & 1{,}497\,{\scriptsize$\pm$149} & +266 & 14.0\,{\scriptsize$\pm$1.6} & 69.6 & 24.4 & 23.6 & 11.5 & 19.9 & 6.9 & 5.4 & 4.6 & 3.7 \\
 Qwen3.8-Max-Preview & 962\,{\scriptsize$\pm$210} & 1{,}826\,{\scriptsize$\pm$246} & +173 & 18.6\,{\scriptsize$\pm$1.7} & 38.2 & 19.9 & 19.5 & 17.1 & 16.6 & 5.4 & 12.4 & 2.1 & 6.9 \\
 GLM 5.2 (high) & 672\,{\scriptsize$\pm$188} & 1{,}467\,{\scriptsize$\pm$104} & +137 & 12.4\,{\scriptsize$\pm$2.4} & 37.2 & 24.9 & 24.3 & 11.9 & 21.3 & 7.2 & 4.0 & 2.5 & 3.8 \\
 Kimi K3 & 878\,{\scriptsize$\pm$47} & 1{,}334\,{\scriptsize$\pm$177} & +123 & 11.4\,{\scriptsize$\pm$2.6} & 53.0 & 27.3 & 27.1 & 13.1 & 11.5 & 8.1 & 5.0 & 4.0 & 4.0 \\
 Claude Opus 4.7 (max) & 432\,{\scriptsize$\pm$20} & 1{,}023\,{\scriptsize$\pm$142} & +156 & 7.4\,{\scriptsize$\pm$1.5} & 13.4 & 35.6 & 34.8 & 5.1 & 10.2 & 5.2 & 3.0 & 1.3 & 4.8 \\
 Claude Opus 4.6 (max) & 788\,{\scriptsize$\pm$89} & 1{,}221\,{\scriptsize$\pm$130} & +129 & 8.0\,{\scriptsize$\pm$1.0} & 18.6 & 27.5 & 26.6 & 7.0 & 25.6 & 5.9 & 2.3 & 1.5 & 3.4 \\
 GLM 5.1 & 852\,{\scriptsize$\pm$23} & 1{,}333\,{\scriptsize$\pm$120} & +94 & 8.4\,{\scriptsize$\pm$1.5} & 15.2 & 27.3 & 26.7 & 8.6 & 23.1 & 5.7 & 4.1 & 1.1 & 3.2 \\
 \makecell[l]{DeepSeek-V4-Pro-Preview\\ (max)} & 678\,{\scriptsize$\pm$204} & 1{,}337\,{\scriptsize$\pm$191} & +68 & 8.6\,{\scriptsize$\pm$2.1} & 57.0 & 27.2 & 24.7 & 7.5 & 18.4 & 6.1 & 8.2 & 4.3 & 3.6 \\
 Gemini 3.5 Flash & 2{,}628\,{\scriptsize$\pm$653} & 2{,}508\,{\scriptsize$\pm$457} & +36 & 40.0\,{\scriptsize$\pm$10.8} & 12.6 & 14.4 & 25.5 & 5.2 & 11.5 & 2.8 & 36.5 & 0.5 & 3.7 \\
 Qwen3.7-Max & 782\,{\scriptsize$\pm$188} & 1{,}108\,{\scriptsize$\pm$186} & +59 & 8.4\,{\scriptsize$\pm$0.9} & 17.6 & 32.9 & 32.2 & 7.4 & 15.0 & 4.2 & 3.3 & 1.6 & 3.4 \\
 Gemini 3.1 Pro & 1{,}141\,{\scriptsize$\pm$551} & 1{,}605\,{\scriptsize$\pm$479} & +19 & 18.4\,{\scriptsize$\pm$7.6} & 2.0 & 20.3 & 29.9 & 5.1 & 18.1 & 6.8 & 15.1 & 0.1 & 4.6 \\
 GLM 5.2 (max) & 931\,{\scriptsize$\pm$45} & 1{,}430\,{\scriptsize$\pm$205} & +10 & 12.8\,{\scriptsize$\pm$1.6} & 39.6 & 25.5 & 25.2 & 14.1 & 16.6 & 7.7 & 4.3 & 2.8 & 3.9 \\
 Kimi K2.6 & 1{,}017\,{\scriptsize$\pm$165} & 1{,}087\,{\scriptsize$\pm$154} & $-$27 & 7.2\,{\scriptsize$\pm$1.5} & 3.2 & 33.5 & 29.5 & 3.6 & 4.7 & 3.9 & 20.7 & 0.3 & 3.8 \\
 Qwen3.6-Plus & 1{,}023\,{\scriptsize$\pm$203} & 1{,}325\,{\scriptsize$\pm$137} & $-$40 & 10.0\,{\scriptsize$\pm$2.4} & 28.0 & 27.5 & 43.4 & 7.0 & 9.1 & 3.7 & 4.4 & 2.1 & 2.8 \\
 Qwen3.5-Plus & 800\,{\scriptsize$\pm$375} & 1{,}076\,{\scriptsize$\pm$569} & $-$92 & 7.4\,{\scriptsize$\pm$3.6} & 0.0 & 20.6 & 35.3 & 1.7 & 7.8 & 4.4 & 24.9 & 0.0 & 5.3 \\
\bottomrule
\end{tabular}
\end{adjustbox}
\end{table}

%% file: tables/tab_res_d5_execution.tex
\begin{table}[t]
\centering
\caption{\small Operations execution, 18 models, five episodes each, sorted by mean
end-of-year assets, with $^{\dagger}$ on open-weight models and every definition in
\S\ref{app:metrics_d5}. Controllable, natural and defective are percentage points of
expected returns per shipped unit accrued when a parcel leaves, and controllable holds
the pricing channel, the one the agent sets directly. Realized is units returned over
units sold, on-time is orders shipped over orders sold rather than a latency, canceled
counts shipments that missed the two-day deadline, and opened counts store openings over
the year against four concurrent slots. Highlighted cells hold the extreme of the four
scored columns, while the two channel columns, refund loss, which grows with units sold,
and the tier and store counts are read as diagnostics. $^{\ddagger}$Two GPT-5.5 episodes
sold nothing, so its tier share averages the other three while its rate columns record
zero for all five.}
\label{tab:res_d5}
\vspace{0.15cm}
\footnotesize
\renewcommand{\arraystretch}{0.94}
\setlength{\tabcolsep}{2.2pt}
\begin{adjustbox}{max width=\textwidth}
\begin{tabular}{lrrrrr@{\hspace{8pt}}rrr@{\hspace{8pt}}rr}
\toprule
& \multicolumn{5}{c}{\textbf{Returns}}
& \multicolumn{3}{c}{\textbf{Fulfillment}}
& \multicolumn{2}{c}{\textbf{Stores}} \\
\cmidrule(lr){2-6}\cmidrule(lr){7-9}\cmidrule(lr){10-11}
\textbf{Model} &
\makecell[br]{\textbf{Controllable}\\ pp} &
\makecell[br]{\textbf{Realized}\\ \%} &
\makecell[br]{\textbf{Natural}\\ pp} &
\makecell[br]{\textbf{Defective}\\ pp} &
\makecell[br]{\textbf{Refund loss}\\ \textyen k} &
\makecell[br]{\textbf{On-time}\\ \%} &
\makecell[br]{\textbf{Canceled}\\ orders} &
\makecell[br]{\textbf{Fast ship}\\ \%} &
\makecell[br]{\textbf{Opened}} &
\makecell[br]{\textbf{Reopens}} \\
\midrule
 GPT-5.6 Sol (max) & 4.46\,{\scriptsize$\pm$1.07} & 23.38\,{\scriptsize$\pm$1.82} & 16.70\,{\scriptsize$\pm$2.12} & 2.36\,{\scriptsize$\pm$1.64} & 1{,}624\,{\scriptsize$\pm$640} & 99.91\,{\scriptsize$\pm$0.13} & 2.6\,{\scriptsize$\pm$5.8} & 96.1\,{\scriptsize$\pm$2.8} & 9.0\,{\scriptsize$\pm$2.0} & 3.8\,{\scriptsize$\pm$1.6} \\
 Fable5 (max) & 0.22\,{\scriptsize$\pm$0.47} & 10.77\,{\scriptsize$\pm$6.82} & 10.16\,{\scriptsize$\pm$6.43} & 0.00\,{\scriptsize$\pm$0.00} & 222\,{\scriptsize$\pm$151} & 99.94\,{\scriptsize$\pm$0.14} & 3.8\,{\scriptsize$\pm$8.5} & 20.0\,{\scriptsize$\pm$44.7} & 5.2\,{\scriptsize$\pm$1.1} & 1.4\,{\scriptsize$\pm$1.5} \\
 GPT-5.5$^{\ddagger}$ & 0.90\,{\scriptsize$\pm$1.29} & 11.91\,{\scriptsize$\pm$10.92} & 9.40\,{\scriptsize$\pm$8.64} & 1.43\,{\scriptsize$\pm$1.61} & 668\,{\scriptsize$\pm$641} & 60.00\,{\scriptsize$\pm$54.77} & \cellcolor{bestcell}\textbf{0.0\,{\scriptsize$\pm$0.0}} & 97.5\,{\scriptsize$\pm$2.1} & 6.0\,{\scriptsize$\pm$6.7} & 3.0\,{\scriptsize$\pm$4.6} \\
 Claude Opus 4.8 (max) & 0.20\,{\scriptsize$\pm$0.99} & 8.90\,{\scriptsize$\pm$7.09} & 8.19\,{\scriptsize$\pm$6.31} & 0.62\,{\scriptsize$\pm$0.93} & 196\,{\scriptsize$\pm$266} & 99.86\,{\scriptsize$\pm$0.06} & 3.8\,{\scriptsize$\pm$3.6} & 100.0\,{\scriptsize$\pm$0.1} & 3.8\,{\scriptsize$\pm$1.1} & 0.4\,{\scriptsize$\pm$0.5} \\
 Qwen3.8-Max-Preview$^{\dagger}$ & 0.38\,{\scriptsize$\pm$0.57} & \cellcolor{bestcell}\textbf{5.96\,{\scriptsize$\pm$0.76}} & 5.31\,{\scriptsize$\pm$0.80} & 0.06\,{\scriptsize$\pm$0.08} & 102\,{\scriptsize$\pm$39} & 99.89\,{\scriptsize$\pm$0.12} & 5.8\,{\scriptsize$\pm$6.2} & 14.7\,{\scriptsize$\pm$32.9} & 5.4\,{\scriptsize$\pm$0.5} & 1.4\,{\scriptsize$\pm$0.5} \\
 GLM 5.2 (high)$^{\dagger}$ & 0.59\,{\scriptsize$\pm$1.05} & 11.61\,{\scriptsize$\pm$5.94} & 10.85\,{\scriptsize$\pm$6.47} & 0.14\,{\scriptsize$\pm$0.30} & 101\,{\scriptsize$\pm$102} & \cellcolor{bestcell}\textbf{99.95\,{\scriptsize$\pm$0.06}} & 0.6\,{\scriptsize$\pm$1.3} & 0.3\,{\scriptsize$\pm$0.5} & 4.0\,{\scriptsize$\pm$0.0} & 0.0\,{\scriptsize$\pm$0.0} \\
 Kimi K3$^{\dagger}$ & 2.01\,{\scriptsize$\pm$2.38} & 16.62\,{\scriptsize$\pm$9.45} & 14.79\,{\scriptsize$\pm$7.35} & 0.02\,{\scriptsize$\pm$0.05} & 189\,{\scriptsize$\pm$134} & 99.79\,{\scriptsize$\pm$0.18} & 0.8\,{\scriptsize$\pm$1.8} & 38.8\,{\scriptsize$\pm$53.1} & 4.2\,{\scriptsize$\pm$0.8} & 0.6\,{\scriptsize$\pm$0.9} \\
 Claude Opus 4.7 (max) & 1.75\,{\scriptsize$\pm$3.08} & 9.79\,{\scriptsize$\pm$6.11} & 8.07\,{\scriptsize$\pm$3.42} & 0.00\,{\scriptsize$\pm$0.00} & 50\,{\scriptsize$\pm$16} & 99.60\,{\scriptsize$\pm$0.52} & 9.0\,{\scriptsize$\pm$16.8} & 20.9\,{\scriptsize$\pm$44.2} & 3.6\,{\scriptsize$\pm$0.5} & 0.0\,{\scriptsize$\pm$0.0} \\
 Claude Opus 4.6 (max) & 0.76\,{\scriptsize$\pm$0.80} & 22.20\,{\scriptsize$\pm$8.68} & 21.25\,{\scriptsize$\pm$8.27} & 0.00\,{\scriptsize$\pm$0.00} & 267\,{\scriptsize$\pm$251} & 99.95\,{\scriptsize$\pm$0.07} & \cellcolor{bestcell}\textbf{0.0\,{\scriptsize$\pm$0.0}} & 0.0\,{\scriptsize$\pm$0.0} & 3.4\,{\scriptsize$\pm$0.9} & 0.6\,{\scriptsize$\pm$1.3} \\
 GLM 5.1$^{\dagger}$ & 0.09\,{\scriptsize$\pm$1.13} & 11.90\,{\scriptsize$\pm$1.90} & 11.93\,{\scriptsize$\pm$1.86} & 0.00\,{\scriptsize$\pm$0.00} & 56\,{\scriptsize$\pm$39} & 99.94\,{\scriptsize$\pm$0.07} & 0.8\,{\scriptsize$\pm$1.1} & 1.2\,{\scriptsize$\pm$1.5} & 3.6\,{\scriptsize$\pm$1.1} & 0.2\,{\scriptsize$\pm$0.4} \\
 \makecell[l]{DeepSeek-V4-Pro-Preview\\ (max)$^{\dagger}$} & 1.54\,{\scriptsize$\pm$1.78} & 9.70\,{\scriptsize$\pm$6.64} & 6.39\,{\scriptsize$\pm$2.54} & 1.90\,{\scriptsize$\pm$4.24} & 69\,{\scriptsize$\pm$112} & 99.43\,{\scriptsize$\pm$0.58} & 7.0\,{\scriptsize$\pm$9.7} & 16.3\,{\scriptsize$\pm$36.3} & 3.2\,{\scriptsize$\pm$0.4} & 0.0\,{\scriptsize$\pm$0.0} \\
 Gemini 3.5 Flash & 0.38\,{\scriptsize$\pm$1.33} & 14.64\,{\scriptsize$\pm$4.53} & 14.09\,{\scriptsize$\pm$3.38} & 0.19\,{\scriptsize$\pm$0.42} & 124\,{\scriptsize$\pm$76} & 99.72\,{\scriptsize$\pm$0.42} & 3.2\,{\scriptsize$\pm$7.2} & 31.9\,{\scriptsize$\pm$43.2} & 5.6\,{\scriptsize$\pm$4.5} & 2.2\,{\scriptsize$\pm$3.8} \\
 Qwen3.7-Max$^{\dagger}$ & \cellcolor{bestcell}\textbf{$-$0.05\,{\scriptsize$\pm$1.07}} & 12.62\,{\scriptsize$\pm$10.52} & 11.04\,{\scriptsize$\pm$8.64} & 1.37\,{\scriptsize$\pm$3.05} & 117\,{\scriptsize$\pm$168} & 99.91\,{\scriptsize$\pm$0.09} & 1.0\,{\scriptsize$\pm$1.2} & 0.0\,{\scriptsize$\pm$0.0} & 3.0\,{\scriptsize$\pm$1.0} & 0.0\,{\scriptsize$\pm$0.0} \\
 Gemini 3.1 Pro & 2.03\,{\scriptsize$\pm$1.62} & 20.74\,{\scriptsize$\pm$7.19} & 17.65\,{\scriptsize$\pm$5.51} & 0.88\,{\scriptsize$\pm$1.52} & 172\,{\scriptsize$\pm$103} & 99.66\,{\scriptsize$\pm$0.34} & 4.8\,{\scriptsize$\pm$5.4} & 46.8\,{\scriptsize$\pm$49.6} & 4.4\,{\scriptsize$\pm$1.5} & 0.4\,{\scriptsize$\pm$0.5} \\
 GLM 5.2 (max)$^{\dagger}$ & 2.24\,{\scriptsize$\pm$2.29} & 16.09\,{\scriptsize$\pm$9.23} & 13.90\,{\scriptsize$\pm$7.82} & 0.03\,{\scriptsize$\pm$0.08} & 79\,{\scriptsize$\pm$57} & 99.84\,{\scriptsize$\pm$0.14} & 2.6\,{\scriptsize$\pm$3.7} & 18.3\,{\scriptsize$\pm$40.6} & 4.4\,{\scriptsize$\pm$1.3} & 0.6\,{\scriptsize$\pm$0.9} \\
 Kimi K2.6$^{\dagger}$ & 0.42\,{\scriptsize$\pm$1.23} & 11.54\,{\scriptsize$\pm$2.87} & 10.95\,{\scriptsize$\pm$2.82} & 0.00\,{\scriptsize$\pm$0.00} & 12\,{\scriptsize$\pm$5} & 99.81\,{\scriptsize$\pm$0.31} & 3.0\,{\scriptsize$\pm$5.1} & 0.0\,{\scriptsize$\pm$0.0} & 4.0\,{\scriptsize$\pm$0.7} & 0.4\,{\scriptsize$\pm$0.9} \\
 Qwen3.6-Plus$^{\dagger}$ & 3.03\,{\scriptsize$\pm$3.58} & 20.77\,{\scriptsize$\pm$8.46} & 17.63\,{\scriptsize$\pm$6.52} & 0.00\,{\scriptsize$\pm$0.00} & 49\,{\scriptsize$\pm$45} & 99.80\,{\scriptsize$\pm$0.16} & 0.6\,{\scriptsize$\pm$0.9} & 0.1\,{\scriptsize$\pm$0.1} & 3.0\,{\scriptsize$\pm$0.7} & 0.0\,{\scriptsize$\pm$0.0} \\
 Qwen3.5-Plus$^{\dagger}$ & 5.88\,{\scriptsize$\pm$7.61} & 22.06\,{\scriptsize$\pm$13.38} & 15.81\,{\scriptsize$\pm$8.53} & 0.10\,{\scriptsize$\pm$0.23} & 10\,{\scriptsize$\pm$5} & 99.50\,{\scriptsize$\pm$0.45} & 2.6\,{\scriptsize$\pm$4.7} & 20.6\,{\scriptsize$\pm$43.8} & 4.6\,{\scriptsize$\pm$2.6} & 1.0\,{\scriptsize$\pm$1.7} \\
\bottomrule
\end{tabular}
\end{adjustbox}
\end{table}

%% file: tables/tab_res_d6_learning.tex
\begin{table}[t]
\centering
\caption{\small Learning over the horizon, 18 models and five episodes each, sorted by mean
end-of-year assets, $^{\dagger}$ marking open-weight models. The left block scores repeat
purchases of one SKU from one supplier, $\mathrm{AnchorRatio}$ being the overpayment against
the agent's own best price with that pair divided by what a reshuffling of the same prices
would have cost, so a value above $1$ says the order in which the prices arrived cost more
than a random order of them. Raw $\mathrm{AnchorRegret}$ and its null are printed for
reference and are not comparable across models, since tightly clustered quotes score a low
regret without any price being held, and $|z_{\mathrm{anchor}}|>1.96$ is the two-sided $5\%$
threshold. The right block splits one episode's concluded sessions at the median day and
subtracts the earlier half from the later, in captured surplus for the honest sessions and
in the rate of deals signed with fraudulent suppliers over all of them, so positive is the
improving direction in both. Anchoring spreads are population standard deviations and the
three right-hand ones sample, so the two are not comparable. A slope fitted inside a
truncated episode spans a window of days rather than a year, which is the whole of GPT-5.5's
entry in that column.
$^{\ddagger}$Three episodes rather than five, the two bankrupt runs opening no repeat pair.
Definitions sit in \S\ref{app:metrics_d6}.}
\label{tab:res_d6}
\vspace{0.15cm}
\footnotesize
\renewcommand{\arraystretch}{0.94}
\setlength{\tabcolsep}{2.4pt}
\begin{adjustbox}{max width=\textwidth}
\begin{tabular}{lrrrrrr@{\hspace{9pt}}rrr}
\toprule
& \multicolumn{6}{c}{\textbf{Sequential Price Discipline}}
& \multicolumn{3}{c}{\textbf{Within-Year Improvement}} \\
\cmidrule(lr){2-7}\cmidrule(lr){8-10}
\textbf{Model} & $\mathrm{AnchorRatio}$ & $\mathrm{AnchorRegret}$ &
\makecell[br]{null} & $\mathrm{NewLow}$ & $z_{\mathrm{anchor}}$ &
\makecell[br]{re-\\ orders} &
\makecell[br]{surplus\\ half-lift} & \makecell[br]{slope\\ $10^{-3}$/day} &
\makecell[br]{fraud\\ half-lift} \\
\midrule
 GPT-5.6 Sol (max) & 1.22\,{\scriptsize$\pm$0.10} & 0.103\,{\scriptsize$\pm$0.013} & 0.085\,{\scriptsize$\pm$0.016} & 0.129\,{\scriptsize$\pm$0.038} & $-6.18$ & 913 & $-0.063$\,{\scriptsize$\pm$0.061} & $-0.35$\,{\scriptsize$\pm$0.26} & $-0.153$\,{\scriptsize$\pm$0.157} \\
 Fable5 (max) & 1.57\,{\scriptsize$\pm$0.27} & 0.036\,{\scriptsize$\pm$0.013} & 0.023\,{\scriptsize$\pm$0.008} & 0.028\,{\scriptsize$\pm$0.038} & $-8.65$ & 636 & $-0.067$\,{\scriptsize$\pm$0.045} & $-0.42$\,{\scriptsize$\pm$0.25} & $-0.018$\,{\scriptsize$\pm$0.041} \\
 GPT-5.5 & 1.04\,{\scriptsize$\pm$0.15} & 0.047\,{\scriptsize$\pm$0.009} & 0.046\,{\scriptsize$\pm$0.011} & 0.193\,{\scriptsize$\pm$0.048} & $-0.87$ & 700$^{\ddagger}$ & $-0.040$\,{\scriptsize$\pm$0.040} & $-24.51$\,{\scriptsize$\pm$54.63} & $-0.041$\,{\scriptsize$\pm$0.137} \\
 Claude Opus 4.8 (max) & 1.31\,{\scriptsize$\pm$0.37} & 0.038\,{\scriptsize$\pm$0.014} & 0.028\,{\scriptsize$\pm$0.005} & 0.086\,{\scriptsize$\pm$0.058} & $-5.95$ & 751 & $-0.055$\,{\scriptsize$\pm$0.057} & $-0.44$\,{\scriptsize$\pm$0.32} & $-0.039$\,{\scriptsize$\pm$0.115} \\
 Qwen3.8-Max-Preview$^{\dagger}$ & \cellcolor{bestcell}\textbf{0.83\,{\scriptsize$\pm$0.20}} & 0.066\,{\scriptsize$\pm$0.010} & 0.083\,{\scriptsize$\pm$0.018} & \cellcolor{bestcell}\textbf{0.256\,{\scriptsize$\pm$0.091}} & $+4.24$ & 773 & $+0.001$\,{\scriptsize$\pm$0.070} & \cellcolor{bestcell}\textbf{$+0.07$\,{\scriptsize$\pm$0.37}} & $-0.062$\,{\scriptsize$\pm$0.025} \\
 GLM 5.2 (high)$^{\dagger}$ & 1.43\,{\scriptsize$\pm$0.10} & 0.115\,{\scriptsize$\pm$0.026} & 0.080\,{\scriptsize$\pm$0.015} & 0.058\,{\scriptsize$\pm$0.050} & $-8.35$ & 703 & $-0.076$\,{\scriptsize$\pm$0.030} & $-0.44$\,{\scriptsize$\pm$0.17} & $-0.007$\,{\scriptsize$\pm$0.025} \\
 Kimi K3$^{\dagger}$ & 1.51\,{\scriptsize$\pm$0.11} & 0.096\,{\scriptsize$\pm$0.053} & 0.062\,{\scriptsize$\pm$0.030} & 0.046\,{\scriptsize$\pm$0.042} & $-8.50$ & 733 & $-0.084$\,{\scriptsize$\pm$0.065} & $-0.48$\,{\scriptsize$\pm$0.31} & $-0.043$\,{\scriptsize$\pm$0.065} \\
 Claude Opus 4.7 (max) & 1.42\,{\scriptsize$\pm$0.41} & 0.031\,{\scriptsize$\pm$0.019} & 0.024\,{\scriptsize$\pm$0.013} & 0.122\,{\scriptsize$\pm$0.086} & $-2.74$ & 222 & \cellcolor{bestcell}\textbf{$+0.011$\,{\scriptsize$\pm$0.063}} & $-0.01$\,{\scriptsize$\pm$0.39} & $-0.004$\,{\scriptsize$\pm$0.009} \\
 Claude Opus 4.6 (max) & 1.27\,{\scriptsize$\pm$0.25} & 0.051\,{\scriptsize$\pm$0.024} & 0.039\,{\scriptsize$\pm$0.015} & 0.202\,{\scriptsize$\pm$0.117} & $-2.83$ & 380 & $-0.052$\,{\scriptsize$\pm$0.039} & $-0.38$\,{\scriptsize$\pm$0.26} & $-0.092$\,{\scriptsize$\pm$0.219} \\
 GLM 5.1$^{\dagger}$ & 1.56\,{\scriptsize$\pm$0.14} & 0.092\,{\scriptsize$\pm$0.046} & 0.059\,{\scriptsize$\pm$0.028} & 0.012\,{\scriptsize$\pm$0.024} & $-7.52$ & 496 & $-0.096$\,{\scriptsize$\pm$0.048} & $-0.50$\,{\scriptsize$\pm$0.37} & $-0.105$\,{\scriptsize$\pm$0.194} \\
 \makecell[l]{DeepSeek-V4-Pro-Preview\\ (max)$^{\dagger}$} & 1.23\,{\scriptsize$\pm$0.21} & 0.037\,{\scriptsize$\pm$0.017} & 0.032\,{\scriptsize$\pm$0.017} & 0.149\,{\scriptsize$\pm$0.143} & $-3.24$ & 473 & $-0.042$\,{\scriptsize$\pm$0.027} & $-0.19$\,{\scriptsize$\pm$0.15} & $-0.011$\,{\scriptsize$\pm$0.174} \\
 Gemini 3.5 Flash & 0.92\,{\scriptsize$\pm$0.46} & 0.029\,{\scriptsize$\pm$0.033} & 0.023\,{\scriptsize$\pm$0.017} & 0.122\,{\scriptsize$\pm$0.065} & $+1.25$ & 175 & $-0.003$\,{\scriptsize$\pm$0.058} & $-0.05$\,{\scriptsize$\pm$0.30} & \cellcolor{bestcell}\textbf{$+0.003$\,{\scriptsize$\pm$0.037}} \\
 Qwen3.7-Max$^{\dagger}$ & 1.43\,{\scriptsize$\pm$0.30} & 0.062\,{\scriptsize$\pm$0.036} & 0.041\,{\scriptsize$\pm$0.019} & 0.144\,{\scriptsize$\pm$0.158} & $-5.42$ & 365 & $-0.045$\,{\scriptsize$\pm$0.026} & $-0.31$\,{\scriptsize$\pm$0.22} & $-0.144$\,{\scriptsize$\pm$0.323} \\
 Gemini 3.1 Pro & 1.38\,{\scriptsize$\pm$0.18} & 0.082\,{\scriptsize$\pm$0.014} & 0.060\,{\scriptsize$\pm$0.009} & 0.081\,{\scriptsize$\pm$0.061} & $-4.05$ & 201 & $-0.064$\,{\scriptsize$\pm$0.061} & $-0.88$\,{\scriptsize$\pm$1.18} & $-0.056$\,{\scriptsize$\pm$0.105} \\
 GLM 5.2 (max)$^{\dagger}$ & 1.36\,{\scriptsize$\pm$0.20} & 0.117\,{\scriptsize$\pm$0.064} & 0.084\,{\scriptsize$\pm$0.038} & 0.065\,{\scriptsize$\pm$0.070} & $-6.90$ & 747 & $-0.066$\,{\scriptsize$\pm$0.025} & $-0.41$\,{\scriptsize$\pm$0.25} & $-0.117$\,{\scriptsize$\pm$0.094} \\
 Kimi K2.6$^{\dagger}$ & 1.55\,{\scriptsize$\pm$0.44} & 0.051\,{\scriptsize$\pm$0.036} & 0.029\,{\scriptsize$\pm$0.017} & 0.071\,{\scriptsize$\pm$0.069} & $-2.76$ & 107 & $-0.077$\,{\scriptsize$\pm$0.081} & $-0.82$\,{\scriptsize$\pm$0.62} & $-0.055$\,{\scriptsize$\pm$0.397} \\
 Qwen3.6-Plus$^{\dagger}$ & 1.33\,{\scriptsize$\pm$0.30} & 0.050\,{\scriptsize$\pm$0.012} & 0.040\,{\scriptsize$\pm$0.014} & 0.096\,{\scriptsize$\pm$0.089} & $-2.60$ & 224 & $-0.077$\,{\scriptsize$\pm$0.081} & $-0.38$\,{\scriptsize$\pm$0.34} & $-0.007$\,{\scriptsize$\pm$0.149} \\
 Qwen3.5-Plus$^{\dagger}$ & 1.86\,{\scriptsize$\pm$0.16} & 0.040\,{\scriptsize$\pm$0.035} & 0.023\,{\scriptsize$\pm$0.021} & 0.109\,{\scriptsize$\pm$0.141} & $-3.01$ & 48 & $-0.050$\,{\scriptsize$\pm$0.042} & $-1.22$\,{\scriptsize$\pm$0.80} & $-0.044$\,{\scriptsize$\pm$0.076} \\
\bottomrule
\end{tabular}
\end{adjustbox}
\end{table}

%% file: tables/tab_4_4_modes.tex
\begin{table}[!htbp]
\centering
\footnotesize
\setlength{\tabcolsep}{4pt}
\renewcommand{\arraystretch}{1.12}
\caption{Recurring failure modes, the mechanism behind each and the mechanical
rule that detects it. \emph{Models} counts the models of 18 with at least one
flagged episode and \emph{Eps.} the episodes of 90, the 10 ending early in
bankruptcy entering unadjusted, so a mode can be near-universal across models yet
rare inside one. Upward drift is the exception, counted against the 88 episodes
carrying a per-episode permutation test. Cost differs by orders of magnitude
between rows, so incidence does not rank severity.}
\label{tab:failure_modes}
\begin{adjustbox}{max width=\textwidth}
\begin{tabular}{@{}p{2.55cm}p{4.55cm}p{4.75cm}rr@{}}
\toprule
Mode & Mechanism & Detection rule & Models & Eps. \\
\midrule
\textit{Cash timing}\newline Bankruptcy & Procurement, storage and the tiered operating charge debit the bank as they fall due while a sale reaches the wallet nine days after shipment, so the account empties before revenue returns & Episode terminated on ten consecutive daily closes with a negative bank balance, \texttt{profitability.bankrupt} & 4 & 10 \\
Overdraft the wallet covered & Settled revenue stays in the platform wallet until an explicit \texttt{withdraw}, which costs 10 simulated minutes and no money & A negative bank close whose shortfall the platform wallet alone would have cleared, terminal post-settlement day excluded & 6 & 10 \\
\addlinespace[2pt]
\textit{Adversarial}\newline Paid a pre-deal fraudulent supplier & A membership fee, a manufactured deadline or a discount promised on the next order raises the price agreed before any goods move & Nonzero spend against \texttt{vip\_fee}, \texttt{fake\_urgency} or \texttt{future\_discount} in \texttt{spend\_by\_fraud\_type}, which records money handed to suppliers of a type and not the amount extracted & 16 & 32 \\
Paid a post-deal fraudulent supplier & A fraudulent supplier ships part of the paid quantity or a defective lot, neither of which is visible when the deal is signed & Nonzero spend against \texttt{qty\_bait} or \texttt{quality\_downgrade} in the same field, read the same way & 16 & 35 \\
Fraudulent supplier re-ordered from & A short or defective delivery is never charged back to the counterpart that sent it, so the next restock repeats the same deal & Two or more agreements on one (supplier, SKU) pair with \texttt{supplier\_type} \texttt{bad} inside one episode & 16 & 38 \\
\addlinespace[2pt]
\textit{Bargaining}\newline Opening quote accepted unimproved & The supplier's first number becomes the settlement price, forfeiting the concession the kernel would have granted over further rounds & Honest agreement whose \texttt{final\_price} is at or above the counterpart's \texttt{initial\_offer} & 15 & 58 \\
Upward drift across re-orders & The best price already won on an item is not carried into the next order for it, so each restock pays more than the running minimum & Within-episode permutation test on re-order prices, $z_{\mathrm{anchor}} \le -1.96$, stricter than the pooled test behind Table~\ref{tab:res_d6}, so a model flagged there can show no episode here & 16 & 44 \\
\addlinespace[2pt]
\textit{Information}\newline Catalog query repeated verbatim & Research whose tool result left the context window is derived again rather than written to memory. \texttt{list\_products} returns exactly what it returned before, while \texttt{supplier\_search} re-shuffles one supplier list on every call, minus any supplier gone bankrupt & Identical \texttt{supplier\_search} or \texttt{list\_products} arguments issued twice in one episode & 18 & 76 \\
\addlinespace[2pt]
\textit{Fulfillment}\newline Deadline cancellation, first 14 days & Sold orders go unshipped for two days and cancel, losing the sale and taking a reputation penalty, in the opening fortnight before any routine is established & An order cancelled for the two-day ship deadline inside the first 14 daily triggers, 2026-01-02 to 2026-01-15 & 12 & 21 \\
Deadline cancellation, last 14 days & Fulfillment is dropped during the year-end wind-down while the stores still sell, so the closing week cancels what it sold & The same cancellation inside the last 14 triggers, 2026-12-19 to the 2027-01-01 settlement trigger, which itself cancels whatever is unshipped, so this row mixes abandoned fulfillment with end-of-episode settlement & 8 & 11 \\
\bottomrule
\end{tabular}
\end{adjustbox}
\end{table}